\documentclass[10pt,journal,compsoc]{IEEEtran}
\ifCLASSOPTIONcompsoc
  \usepackage[nocompress]{cite}
\else
  \usepackage{cite}
\fi
\ifCLASSINFOpdf
\else
\fi

\usepackage[utf8]{inputenc} 
\usepackage[T1]{fontenc}    
\usepackage[pagebackref=true,breaklinks=true,colorlinks,bookmarks=false]{hyperref}
\usepackage{url}            
\usepackage{booktabs}       
\usepackage{amsfonts}       
\usepackage{nicefrac}       
\usepackage{microtype}      
\usepackage{xcolor}         
\usepackage{acronym}
\usepackage{multirow}
\usepackage{nccmath}
\usepackage{subfig}
\usepackage{graphicx}
\usepackage{wrapfig, soul}
\usepackage[rightcaption]{sidecap}
\sidecaptionvpos{figure}{c}
\usepackage{enumitem}
\usepackage{array}
\usepackage{tcolorbox}

\newcommand{\etal}{\textit{et al. }}
\newcommand{\state}{state-of-the-art }

\usepackage{bbding} 
\newacro{ai}[AI]{artificial intelligence} 
\newacro{ml}[ML]{machine learning} 
\newacro{dl}[DL]{deep learning}
\newacro{sgd}[SGD]{stochastic gradient descent}
\newacro{ae}[AE]{adversarial example}
\newacro{aa}[AA]{adversarial attack}
\newacro{at}[AT]{adversarial training}
\newacro{dnn}[DNN]{deep neural network}
\newacro{cnn}[CNN]{convolutional neural network}
\newacro{nn}[NN]{neural network}
\newacro{nlp}[NLP]{natural language processing}
\newacro{vit}[ViT]{vision transformer}
\newacro{t2t}[T2T-ViT]{tokens-to-token ViT}
\newacro{tnt}[TNT]{transformer-in-transformer}
\newacro{fastrcnn}[Fast-RCNN]{fast region convolutional neural network}
\newacro{pgd}[PGD]{projected gradient descent}
\newacro{autopgd}[Auto-PGD]{auto projected gradient descent}
\newacro{fgsm}[FGSM]{fast gradient sign method}
\newacro{bpda}[BPDA]{backward pass differentiable approximation}
\newacro{cw}[CW]{Carlini-Wagner}
\newacro{mim}[MI-FGSM]{momentum iterative gradient-based method}
\newacro{saga}[SAGA]{self-attention gradient attack}
\newacro{mlp}[MLP]{multi-layer perceptron}
\newacro{uap}[UAP]{universal adversarial perturbations}
\newacro{jsm}[JSMA]{jacobian saliency map attack}
\newacro{sa}[SA]{square attack}
\newacro{aa}[AA]{autoattack}
\newacro{rays}[RayS]{ray searching}
\newacro{fab}[FAB]{fast adaptive boundary}
\newacro{bfg}[L-BFGS]{limited memory Broyden-Fletcher-Goldfarb-Shanno}
\newacro{bim}[BIM]{basic iterative method}
\newacro{df}[DF]{DeepFool}
\newacro{eot}[EOT]{expectation over transformation}
\newacro{asr}[ASR]{attack success rate}
\newacro{ss}[SS]{local spatial smoothing}
\newacro{nlm}[NLM]{non-local mean}
\newacro{tvm}[TVM]{total variation minimization}
\newacro{cr}[CR]{cropping and re-scaling}
\newacro{jpg}[JPEG]{JPEG compression}
\newacro{ccp}[CCP]{color channel perturbations}
\newacro{dct}[DCT]{discrete cosine transform}
\newacro{gcam}[Grad-CAM]{gradient-weighted class activation mapping}
\newacro{ssim}[SSIM]{structural similarity index measure}
\newacro{mad}[MAD]{most apparent distortion}
\newacro{bit}[BiT]{big transfer}
\newacro{mhsa}[MHSA]{multi-head self-attention}
\newacro{shsa}[SHSA]{single-head self-attention}
\newacro{psnr}[PSNR]{peak signal to noise ratio}
\newacro{gelu}[GELU]{Gaussian error linear units}

\begin{document}
%
\title{Reveal of Vision Transformers Robustness against Adversarial Attacks}
%
%
%
%

\author{Ahmed~Aldahdooh,Wassim~Hamidouche,and~Olivier~D\'eforges
\IEEEcompsocitemizethanks{\IEEEcompsocthanksitem All authors are with INSA Rennes, CNRS, IETR - UMR 6164, University of Rennes, 35000, Rennes, France.\protect\\
\IEEEcompsocthanksitem Corresponding author: A. Aldahdooh,\\E-mail: \href{mailto:ahmed.aldahdooh@insa-rennes.fr}{ahmed.aldahdooh@insa-rennes.fr}\\
\IEEEcompsocthanksitem This work has been submitted to the IEEE for possible publication. Copyright may be transferred without notice, after which this version may no longer be accessible.}
}

\markboth{IEEE Transactions on XXXX}%
{Aldahdooh \MakeLowercase{\textit{et al.}}: Reveal of Vision Transformers Robustness against Adversarial Attacks}
%



\IEEEtitleabstractindextext{%
\begin{abstract}
The major part of the vanilla \acf{vit} is the attention block that brings the power of mimicking the global context of the input image. For better performance, \ac{vit} needs large-scale training data. To overcome this data hunger limitation, many \ac{vit}-based networks, or hybrid-\ac{vit}, have been proposed to include local context during the training. The robustness of \acp{vit} and its variants against adversarial attacks has not been widely investigated in the literature like \acp{cnn}. This work studies the robustness of \ac{vit} variants 1) against different $L_p$-based adversarial attacks in comparison with \acp{cnn}, 2) under \acfp{ae} after applying preprocessing defense methods and 3) under the adaptive attacks using \acf{eot} framework. To that end, we run a set of experiments on 1000 images from ImageNet-1k and then provide an analysis that reveals that vanilla \ac{vit} or hybrid-\ac{vit} are more robust than \acp{cnn}. For instance, we found that 1) Vanilla \acp{vit} or hybrid-\acp{vit} are more robust than \acp{cnn} under $L_p$-based attacks and under adaptive attacks. 2) Unlike hybrid-\acp{vit}, Vanilla \acp{vit} are not responding to preprocessing defenses that mainly reduce the high frequency components. Furthermore, feature maps, attention maps, and Grad-CAM visualization jointly with image quality measures, and perturbations' energy spectrum are provided for an insight understanding of attention-based models. 
\end{abstract}

\begin{IEEEkeywords}
Vision transformer, convolutional neural network, robustness, adversarial attacks, deep learning.
\end{IEEEkeywords}}

\maketitle

\IEEEdisplaynontitleabstractindextext

%
\IEEEpeerreviewmaketitle

\acresetall


%
%
%
%
\section{Introduction}\label{sec:introduction}
\IEEEPARstart{I}{mage} classification task models have remarkable progress in its prediction accuracy especially when convolutional blocks serve as the main building block of the model \cite{he2016deep}. Convolutional blocks have the ability to exploit the spatial features and in particular the low-level features \cite{yuan2021tokens}. On the other hand, self-attention blocks in Transformers \cite{vaswani2017attention} showed great success in \ac{nlp} models \cite{vaswani2017attention}, and recently, Dosovitskiy \etal proposed \ac{vit}, vanilla \ac{vit}, the first image classification model that uses the pure transformer encoder blocks \cite{dosovitskiy2020image} and image patches, as tokenization, to build the classifier. To overcome the lack of the inductive biases inherent to \acp{cnn}, it was shown that \ac{vit} achieves better performance than \state \acp{cnn} models of similar capacity, such as ResNet~\cite{he2016deep} and its variants if \ac{vit} is trained with significantly large-scale training datasets, such as JFT-300M \cite{sun2017revisiting,dosovitskiy2020image}. Moreover, \ac{vit} models that are trained on large-scale datasets can be downgraded to smaller datasets, such as ImageNet-1k \cite{imagenet_cvpr09}, via transfer learning, leading to performance comparable to or better than \state \acp{cnn} models. Given the advantage of the \ac{vit} and on the other side its limitation to the huge need of the data, other models that combine the vanilla \ac{vit} with other modules, like \ac{t2t} \cite{yuan2021tokens}, \ac{tnt} \cite{han2021transformer}, and CvT \cite{wu2021cvt} models were proposed for better learning the low-level features in the transformer and hence, reduce their dependency on large datasets. These models are also known as hybrid-\ac{vit} models

Given many \ac{cnn} image classification task models \cite{he2016deep,simonyan2014very,szegedy2016rethinking, howard2017mobilenets}, their properties are well studied and analyzed including the robustness against \acp{ae} \cite{szegedy2013intriguing}. \ac{ae} is a combination of the original image and a carefully crafted perturbation \cite{szegedy2013intriguing}. This perturbation is hardly perceptible to humans, while it causes the \ac{dl} model to misclassify the input image. Since the feature space identification of the \ac{ae} is hard to predict ~\cite{carlini2017adversarial,ilyas2019adversarial}, the adversarial attacks threat is very challenging. Adversary can generate \acp{ae} under white box, black box, and gray box attack scenarios~\cite{akhtar2018threat,hao2020adversarial}.

\begin{figure*}[t]
    \begin{center}
        \includegraphics[width=0.95\textwidth, keepaspectratio]{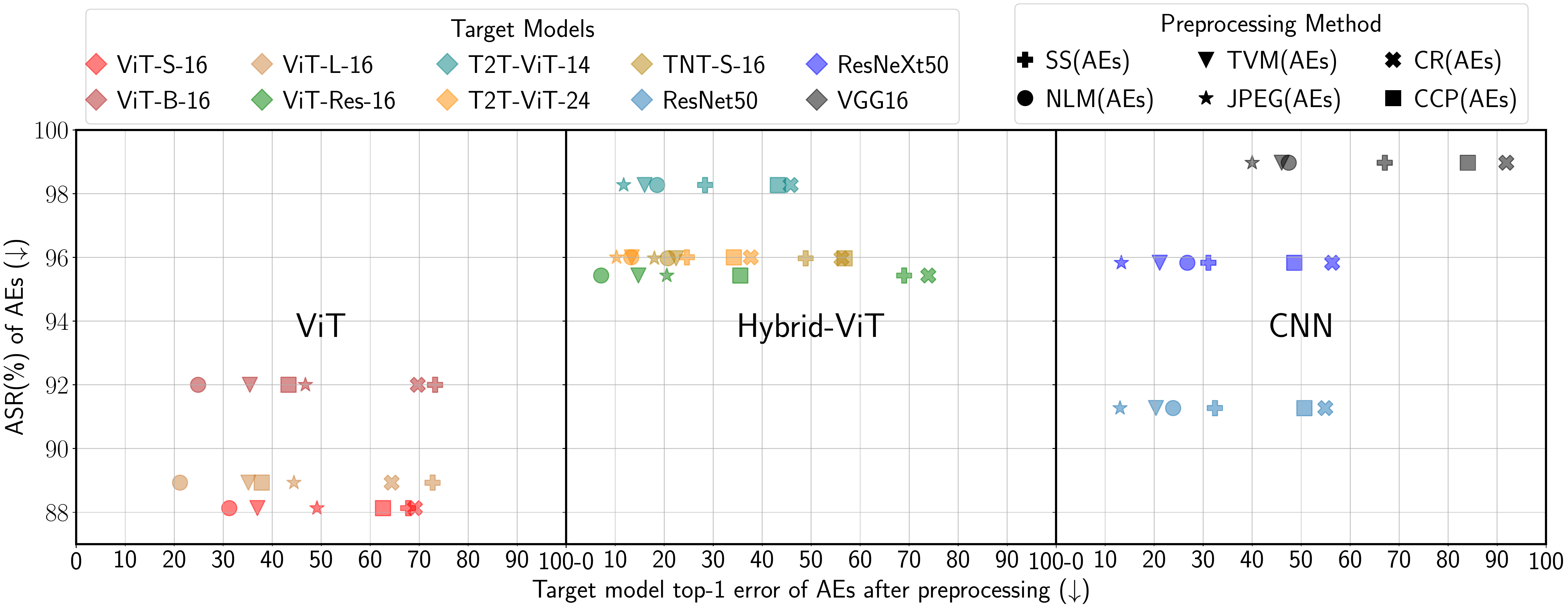}
    \end{center}
    \vspace{-3mm}
    \caption{\begin{footnotesize}The \acf{asr} of target models, on 1000 images from ImageNet-1k, against AutoAttack, in average $\epsilon=\{1,2,4\}/255$, and the target model top-1 error of the preprocessed \acp{ae} for six different preprocessing defense methods including SS: \acl{ss}, NLM: \acl{nlm}, TVM: \acl{tvm}, JPEG: \acl{jpg}, CR: \acl{cr} and CCP: \acl{ccp}. \end{footnotesize}}
    \label{fig:defense_all}
    \vspace{-4mm}
\end{figure*}

Due to the success of \ac{vit} and its variants in various computer vision tasks, the insight properties and robustness studies for such transformers are yet under investigation. Recently, four studies \cite{bhojanapalli2021understanding,shao2021adversarial,mahmood2021robustness,paul2021vision} showed some of these robustness properties for the vision transformers and compared them with the competitive \ac{cnn} models. In this work, we experimentally investigate the robustness of different \ac{vit} variants against different $L_{p}$-based and \ac{ccp} attacks and their strength of predicting the preprocessed \acp{ae}, as defense. The results are compared with competitive \ac{cnn} models. We attend to answer the following research questions 1) Are \ac{vit} variants more robust than \acp{cnn} against $L_0$, $L_1$, $L_2$, and $L_\infty$ based attacks and against \ac{ccp} attack? 2) Are \ac{vit} variants more robust than \acp{cnn} under the preprocessed \acp{ae}? 3) Is increasing the number of attention blocks has an effect on the robustness against the \acp{ae} and under the preprocessed \acp{ae}? 4) Is enhancing \ac{vit} tokenization method has an effect on increasing the robustness against the \acp{ae} and under the preprocessed \acp{ae}? Hence, this work will provide researchers with an in-depth understanding on how vision transformers behave against different attack settings with preprocessing defenses, see Figure \ref{fig:defense_all}. The main contribution of this work is to highlight the following observations:
\begin{itemize}
    \item \ac{vit}-based models are more robust than \acp{cnn} against $L_p$-norm attacks.
    \item Neither larger model's architecture, like \ac{vit}-L, nor bringing convolutional modules for tokenization in \acp{vit}, like \ac{tnt}, will necessarily enhance the robustness.
    \item In general, transferability exists within the model architecture family and the transferability became lower when transfer from large to small variants and vice versa.
    \item Black-box attacks are transferable to \acp{cnn} when they are generated using vanilla \acp{vit} and hybrid-\acp{vit} and not vice versa.
    \item Vanilla \acp{vit} are not responding to preprocessing defenses that mainly reduce the high frequency components, such as \ac{ss} and \ac{jpg}. Hybrid-\acp{vit} are more responsive to preprocessing defenses than vanilla \acp{vit} and \acp{cnn}, see Figure \ref{fig:defense_all}. 
    \item Hybrid-\ac{vit} models show better robustness than vanilla \acp{vit} and ResNets under the \ac{eot} robustness test. 
\end{itemize}

\section{Related work}
Robustness of computer vision \acp{cnn} against \acp{ae} is well studied in the literature~\cite{szegedy2013intriguing}, and many~ countermeasures~\cite{akhtar2018threat,hao2020adversarial,yuan2019adversarial,chakraborty2018adversarial,aldahdooh2021adversarial}, i.e. defenses and detectors, were implemented to characterize the feature space of the \acp{ae}. 
On the other hand, self-attention based computer vision models are under investigation. The work in~\cite{alamri2021transformer} used vanilla self-attention transformer to improve the object detection task detector's robustness against \acp{ae}. The new module used the self-attention transformer to model the features that are extracted from \ac{fastrcnn}~\cite{ren2016faster} before running the detection classifier. Recently, four studies~\cite{bhojanapalli2021understanding,shao2021adversarial,mahmood2021robustness,paul2021vision} provided some understanding of the robustness of \ac{vit}, and its variants. Authors in~\cite{bhojanapalli2021understanding,mahmood2021robustness,paul2021vision}~studied the robustness of the vanilla \acp{vit}, while the work in~\cite{shao2021adversarial} studied the robustness of vanilla \acp{vit} and of other models that combine vanilla \ac{vit} with other modules like convolutional blocks, hybrid-\acp{vit}. The work in~\cite{bhojanapalli2021understanding} investigated the \ac{vit} robustness with respect to the perturbations of both the input and model. The study of input perturbations included natural corruption, real-world distribution shifts, natural adversarial perturbations, adversarial attack perturbations, adversarial spatial perturbations, and texture bias. The adversarial attack perturbations are generated using $L_\infty$ bounded ($\epsilon$) \acp{ae} using \ac{fgsm}~\cite{goodfellow2014explaining} and \ac{pgd}~\cite{madry2017towards} attacks where $\epsilon$ was set to one gray-level (1/255). The input perturbations are tested on models that are pretrained on ImageNet-1k \cite{imagenet_cvpr09}, ImageNet-21k \cite{imagenet21}, and JFT-300M \cite{sun2017revisiting} datasets. It was shown that, like \acp{cnn}, \acp{vit} are vulnerable to \acp{ae} and it is more robust than \acp{cnn} with comparable capacity when trained on sufficient training data. Similarly, the work in~\cite{mahmood2021robustness} investigated the robustness of \ac{vit} and \ac{bit} \ac{cnn}~\cite{kolesnikov2020big} against common corruptions and perturbations, distribution shifts, and natural adversarial examples. The work in \cite{shao2021adversarial} showed that this observation is not necessarily true and claimed that the robustness of \ac{vit} is more related to transformer architecture rather than the pre-training. In \cite{shao2021adversarial}, it was shown that introducing convolutional blocks or token\st{s}-to-token \cite{yuan2021tokens} modules compromises the model robustness when tested against $L_\infty$ \ac{pgd} \cite{madry2017towards} and \ac{autopgd} \cite{croce2020reliable} attacks. Moreover, \cite{shao2021adversarial} showed that \acp{cnn} are less robust than \acp{vit} since \ac{cnn} tends to learn low-level features that are much vulnerable to adversarial perturbations. The three works in~\cite{bhojanapalli2021understanding,shao2021adversarial,mahmood2021robustness} showed that \acp{ae} do not transfer across different model architectures' families which leads to have an ensemble models approach to defend against the attacks. Hence, the work in \cite{mahmood2021robustness}, included more $L_\infty$ attacks like; \ac{cw} \cite{carlini2017towards}, \ac{mim} \cite{dong2018boosting}, and \ac{bpda} \cite{athalye2018obfuscated}, and focused on the transferability of the attacks to introduce the \ac{saga} that can fool the ensemble defense model. In this work, we investigate the robustness against different $L_p$ norms attacks and the robustness under the pre-processed \acp{ae}, as a defense technique.

\section{Preliminaries}
\subsection{Models' architectures}
In this section, a brief description of models that are investigated in our experiments is introduced.  Table \ref{tab:model_var} gives the parameters of the investigated models\footnote{The weights are available \href{https://github.com/rwightman/pytorch-image-models}{here} , for all models except for \acp{t2t} which are available \href{https://github.com/yitu-opensource/T2T-ViT}{here}.}.

\textbf{Vanilla transformer~\cite{dosovitskiy2020image}}. The vanilla self-attention Transformer was introduced for \ac{nlp} tasks. It stacks the encoder and the decoder blocks, where each block stacks $N$ attention blocks. The attention block has two sub-layers. The first one is a multi-head self-attention mechanism, and the second one is a simple, position-wise fully connected feed-forward network. The residual connection around each of the two sub-layers followed by layer normalization are employed. Encoder and decoder input tokens are converted to a sequence of vectors using learned embeddings. Finally, the learned linear transformation and \textit{softmax} function are used to convert the decoder output to predict next-token probabilities.

\textbf{\Acf{vit}~\cite{dosovitskiy2020image}}. It is the first model that employed the vanilla transformer architecture and achieved \state performance on image classification task. To adapt the vanilla transformer for image classification, \ac{vit} 1) divides the input image into a sequence of patches and then linearly projects them to the transformer, 2) appends [CLS] token to the input and output representations that are passed to \ac{mlp} for the classification task. In our investigation, we consider different \ac{vit} models including \ac{vit}-S/B/L-16 models. In the \ac{shsa} of \acp{vit} \cite{dosovitskiy2020image}, each image $X \in \mathbb{R}^{n\times d}$ is presented as a sequence of $n$ patches ($x_1, x_2, \dots x_n$), where $d$ is the embedding dimension to represent each patch. Then, the image sequence, $X$, is linearly projected onto query ($Q=XW_q$), key ($K=XW_k$), and value ($V=XW_v$), where $W_q \in \mathbb{R}^{d\times d_q}$, $W_k \in \mathbb{R}^{d\times d_k}$, and $W_v \in \mathbb{R}^{d\times d_v}$. In this way, the image patch is encoded in terms of the global information, which enables the self-attention structure mechanism to capture the interactions among image patches. By applying scaled dot-product attention mechanism, the output of one self-attention $Z$ is computed as:

\begin{ceqn}
    \begin{equation} \label{eq:attention}
        Z = Attention(Q, K, V) = \textbf{softmax}\left(\frac{Q K^T}{\sqrt{d_k}}\right) V.
    \end{equation}
\end{ceqn}

According to \eqref{eq:attention}, firstly, the attention scores of input query and key are calculated, $S=Q K^T$, which is then normalized by $\sqrt{d}$ to prevent pushing the $\textbf{softmax}$ function into regions where it has extremely small gradients when $d$ becomes large. Then, the probabilities of the normalized scores are calculated, $P=\textbf{softmax}(S/\sqrt{d})$. Finally, the value vector $V$ is multiplied by the probabilities $P$ to calculate $Z$ in which larger probabilities are the focus in the following layers.

The \ac{shsa} limits the capability to highlight the importance of other equally important patches at the same time. Having \acf{mhsa} can mitigate this problem by having different $Q$, $K$, and $V$ vectors for each self-attention, and hence, can jointly attend to information from different representation subspaces at different positions. In \ac{mhsa}, the $Q$, $K$, and $V$ have the size of $X \in \mathbb{R}^{n\times d/h}$, where $h$ is the number of heads, and $d_q=d_k=d_v=d/6$. Hence, the output of \ac{mhsa} has the output of size $X \in \mathbb{R}^{n\times d}$ and is calculated as follows:

\begin{ceqn}
    \begin{equation}\label{eq:mhsa}
    \begin{gathered}
        MultiHead(Q^\prime, K^\prime, V^\prime) = (head_1 \oplus   \dots \oplus head_h) W^o,\\
        head_i = Attention(Q_i, K_i, V_i),  \; \forall i \in \{1, \cdots, h\},
    \end{gathered}
    \end{equation}
\end{ceqn}

where $Q^\prime$, $K^\prime$, $V^\prime$ are the concatenation of $\{Q_i\}_{i=1}^{h}$, $\{K_i\}_{i=1}^{h}$, and $\{V_i\}_{i=1}^{h}$ respectively, and $W^o \in \mathbb{R}^{d\times d}$ is the linear projection matrix. $\oplus$ stands for vector concatenation operation. 


The \ac{vit}'s encoder block stacks $N$ layers of attention blocks. The attention block has two sub-layers. The first one is a \ac{mhsa}, and the second one is a simple \ac{mlp} layer, also known as position-wise fully connected feed-forward network (FFN). Layernorm (LN) is applied before every sub-layer, and residual connections are applied after each sub-layer. This latter consists of two linear transformation layers and a nonlinear activation function, \ac{gelu} \cite{hendrycks2020gaussian}, in between.

The work in \cite{shao2021adversarial} showed that the fact of having \acp{vit} trained on large-scale dataset yield to more robust models than \acp{cnn} is not necessarily true and claimed that the robustness of \acp{vit} is more related to transformer structure rather than the pre-training. Hence, in our investigation, Vanilla \acp{vit} are pretrained on ImageNet \cite{imagenet21} and fine-tuned on ImageNet-1k \cite{imagenet_cvpr09}, while hybrid-\acp{vit} and \ac{cnn} are trained from scratch on ImageNet-1k. 

\begin{table*}[!ht]
\renewcommand{\arraystretch}{1.20}
    \centering
    \caption{\ac{vit}, \ac{vit}-based, and \ac{cnn} model variants that are investigated in this work.}
    \label{tab:model_var}
    \resizebox{0.85\textwidth}{!}{%
    \begin{tabular}{|c|l|c|c|c|c|c|}
        \hline
         \multirow{2}{*}{Category} & \multirow{2}{*}{Model} & \multicolumn{3}{c|}{\ac{vit} Backbone}  & \multirow{2}{*}{Params (M)} & \multirow{2}{*}{\shortstack{Top-1 Acc.(\%)\\on ImageNet-1k}}\\ \cline{3-5}
         & & Layers & Hidden size & \ac{mlp} size & & \\ \hline\hline
         \multirow{3}{*}{Vanilla \ac{vit}$^{\dagger}$} & \ac{vit}-S-16 \cite{rw2019timm} & 8& 786& 2358& 49& 77.858 \\ \cline{2-7}
         & \ac{vit}-B-16 \cite{dosovitskiy2020image} & 12& 786& 3072& 87& 81.786 \\ \cline{2-7}
         & \ac{vit}-L-16 \cite{dosovitskiy2020image}& 24& 1024& 4096& 304& 83.062 \\ \hline\hline
         \multirow{4}{*}{Hybrid-\ac{vit}$^{\ast}$} & \ac{vit}-Res-16 (384) \cite{dosovitskiy2020image} & 12& 786& 3072& 87& 84.972 \\\cline{2-7}
         & \ac{t2t}-14 \cite{yuan2021tokens}& 14& 384& 1152& 22& 81.7 \\ \cline{2-7}
         & \ac{t2t}-24 \cite{yuan2021tokens}& 24& 512& 1536& 64& 82.6 \\ \cline{2-7}
         & \ac{tnt}-S-16 \cite{han2021transformer} & 12& 384& 1536& 24& 81.518 \\ \hline\hline
         \multicolumn{7}{c}{\ac{cnn} Backbone Conv. Layers}\\ \hline\hline
         \multirow{4}{*}{\ac{cnn}$^{\ast}$} & ResNet50 \cite{he2016deep} & \multicolumn{3}{c|}{49}& 23& 79.038 \\ \cline{2-7}
         & \shortstack{ResNet50-32x4d\\(ResNeXt50) \cite{xie2017aggregated}} & \multicolumn{3}{c|}{\multirow{-1.5}{*}{49}}& \multirow{-1.5}{*}{25}& \multirow{-1.5}{*}{79.676} \\ \cline{2-7}
         & VGG16 \cite{simonyan2014very} & \multicolumn{3}{c|}{13}& 138& 71.594 \\ \hline
         \multicolumn{7}{c}{\parbox{0.7\textwidth}{$^{\dagger}$ Pre-trained on ImageNet-21k~\cite{imagenet21} and fine tuned on ImageNet-1k~\cite{imagenet_cvpr09}, \\$^{\ast}$ Trained from scratch on ImageNet-1k~\cite{imagenet_cvpr09}}}
    \end{tabular}
    }%
    \vspace{-5mm}
\end{table*}


\textbf{Hybrid-\ac{vit} models~\cite{dosovitskiy2020image,yuan2021tokens,han2021transformer}}. One key of \acp{cnn} success is the ability to learn local features, while the key success of self-attention based transformer is its ability to learn the global features. Hence, many approaches were introduced to integrate local feature representation in \ac{vit} model. The work in \cite{dosovitskiy2020image}, replaced input image patches with the \ac{cnn} feature map patches and introduced \ac{vit}-Res that used the flattened ResNet feature maps to generate the input sequence. In \cite{yuan2021tokens}, \acf{t2t} is introduced. It replaces input image patches with a layer-wise token\st{s}-to-token (T2T) transformation, then, the features that are learned by T2T module are passed to the \ac{vit}. The aim of T2T module is to progressively structurize the image into tokens by recursively aggregating neighboring tokens into one token. The work in \cite{han2021transformer} proposed a \acf{tnt} architecture to model both patch-level and pixel-level representations. The \ac{tnt} is made up by stacking \ac{tnt} blocks. Each \ac{tnt} block has an inner transformer and an outer transformer. The inner transformer block extracts local features from pixel embeddings. In order to add the output of the inner transformer into the patch embeddings, the output of the inner transformer is projected to the space of patch embedding using a linear transformation layer. The outer transformer block is used to process patch embeddings. In our investigation, we consider \ac{t2t}-14/24, and \ac{tnt}-S-16 models.

\textbf{\Acf{cnn}}. The convolutional layer is the basic building block for the \ac{cnn} models. It has a set of learnable small receptive field filters. These filters are convolved across the full depth of the input to produce the feature maps. Hence, feature maps can remarkably represent local structure of the input image. Many architectures have been proposed in the literature \cite{li2020survey}. In our investigation, we consider ResNet50 \cite{he2016deep}, ResNeXt50\_32x4d \cite{xie2017aggregated}, and VGG16 \cite{simonyan2014very} models.

\subsection{Adversarial attacks}
Our investigation assumes that the adversary can create an \ac{ae} $x^\prime$ by perturbing the input image $x$ with a certain amount of noise $\epsilon$, such that $||x-x^\prime||_p \leq \epsilon$, where $p \in \{0, 1, 2, \dots, \infty\}$, under different attack scenarios; white-, black-, hybrid-, and gray-box attacks. We investigate the target models with 1000 images that are correctly classified by all the target models. The test samples are collected from ImageNet-1k validation dataset \cite{imagenet_cvpr09}. Our investigation, included five white-box attacks: \ac{jsm} \cite{papernot2016limitations}, \ac{fgsm} \cite{goodfellow2014explaining}, \ac{pgd} \cite{madry2017towards}, \ac{uap}~\cite{moosavi2017universal},  and \ac{cw}~\cite{carlini2017towards}. For the black-box attacks, we consider 
\ac{sa} \cite{andriushchenko2020square}, \ac{rays}~\cite{chen2020rays}, and~\ac{ccp}~\cite{colorchannel} attacks. Moreover, we consider ~\ac{aa} \cite{croce2020reliable}, as hybrid-box attack, which is an ensemble attack that runs three attacks to generate the \acp{ae}: the Auto-\ac{pgd}, ac{fab} \cite{croce2020minimally}, and \ac{sa}. For the gray-box attacks, the adversary has knowledge about training data but not the model architecture and depends on the transferability property of the attacks to generate the \acp{ae} using a surrogate model. Given the surrogate model, the aforementioned attacks can be used to generate \acp{ae}. In our investigation, one of our target models will be considered as a surrogate model to generate the \acp{ae} and these \acp{ae} will be tested on the other target models. Table \ref{tab:list_adv_param} lists the attack's parameters that are used in our experiments. All the attacks are generated using adversarial robustness toolbox (ART)\footnote{\url{https://github.com/Trusted-AI/adversarial-robustness-toolbox}}, except \ac{rays} attack that is generated using the official implementation \footnote{\url{https://github.com/uclaml/RayS}}.

In \cite{shao2021adversarial}, a preliminary results were introduced to train robust \ac{vit} classifiers using \ac{at}. It is stated that ``\ac{vit} does not advance the robust accuracy after adversarial training compared to large \acp{cnn} such as WideResNet-34-10''. That is because \ac{vit} may need larger training data or longer training epochs to further improve its robust training performance. Moreover, it is found that \ac{at} can still cause catastrophic overfitting for \ac{vit} when fast \ac{at} training is conducted, to mitigate the overfitting, with \ac{fgsm}, further adjustments were needed to propose \ac{at} for \ac{vit}. Due to such reasons we couldn't conduct a fair \ac{at} comparison with \ac{cnn} and we left it as future work.

\begin{table*}[t]
\renewcommand{\arraystretch}{1.20}
    \centering
    \caption{Considered adversarial attacks and their parameters.}
    \label{tab:list_adv_param}
    \resizebox{0.88\textwidth}{!}{%
    \begin{tabular}{|l|c|c|p{0.52\textwidth}|}
        \hline
         Scenario & Attack & norm & Parameters  \\ \hline\hline
         \multirow{8}{*}{White box} & \ac{fgsm} & $L_\infty$ & $\epsilon \in \{1, 2, 4, 8, 16, 24\}/255 $ \\ \cline{2-4}
         & \ac{pgd} & $L_1$ & $\epsilon \in \{100, 150, 200, 400, 600, 800, 1000\}$, $\epsilon_{step}=\epsilon/10$, \textit{max. iterations}=10 \\ \cline{2-4}
         & \ac{pgd} & $L_2$ & $\epsilon \in \{0.5, 1, 2, 3, 4, 5\}$, $\epsilon_{step}=\epsilon/10$, \textit{max. iterations}=10 \\ \cline{2-4}
         & \ac{pgd} & $L_\infty$ & $\epsilon \in \{1, 2, 4\}/255$, $\epsilon_{step}=\epsilon/10$, \textit{max. iterations}=10 \\ \cline{2-4}
         & \ac{cw} & $L_2$ & \textit{max. iterations}=10, \textit{learning rate}=5e-3, \textit{initial const}=2/255, \textit{binary search steps=}10  \\ \cline{2-4}
         & \ac{cw} & $L_\infty$ & $\epsilon=8/255$, \textit{confidence}=0, \textit{max. iterations}=50, \textit{learning rate}=5e-3  \\ \cline{2-4}
         & \ac{uap} & $L_\infty$ & $\epsilon \in \{1, 2, 4\}/255$, \textit{attacker}=\acs{bim}, $\epsilon_{step}=\epsilon/10$, \textit{max. iterations}=10  \\ \cline{2-4}
         & \ac{jsm} & $L_0$ & $\theta=0.1$, $\gamma=1$,  \\ \hline\hline
          \multirow{2}{*}{Black box}& \ac{sa} & $L_\infty$ & $\epsilon \in \{8, 16\}/255$, $p=0.05$, \textit{max. iterations}=300, \textit{restarts}=1 \\ \cline{2-4}
          & \ac{rays} & $L_\infty$ & $\epsilon =8/255$, \textit{query}=2000\\ \hline\hline
         Hybrid box& \ac{aa} & $L_\infty$ & $\epsilon \in \{1, 2, 4\}/255$ \\ \hline\hline
         Other Attacks& \ac{ccp} & - & \textit{seed}=0 for fixed random-weight based \ac{ccp} attack, $s=2$ and $b=30$.  \\ \hline
    \end{tabular}
    }%
    \vspace{-4mm}
\end{table*}

\subsection{Pre-processing defense methods} \label{sec:defense_methods}
We consider pre-processing methods that were previously applied to the \acp{ae} before being forwarded to the \acp{cnn} in order to alleviate the effect of added perturbations. In our investigation, we consider local smoothing \cite{xu2017feature}, non-local mean denoising \cite{xu2017feature}, total variation minimization \cite{guo2017countering} , JPEG image compression \cite{dziugaite2016study,das2017keeping}, and crop and re-scaling \cite{graese2016assessing}.

\textbf{Local smoothing~\cite{xu2017feature}}. Is a method that uses a pixel neighborhood to smooth out each pixel. For a given $n\times n$ sliding window, local smoothing changes each pixel, center of the sliding window, with the mean, the median, or Gaussian smooth of the window. In \cite{xu2017feature}, it was shown that median filter is more efficient in projecting the \ac{ae} back to the data manifold especially for $L_0$ based attacks, since it is capable of removing sparsely pixels in the input image, and at the same time it preserves edges. In our investigation we use the median filter with $3\times 3$ window size.

\textbf{Non-local mean denoising~\cite{xu2017feature}}. As the name indicates, the non-local smoothing not only uses the nearby pixels, but uses several similar patches within a search window, as well, to smooth out the current patch. While preserving image edges, it is assumed that averaging similar patches to smooth the current patch will remove the perturbations when the mean of the noise is zero. In our investigation, we use the patch size of $7\times 7$ and the search window of $23\times 23$ and the strength of 0.07. We use \textbf{skimage} restoration library to run the non-local mean denoising. For each noisy image, i.e. the \ac{ae}, the robust wavelet-based estimator of the Gaussian noise standard deviation is used before applying the \ac{nlm} denoiser \cite{donoho1994ideal}.

\textbf{Total variation minimization \cite{guo2017countering}}. Total variation denoising was shown to be effective in removing adversarial perturbations due to its ability to minimize the total variation of the image by producing similar images to the noisy images. It was shown that total variation simultaneously preserves edges and denoises flat regions. We use the Chambolle algorithm \cite{chambolle2004algorithm} that is implemented in \textbf{skimage} library with strength of 0.1 and $\epsilon=2e-4$.

\textbf{JPEG image compression \cite{dziugaite2016study,das2017keeping}}. Compression is used to remove redundancy in images by removing high frequency components that are imperceptible to humans. On the other hand, adversarial perturbation is imperceptible and hence it is assumed that the high frequency details of the image are more vulnerable to adversarial perturbations. In our experiments, we compress the \acp{ae} at quality level of 65\%.

\textbf{Cropping and re-scaling \cite{graese2016assessing}}. It was shown that image cropping and re-scaling method is an effective way to remove the adversarial perturbations effect due to the spatial re-positioning of the pixels. This method may harm the structure of the carefully crafted adversarial perturbations. We perform center cropping with 2-pixel margin for top, bottom, left, and right of the image, and then re-scale the cropped image to the input size.

Finally, it is worth mentioning that the pre-processing defense can be easily fooled using \ac{bpda} \cite{athalye2018obfuscated} and \ac{eot} \cite{athalye2018synthesizing} techniques. In our investigation, we consider the \ac{eot} attack to have a deep insight in the robustness of the \acp{vit} based models.

\subsection{Analysis tools}
We rely our investigations on tools that can reveal the properties of the model's behavior such as feature \cite{yuan2021tokens}, attention \cite{dosovitskiy2020image}, and \ac{gcam} \cite{Selvaraju_2019} maps. Moreover, we use \ac{dct} based decomposition of the perturbations \cite{ortizjimenez2020hold} and the visual quality assessment of the \ac{ae} \cite{fezza2019perceptual} as tools to assess the robustness of target models against the \acp{ae}. We consider three objective visual quality assessment metrics including \ac{psnr}, \ac{ssim} \cite{wang2004image}, and \ac{mad} \cite{larson2010most}. 

\textbf{Feature and attention maps:} In \cite{yuan2021tokens}, a visualization method to visualize the feature maps for \acp{cnn} and \acp{vit} is recommended. For simplicity, we get the feature maps of clean and adversarial samples and then visualize the difference of one channel only, randomly selected, and finally normalize it to [0, 1] scale. As recommended in \cite{yuan2021tokens}, input sample is upsampled to clearly visualize the \ac{vit} feature maps. To visualize the attention map, we follow the rollout-based attention visualization method that is provided by \cite{dosovitskiy2020image} and implemented in the ViT-pytorch github repository\footnote{\url{https://github.com/jeonsworld/ViT-pytorch}}. 

\textbf{\ac{gcam} \cite{Selvaraju_2019}} uses the gradients of any target class flowing into the final convolutional layer to produce a coarse localization map highlighting the important regions in the image in the class prediction. Since convolutional blocks are missing in \acp{vit}, \cite{jacobgilpytorchcam} uses the LN output of the last attention block to calculate the gradients and then \cite{jacobgilpytorchcam} reshapes the activation and gradients to 2D spatial images to fit the \ac{gcam} algorithm. 

\begin{figure}[!ht]
    \centering
    \resizebox{0.8\linewidth}{!}{%
    \begin{tabular}{ccc}
         \Large Clean Image& \Large Perturbation from \ac{vit}-B-16 &  \Large Perturbation from ResNet50 \\
         \includegraphics[width=0.35\textwidth]{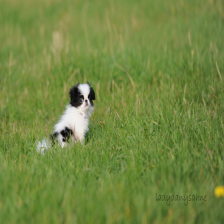}& \includegraphics[width=0.35\textwidth]{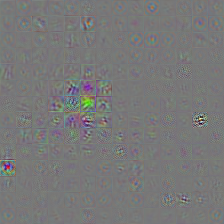}&\includegraphics[width=0.35\textwidth]{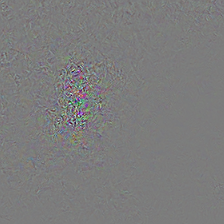}\\
         & \includegraphics[width=0.35\textwidth, trim={0 0 0 1.4cm},clip]{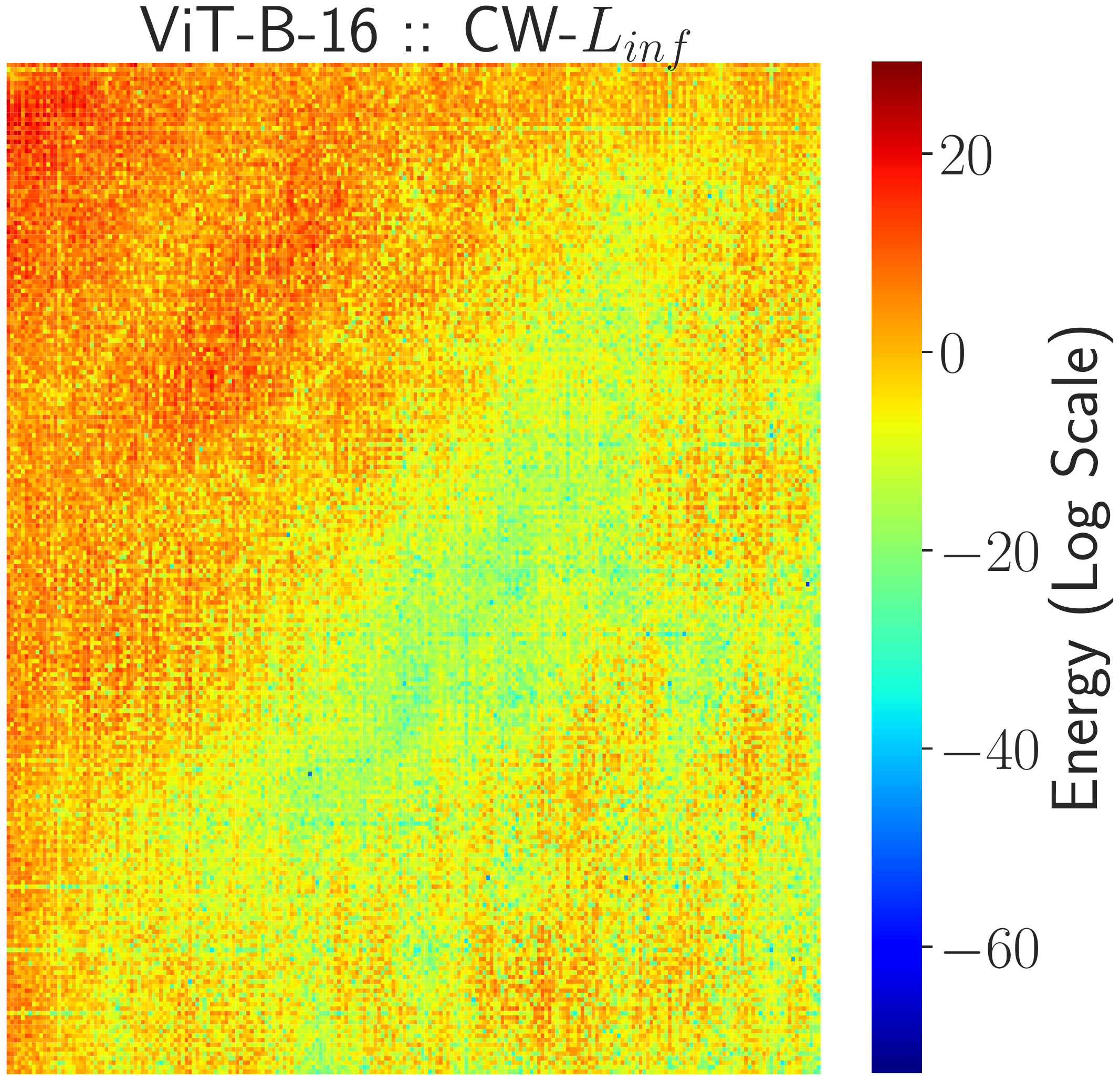}& \includegraphics[width=0.35\textwidth, trim={0 0 0 1.4cm},clip]{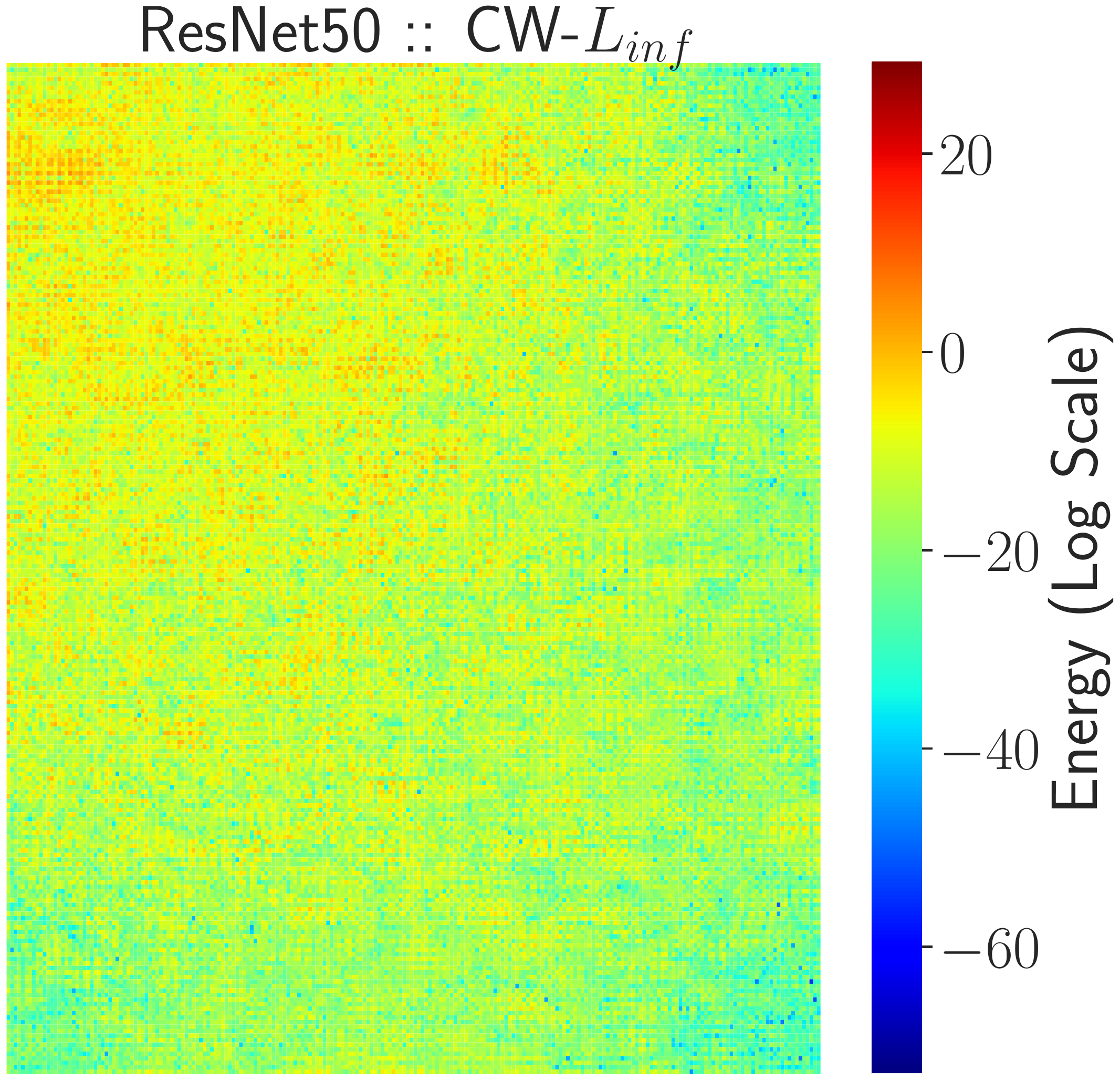}\\
    \end{tabular}
    }%
    \vspace{-3mm}
    \caption{\begin{footnotesize}The perturbation (top), generated using \ac{cw}-$L_\infty$ attack with \ac{vit}-B-16 (middle) and ResNet50(left), and the corresponding \acs{dct}-based spectral decomposition heatmap. Perturbation is scaled from [-1, 1] to [0, 255]. \end{footnotesize}}
    \label{fig:energy_pertub_1}
    \vspace{-3mm}
\end{figure}

\begin{figure}[!ht]
    \centering
    \resizebox{0.8\linewidth}{!}{%
    \begin{tabular}{ccc}
         \Large Clean Image& \Large Perturbation from \ac{t2t}-14 &  \Large Perturbation from ResNet50 \\
         \includegraphics[width=0.35\textwidth]{imgs/clean_imagenet-1k_81.png}& \includegraphics[width=0.35\textwidth]{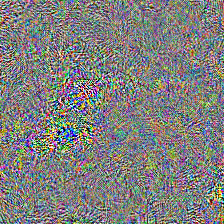}&\includegraphics[width=0.35\textwidth]{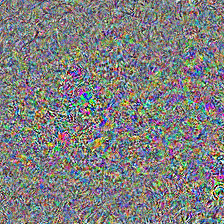}\\
         & \includegraphics[width=0.35\textwidth, trim={0 0 0 1.4cm},clip]{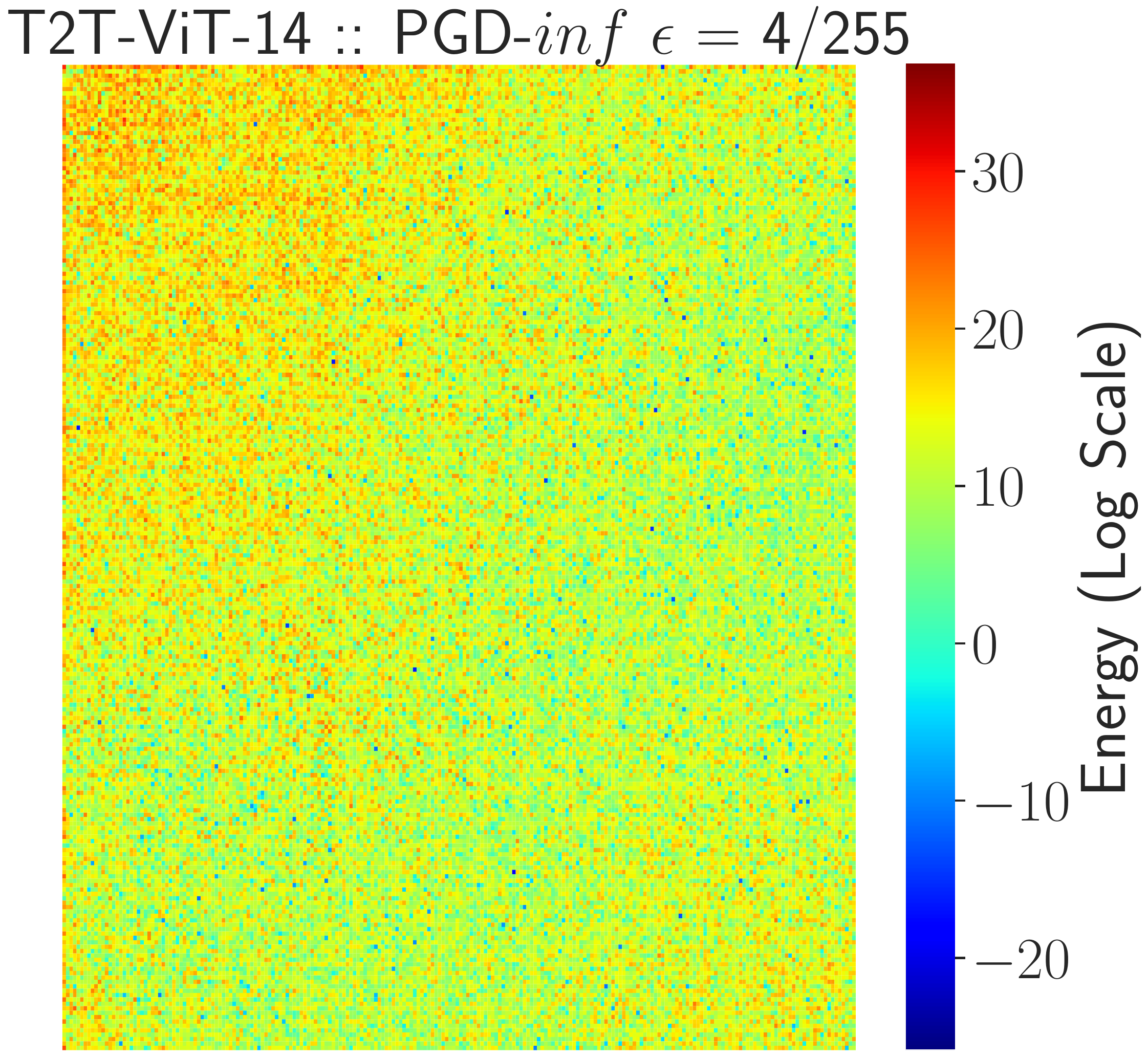}& \includegraphics[width=0.35\textwidth, trim={0 0 0 1.4cm},clip]{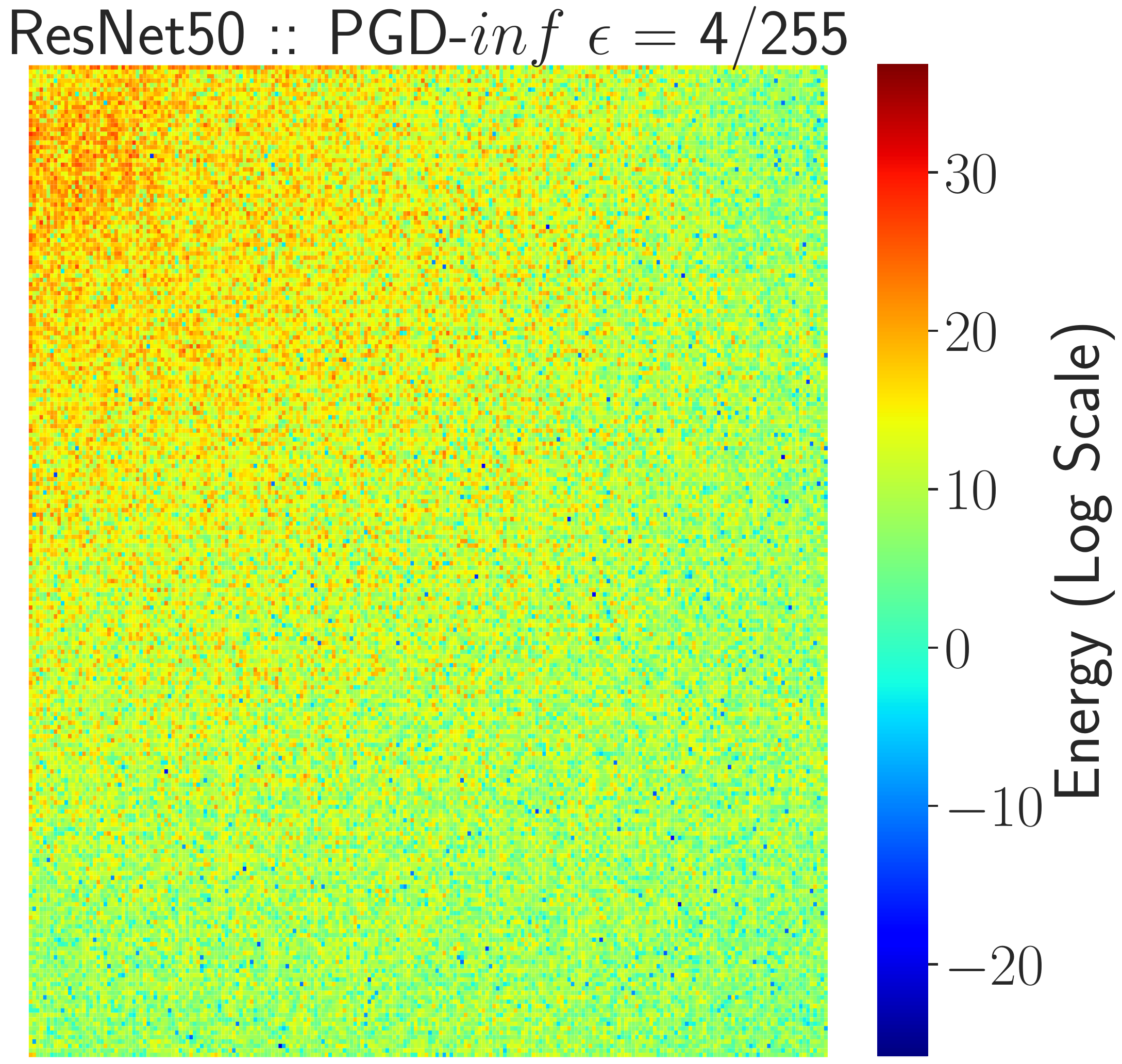}\\
    \end{tabular}
    }%
    \vspace{-3mm}
    \caption{\begin{footnotesize}The perturbation (top), generated using \ac{pgd}-$L_\infty$ $\epsilon=4/255$ attack with \ac{t2t}-14 (middle) and ResNet50(left), and the corresponding \acs{dct}-based spectral decomposition heatmap. Perturbation is scaled from [-1, 1] to [0, 255]. \end{footnotesize}}\vspace{-2mm}
    \label{fig:energy_pertub_2}
\end{figure}

\textbf{\ac{dct}-based decomposition of perturbations}. In \cite{ortizjimenez2020hold}, it was shown that construction of the class decision boundary is extremely sensitive to small perturbations of the training samples.  Moreover, it was shown that \acp{cnn} mainly exploit discriminative features in the low frequencies of MNIST, CIFAR-10, and ImageNet-1k datasets. This explains why, in some query-based black-box attacks like in \cite{sharma2019effectiveness}, using low-frequency perturbations improves the efficiency of the attack's query. To show how we take advantage of this observation in our investigation, we consider the \ac{dct} decomposition of the perturbations for non norm-constrained\footnote{In \cite{ortizjimenez2020hold}, to measure perturbation distance to the decision boundary, they found that non norm-constrained attacks are more suitable for the study. In our experiments, we show that, for ResNet, the perturbations' discriminative features are more centered around low frequency in non norm-constrained attacks.} attacks, like \ac{cw} attacks, and for norm-constrained attacks, like \ac{fgsm} and \ac{pgd}-based attacks as shown in Figure \ref{fig:energy_pertub_1} and Figure \ref{fig:energy_pertub_2}, respectively. For ResNet, it is very clear that the perturbations' discriminative features are more centered around low frequency components. While, for \acp{vit}, the perturbations' discriminative features are spread in all frequency spectrum region. We conclude that, the wider the spread of perturbations' discriminative features the more robust the model against the attacks, because the adversarial attack algorithms have to affect a wider range of frequency spectrum. 

\textbf{\ac{psnr}}. It is the most widely used metric because it is simple and mathematically convenient for optimization. Unfortunately, \ac{psnr} doesn't correlate well with human perception. It is calculated as: 
\begin{ceqn}
    \begin{equation} \label{eq:psnr}
    \begin{gathered}
        PSNR = 10 \; log_{10} \left(\frac{MAX^2}{MSE}\right)\\
        MSE = \frac{1}{N} \sum_{i=1}^{m}\sum_{j=1}^{n} [x(i,j)-x^\prime(i,j)]^2,
        \end{gathered}
    \end{equation}
\end{ceqn}

where $MAX$ is the maximum density value, $2^8-1$ in 8-bit images, $MSE$ is the mean squared error, $m$ and $n$ are the image width and height, $N=mn$, and $\ac{psnr}$ $\in [0, \infty[$. The higher the $\ac{psnr}$ the better the quality and $PSNR \to \infty$ means that the two images are identical. 

\textbf{\ac{ssim}~\cite{wang2004image}}. It is a full reference image quality metric that relies on the fact that the human visual system (HVS) highly tends to extract structural information from the image. \ac{ssim} extracts information that is related to the luminance ($l$), the contrast ($c$), and the structure ($s$) from two images, here $x$ and $x^\prime$, and measures the similarity between them as: 

\begin{ceqn}
    \begin{equation} \label{eq:ssim}
    \begin{gathered}
        SSIM(x,x^\prime)= l(x,x^\prime) c(x,x^\prime) s(x,x^\prime)\\=\frac{(2\mu_x \mu_{x^\prime}+c_1)(2\sigma_{xx^\prime}+c_2)}{(\mu_x^2+\mu_{x^\prime}^2+c_1)(\sigma_x^2+\sigma_{x^\prime}^2+c_2)},
        \end{gathered}
    \end{equation}
\end{ceqn}

where $\mu_x$ is the average of $x$, $\mu_{x^\prime}$ is the average of $x^\prime$, $\sigma_x$ is the variance of $x$, $\sigma_{x^\prime}$ is the variance of $x^\prime$, $\sigma_{xx^\prime}$ is the covariance of $x$ and $x^\prime$, and $c_1$ and $c_2$ are hyperparameter to avoid instability when denominator is close to zero. The range of $\ac{ssim}$ is $[0, 1]$, where $\ac{ssim}=1$ means that the two images are identical. In our experiments, we assume that the target model $\mathcal{M}$ is more robust than other models if the \ac{ae} that is generated using the $\mathcal{M}$ has lower \ac{ssim} score than other models because the adversarial attack algorithm has to generate \ac{ae} with higher perturbation that influences the structure of the \acp{ae}. 

\begin{figure}[t]
    \begin{center}
        \includegraphics[width=0.85\linewidth, keepaspectratio]{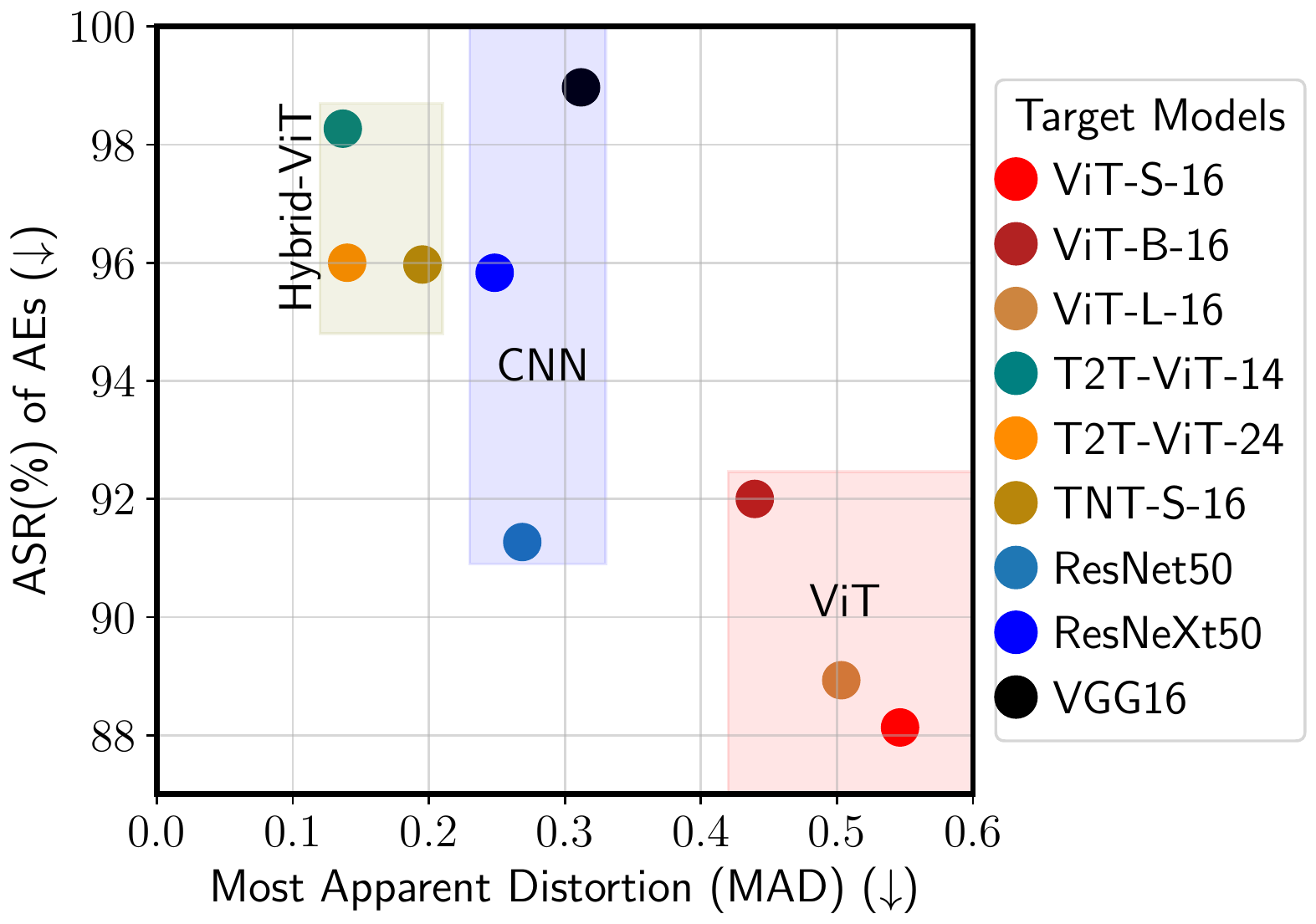}
    \end{center}
    \vspace{-5mm}
    \caption{\begin{footnotesize}The \acf{asr} of target models, on 1000 images from ImageNet-1k against AutoAttack, in average $\epsilon=\{1,2,4\}/255$, and the \acf{mad} \end{footnotesize}}
    \label{fig:asr_vs_mad_autoattack}
    \vspace{-5mm}
\end{figure}

\textbf{\ac{mad} \cite{larson2010most}}. Adversary usually generates \ac{ae} that are imperceptible to human as much as they can by adding perturbations that are near threshold distortions. On the other hand, \ac{mad} attempts to explicitly model two strategies employed by the HVS. The first one is a detection-based strategy for high-quality images containing near threshold distortions\footnote{Some image quality assessment methods assume that there is a distortion level at which the distortions start be visible to the observers. When there is a distortion in an image and is not perceptible, it is called near threshold distortion. When the distortion is visible with high magnitude in the image, it is called suprathreshold distortion.} that are typical situation for \acp{ae} with low perturbation. The second one is an appearance-based strategy for low-quality images containing clearly suprathreshold distortions, that are typical situation for \acp{ae} with high perturbation. As stated in \cite{larson2010most}, local luminance and contrast masking are used to estimate detection-based perceived distortion in high-quality images, whereas changes in the local statistics of spatial-frequency components are used to estimate appearance-based perceived distortion in low-quality images. Moreover, it was shown in \cite{fezza2019perceptual} that \ac{mad} has better correlation to subjective scores for \acp{ae}. The range of \ac{mad} score is $[0, \infty]$ where $\ac{mad}=0$ is close to the clean image. In our experiments, we assume, as shown in Figure \ref{fig:asr_vs_mad_autoattack}, that the target model $\mathcal{M}$ is more robust than other models if the \ac{ae} that is generated using the $\mathcal{M}$ has higher \ac{mad} score than other models because the adversarial attack algorithm don't be able to prevent the perception of the perturbation.

\begin{SCfigure*}[1][t]
\resizebox{0.8\textwidth}{!}{%
    \setlength\tabcolsep{1.5pt}
    \tiny
    \begin{tabular}{ccccccccccc}
    \hline
    Clean &
    ViT-S-16 & 
    ViT-B-16 &
    ViT-L-16 &
    ViT-Res-16 &
    T2T-ViT-14 &
    T2T-ViT-24 &
    TNT-S-16 &
    ResNet50 &
    ResNeXt50 &
    VGG16 \\ \Large
    
    \includegraphics[width=0.07\textwidth]{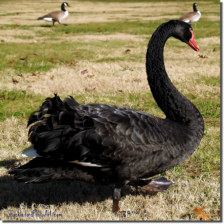} &
    \includegraphics[width=0.07\textwidth]{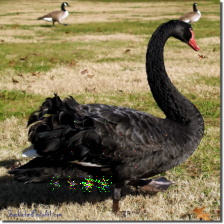} &
    \includegraphics[width=0.07\textwidth]{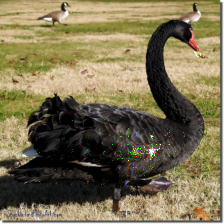} &
    \includegraphics[width=0.07\textwidth]{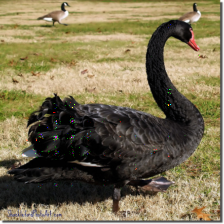} &
    \includegraphics[width=0.07\textwidth]{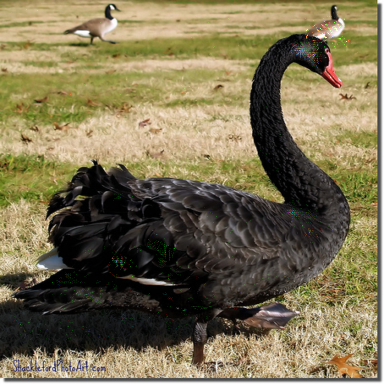} &
    \includegraphics[width=0.07\textwidth]{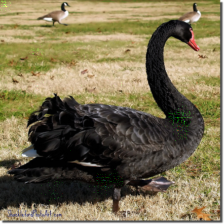} &
    \includegraphics[width=0.07\textwidth]{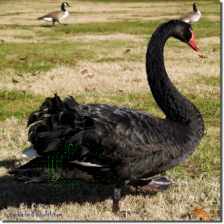} &
    \includegraphics[width=0.07\textwidth]{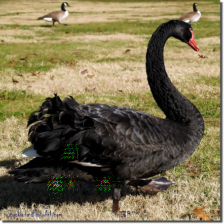} &
    \includegraphics[width=0.07\textwidth]{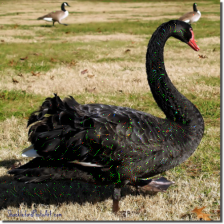} &
    \includegraphics[width=0.07\textwidth]{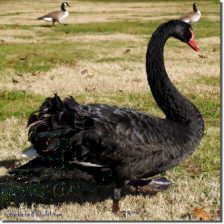} &
    \includegraphics[width=0.07\textwidth]{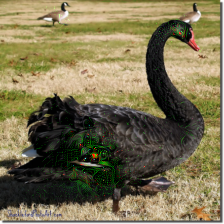} \\ \cline{0-0}
    
    \multirow{-3}{0.07\textwidth}{\subfloat[]{\label{fig:sal_maps_a}}} &
    \includegraphics[width=0.07\textwidth]{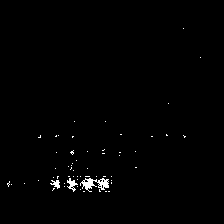} &
    \includegraphics[width=0.07\textwidth]{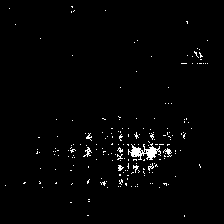} &
    \includegraphics[width=0.07\textwidth]{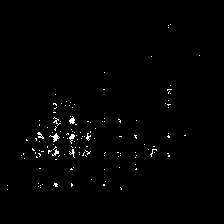} &
    \includegraphics[width=0.07\textwidth]{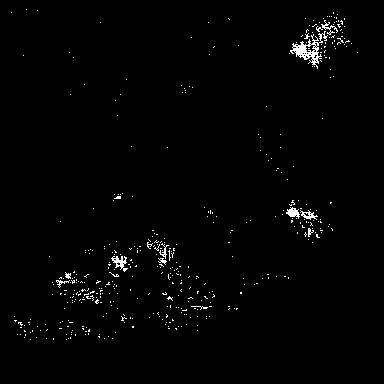} &
    \includegraphics[width=0.07\textwidth]{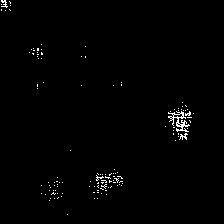} &
    \includegraphics[width=0.07\textwidth]{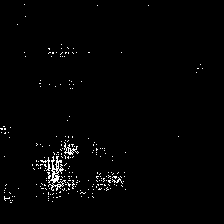} &
    \includegraphics[width=0.07\textwidth]{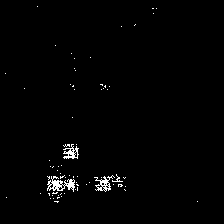} &
    \includegraphics[width=0.07\textwidth]{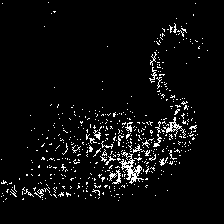} &
    \includegraphics[width=0.07\textwidth]{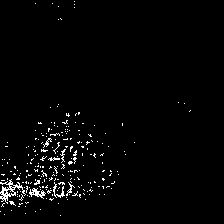} &
    \includegraphics[width=0.07\textwidth]{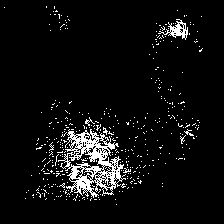} \\
    
    &
    \includegraphics[width=0.07\textwidth]{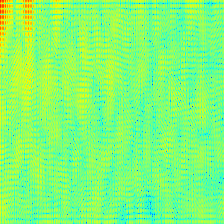} &
    \includegraphics[width=0.07\textwidth]{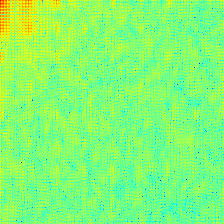} &
    \includegraphics[width=0.07\textwidth]{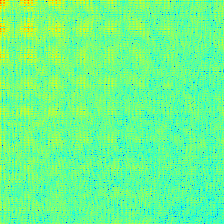} &
    \includegraphics[width=0.07\textwidth]{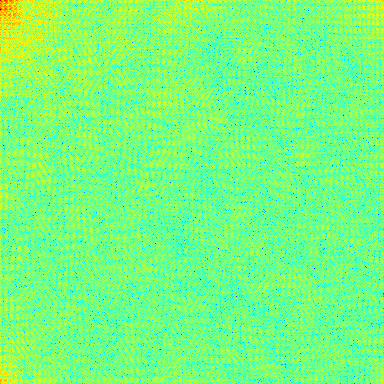} &
    \includegraphics[width=0.07\textwidth]{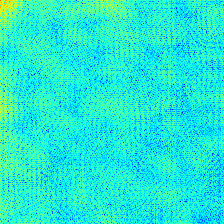} &
    \includegraphics[width=0.07\textwidth]{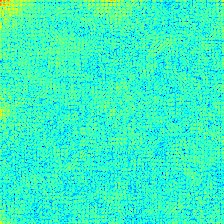} &
    \includegraphics[width=0.07\textwidth]{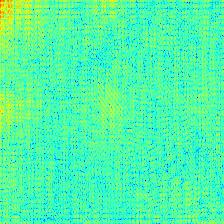} &
    \includegraphics[width=0.07\textwidth]{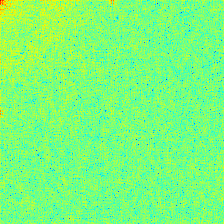} &
    \includegraphics[width=0.07\textwidth]{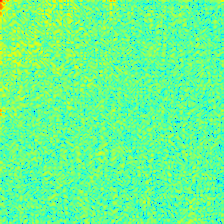} &
    \includegraphics[width=0.07\textwidth]{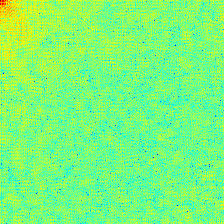} \\ \cline{0-0}
    
    \multirow{-3}{0.08\textwidth}{\subfloat[]{\label{fig:sal_maps_b}}}&
    \includegraphics[width=0.07\textwidth]{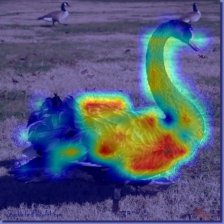} &
    \includegraphics[width=0.07\textwidth]{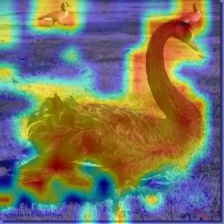} &
    \includegraphics[width=0.07\textwidth]{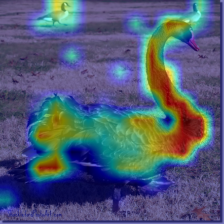} &
    \includegraphics[width=0.07\textwidth]{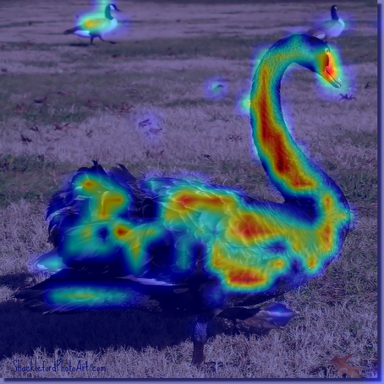} &
    \includegraphics[width=0.07\textwidth]{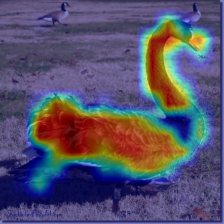} &
    \includegraphics[width=0.07\textwidth]{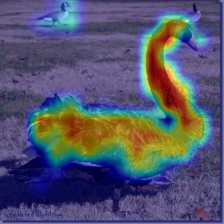} &
    \includegraphics[width=0.07\textwidth]{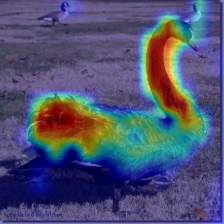} &
    \includegraphics[width=0.07\textwidth]{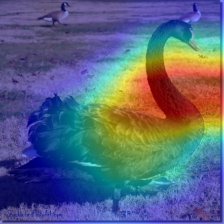} &
    \includegraphics[width=0.07\textwidth]{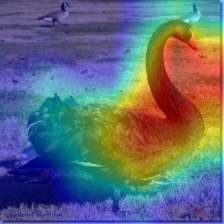} &
    \includegraphics[width=0.07\textwidth]{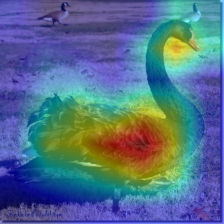} \\
    
    &
    \includegraphics[width=0.07\textwidth]{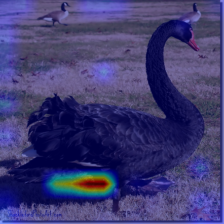} &
    \includegraphics[width=0.07\textwidth]{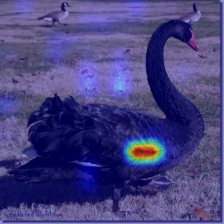} &
    \includegraphics[width=0.07\textwidth]{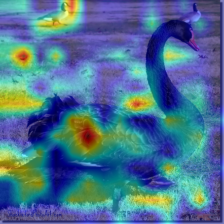} &
    \includegraphics[width=0.07\textwidth]{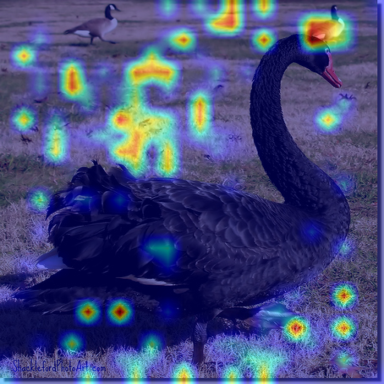} &
    \includegraphics[width=0.07\textwidth]{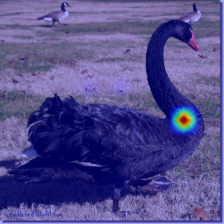} &
    \includegraphics[width=0.07\textwidth]{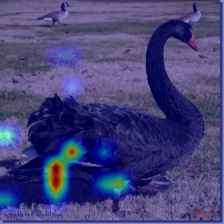} &
    \includegraphics[width=0.07\textwidth]{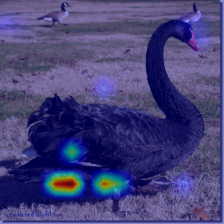} &
    \includegraphics[width=0.07\textwidth]{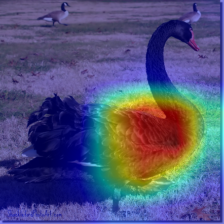} &
    \includegraphics[width=0.07\textwidth]{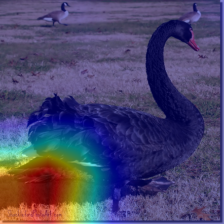} &
    \includegraphics[width=0.07\textwidth]{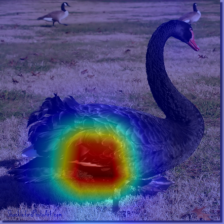} \\ \cline{0-0}
    
    \multirow{-6}{0.08\textwidth}{\subfloat[]{\label{fig:sal_maps_c}}} &
    \includegraphics[width=0.07\textwidth]{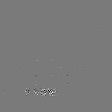} &
    \includegraphics[width=0.07\textwidth]{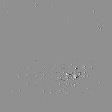} &
    \includegraphics[width=0.07\textwidth]{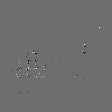} &
    \includegraphics[width=0.07\textwidth]{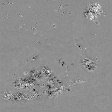} &
    \includegraphics[width=0.07\textwidth]{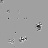} &
    \includegraphics[width=0.07\textwidth]{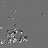} &
    \includegraphics[width=0.07\textwidth]{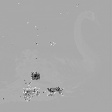} &
    \includegraphics[width=0.07\textwidth]{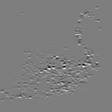} &
    \includegraphics[width=0.07\textwidth]{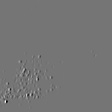} &
    \includegraphics[width=0.07\textwidth]{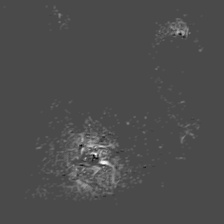} \\ \cline{0-0}
    
    \multirow{-3}{0.08\textwidth}{\subfloat[]{\label{fig:sal_maps_d}}} &
    \includegraphics[width=0.07\textwidth]{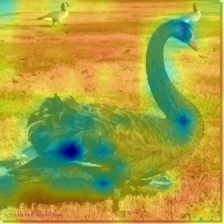} &
    \includegraphics[width=0.07\textwidth]{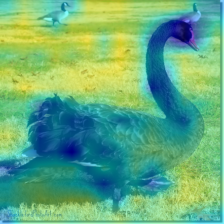} &
    \includegraphics[width=0.07\textwidth]{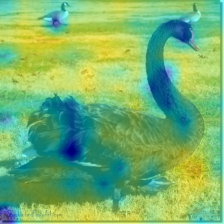} &
    \includegraphics[width=0.07\textwidth]{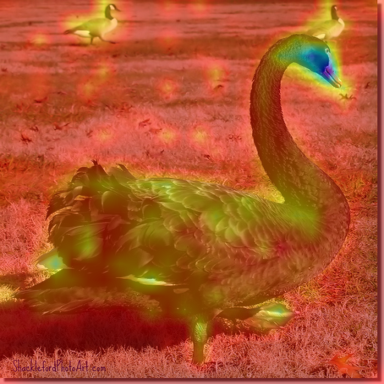} &
    \includegraphics[width=0.07\textwidth]{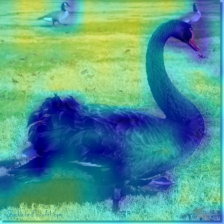} &
    \includegraphics[width=0.07\textwidth]{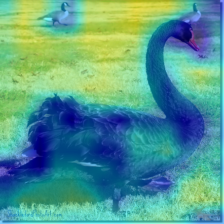} &
    \includegraphics[width=0.07\textwidth]{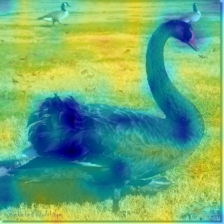} &
    &
    &
    \\
    
    &
    \includegraphics[width=0.07\textwidth]{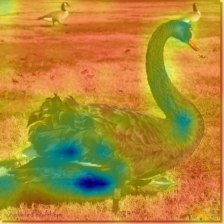} &
    \includegraphics[width=0.07\textwidth]{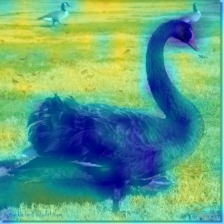} &
    \includegraphics[width=0.07\textwidth]{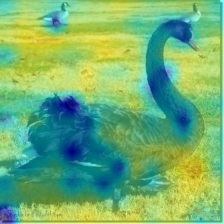} &
    \includegraphics[width=0.07\textwidth]{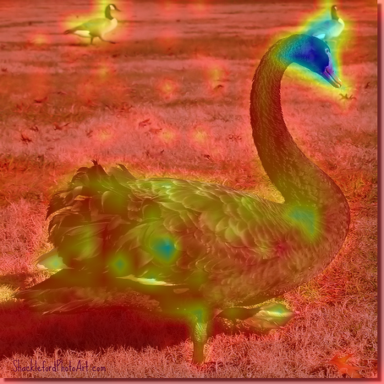} &
    \includegraphics[width=0.07\textwidth]{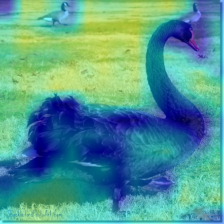} &
    \includegraphics[width=0.07\textwidth]{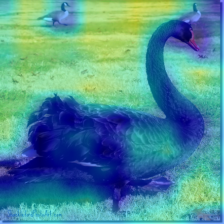} &
    \includegraphics[width=0.07\textwidth]{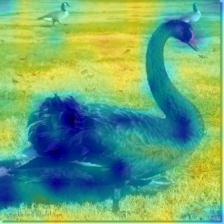} &
    &
    &
    \\ \hline
    \end{tabular}
}%
    
    \hspace{-3mm}\caption{\protect\rule{0ex}{5ex}\begin{footnotesize}\textbf{\ac{jsm} attack:} The first row shows the clean sample and the \acp{ae}. The clean image is correctly classified by tested models and all \acp{ae} are successful attacks. (a) The perturbation (top) and the corresponding \acs{dct}-based spectral decomposition heatmap. Perturbation is shown in black and white colors only. (b) \acs{gcam} of the clean (top) and \ac{ae} samples. (c) Feature map difference between clean and \ac{ae} feature maps that are computed after the first basic block of the model, attention block for \acp{vit} and convolutional layer for \acp{cnn}. (d) The attention map from last attention block for clean (top) and \ac{ae} samples.\end{footnotesize}}\vspace{-5mm}
    \label{fig:sal_maps}
\end{SCfigure*}

\begin{figure*}[h] 
\vspace{-5mm}
\resizebox{\textwidth}{!}{%
    \setlength\tabcolsep{1.5pt}
    \begin{tabular}{cc}
        \begin{tabular}{l}\subfloat[]{\includegraphics[width=0.6\textwidth]{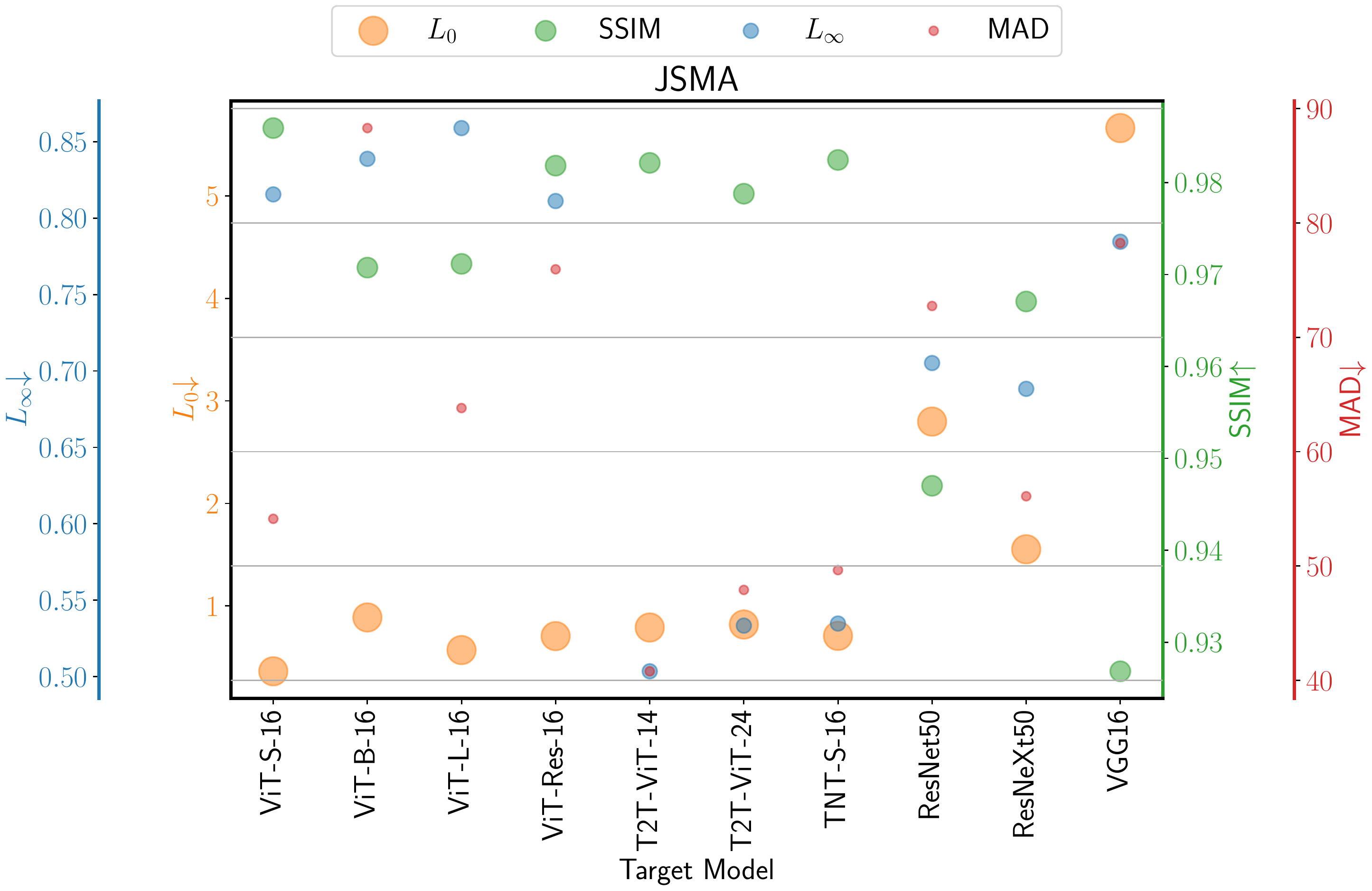}\label{fig:sal_quality}}\end{tabular} &
        \begin{tabular}{l}\subfloat[]{\includegraphics[width=0.4\textwidth]{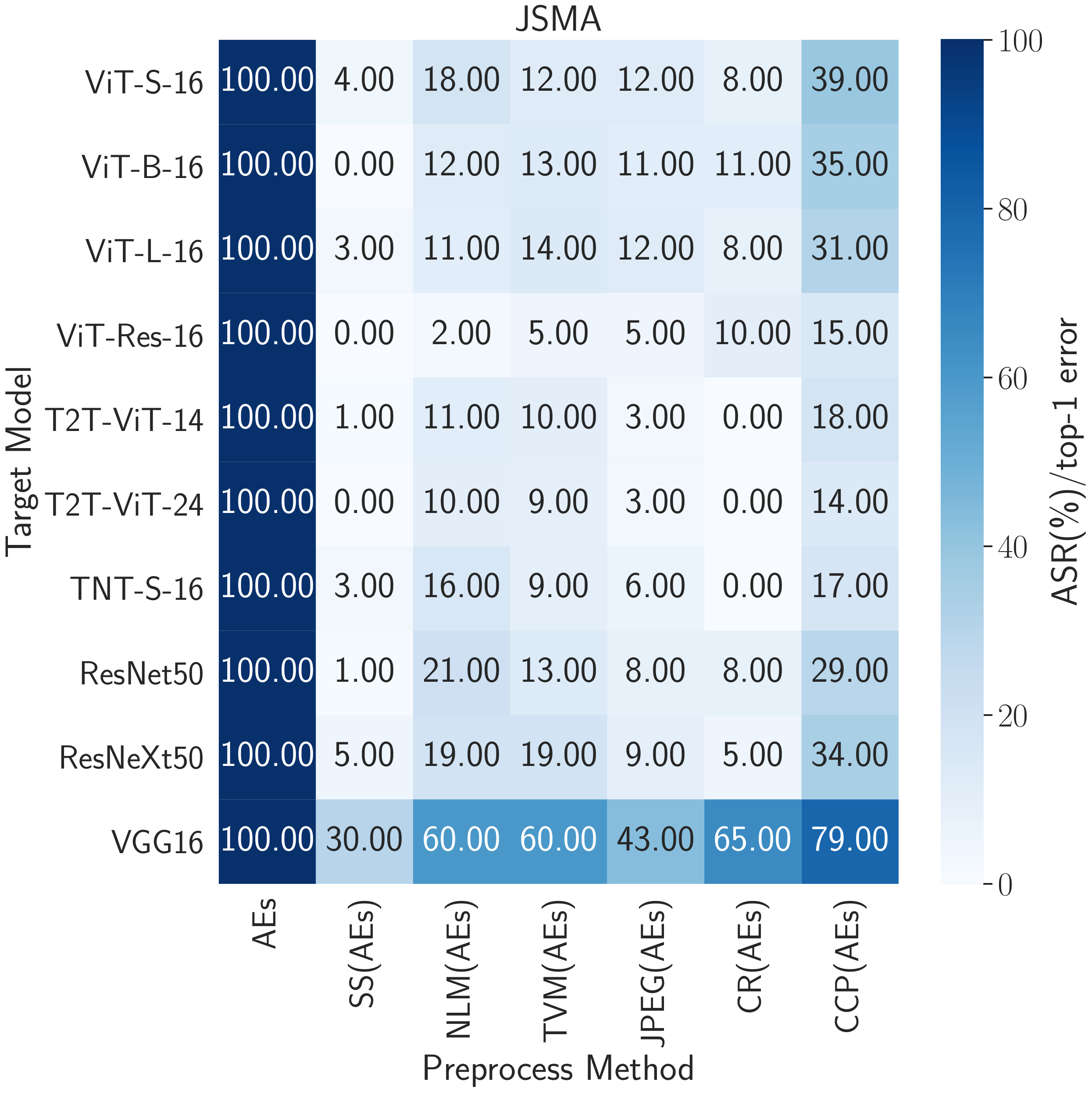}\label{fig:sal_defense}} \end{tabular} \\
    \end{tabular}
}%
\vspace{-8mm}
    \hspace{-3mm}\caption{\protect\rule{0ex}{5ex}\begin{footnotesize}\textbf{\ac{jsm} attack:} (a) \acp{ae} quality assessment measures. (b) The \ac{asr} of the \acp{ae} and the top-1 error of the pre-processed \acp{ae} on 100 images from imagenet-1k. SS: \acl{ss}. NLM: \acl{nlm}. TVM: \acl{tvm}. JPEG: \acl{jpg}. CR: \acl{cr}. CCP: \acl{ccp}. \end{footnotesize}}
    \vspace{-5mm}
    \label{fig:sal_defense_quality}
\end{figure*}

\section{Reveal of the robustness attributes}
\subsection{General observation}
\begin{SCfigure*}[1][t]
\resizebox{0.8\textwidth}{!}{%
    \setlength\tabcolsep{1.5pt}
    \tiny
    \begin{tabular}{ccccccccccc}
    
    \hline
    Clean &
    ViT-S-16 & 
    ViT-B-16 &
    ViT-L-16 &
    ViT-Res-16 &
    T2T-ViT-14 &
    T2T-ViT-24 &
    TNT-S-16 &
    ResNet50 &
    ResNeXt50 &
    VGG16 \\ \Large
    
    \includegraphics[width=0.07\textwidth]{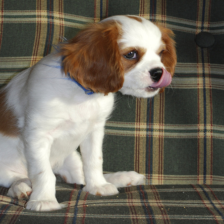} &
    \includegraphics[width=0.07\textwidth]{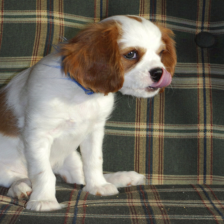} &
    \includegraphics[width=0.07\textwidth]{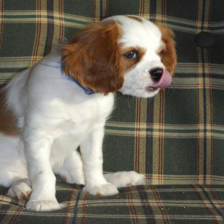} &
    \includegraphics[width=0.07\textwidth]{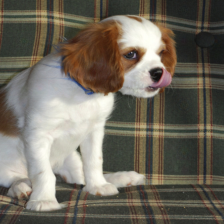} &
    \includegraphics[width=0.07\textwidth]{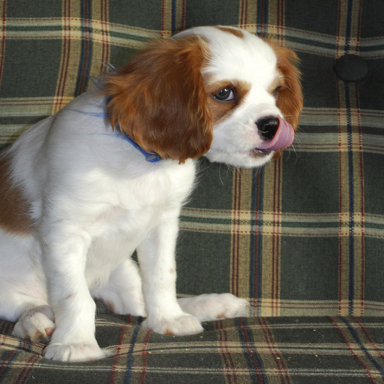} &
    \includegraphics[width=0.07\textwidth]{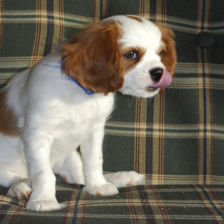} &
    \includegraphics[width=0.07\textwidth]{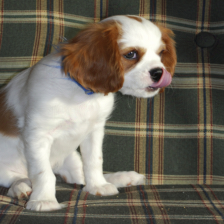} &
    \includegraphics[width=0.07\textwidth]{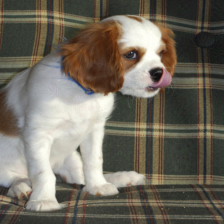} &
    \includegraphics[width=0.07\textwidth]{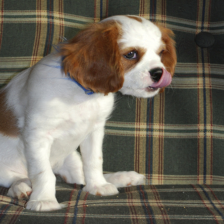} &
    \includegraphics[width=0.07\textwidth]{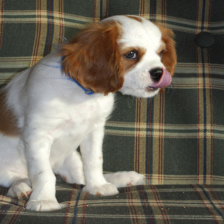} &
    \includegraphics[width=0.07\textwidth]{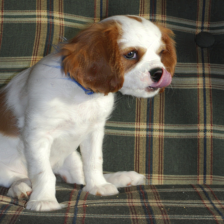} \\ \cline{0-0}
    
    \multirow{-3}{0.08\textwidth}{\subfloat[]{\label{fig:pgd_l1_400_maps_a}}} &
    \includegraphics[width=0.07\textwidth]{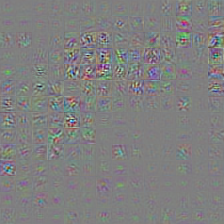} &
    \includegraphics[width=0.07\textwidth]{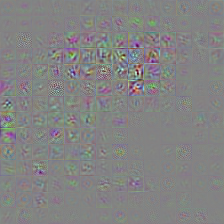} &
    \includegraphics[width=0.07\textwidth]{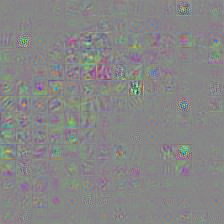} &
    \includegraphics[width=0.07\textwidth]{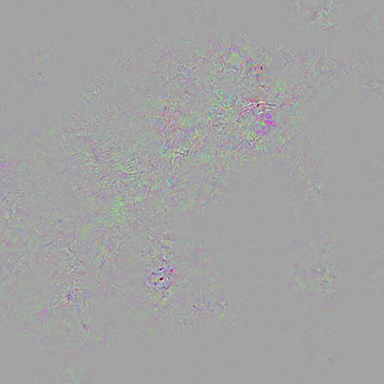} &
    \includegraphics[width=0.07\textwidth]{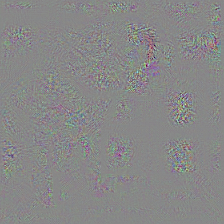} &
    \includegraphics[width=0.07\textwidth]{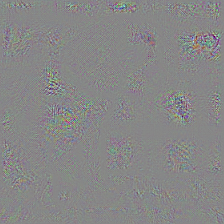} &
    \includegraphics[width=0.07\textwidth]{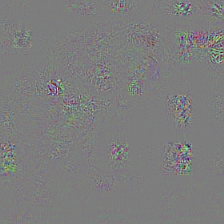} &
    \includegraphics[width=0.07\textwidth]{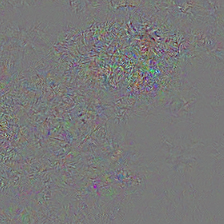} &
    \includegraphics[width=0.07\textwidth]{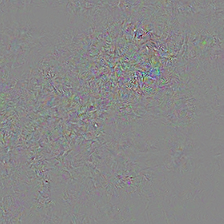} &
    \includegraphics[width=0.07\textwidth]{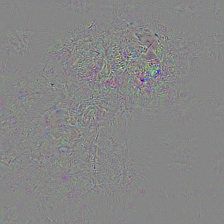} \\
    
    &
    \includegraphics[width=0.07\textwidth]{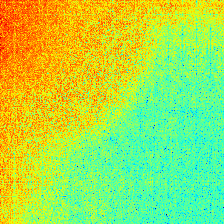} &
    \includegraphics[width=0.07\textwidth]{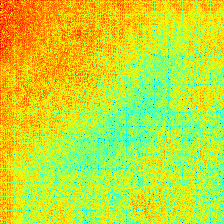} &
    \includegraphics[width=0.07\textwidth]{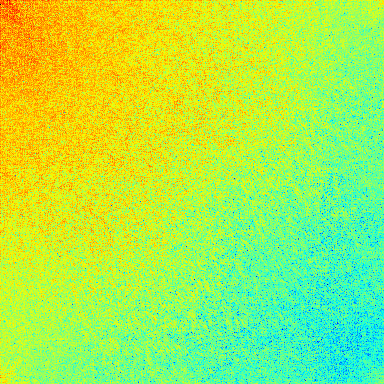} &
    \includegraphics[width=0.07\textwidth]{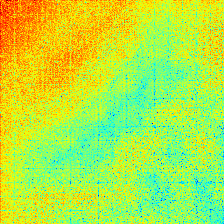} &
    \includegraphics[width=0.07\textwidth]{imgs/energy_perturb_pgd_l1_400_vit_base_resnet50_384_imagenet-1k_87.png} &
    \includegraphics[width=0.07\textwidth]{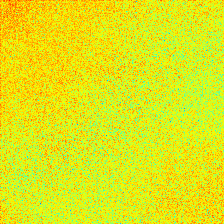} &
    \includegraphics[width=0.07\textwidth]{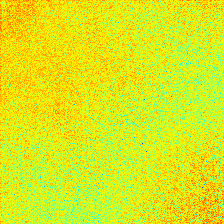} &
    \includegraphics[width=0.07\textwidth]{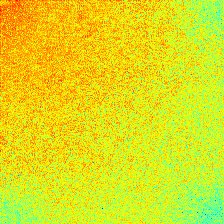} &
    \includegraphics[width=0.07\textwidth]{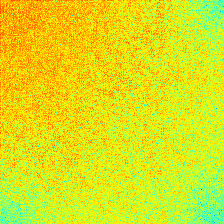} &
    \includegraphics[width=0.07\textwidth]{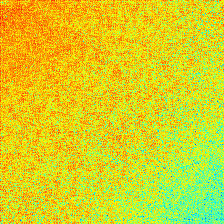} \\ \cline{0-0}
    
    \multirow{-3}{0.08\textwidth}{\subfloat[]{\label{fig:pgd_l1_400_maps_b}}}&
    \includegraphics[width=0.07\textwidth]{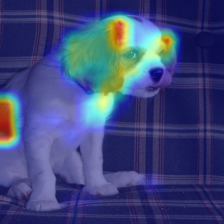} &
    \includegraphics[width=0.07\textwidth]{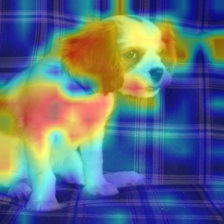} &
    \includegraphics[width=0.07\textwidth]{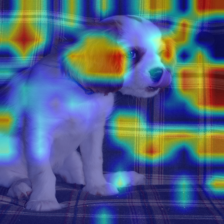} &
    \includegraphics[width=0.07\textwidth]{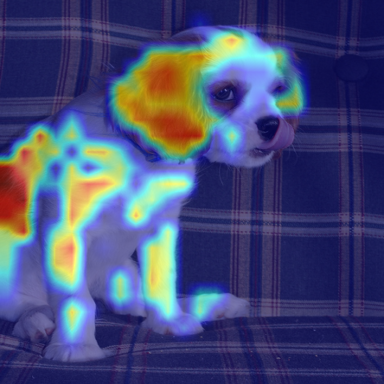} &
    \includegraphics[width=0.07\textwidth]{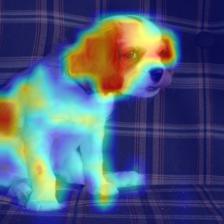} &
    \includegraphics[width=0.07\textwidth]{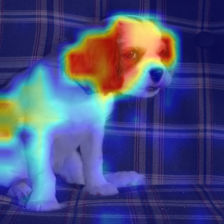} &
    \includegraphics[width=0.07\textwidth]{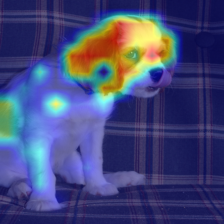} &
    \includegraphics[width=0.07\textwidth]{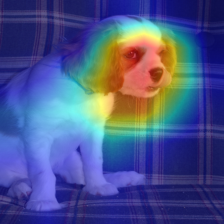} &
    \includegraphics[width=0.07\textwidth]{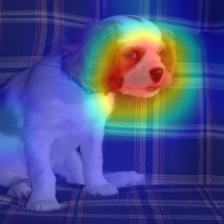} &
    \includegraphics[width=0.07\textwidth]{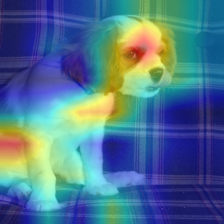} \\
    
    &
    \includegraphics[width=0.07\textwidth]{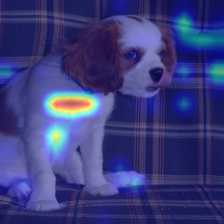} &
    \includegraphics[width=0.07\textwidth]{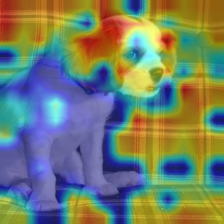} &
    \includegraphics[width=0.07\textwidth]{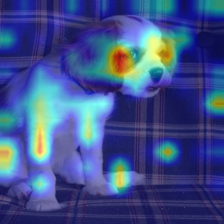} &
    \includegraphics[width=0.07\textwidth]{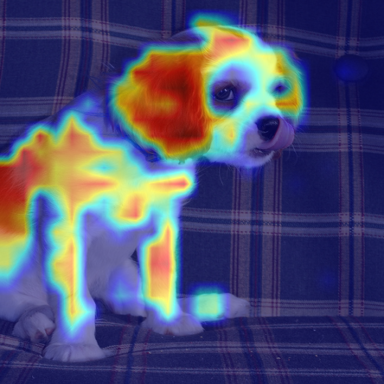} &
    \includegraphics[width=0.07\textwidth]{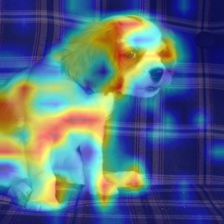} &
    \includegraphics[width=0.07\textwidth]{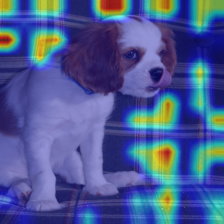} &
    \includegraphics[width=0.07\textwidth]{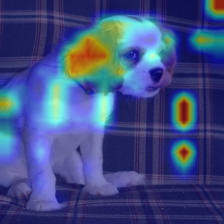} &
    \includegraphics[width=0.07\textwidth]{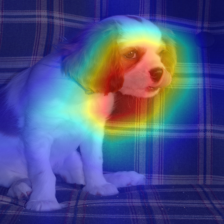} &
    \includegraphics[width=0.07\textwidth]{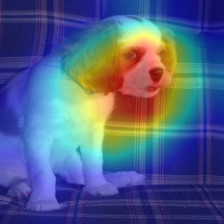} &
    \includegraphics[width=0.07\textwidth]{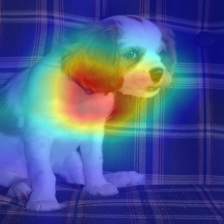} \\ \cline{0-0}
    
    \multirow{-6}{0.08\textwidth}{\subfloat[]{\label{fig:pgd_l1_400_maps_c}}} &
    \includegraphics[width=0.07\textwidth]{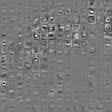} &
    \includegraphics[width=0.07\textwidth]{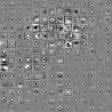} &
    \includegraphics[width=0.07\textwidth]{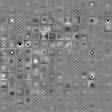} &
    \includegraphics[width=0.07\textwidth]{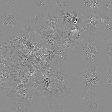} &
    \includegraphics[width=0.07\textwidth]{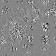} &
    \includegraphics[width=0.07\textwidth]{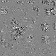} &
    \includegraphics[width=0.07\textwidth]{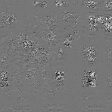} &
    \includegraphics[width=0.07\textwidth]{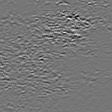} &
    \includegraphics[width=0.07\textwidth]{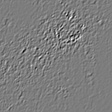} &
    \includegraphics[width=0.07\textwidth]{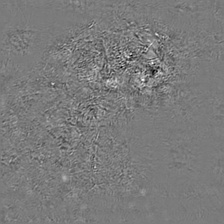} \\ \cline{0-0}
    
    \multirow{-3}{0.08\textwidth}{\subfloat[]{\label{fig:pgd_l1_400_maps_d}}} &
    \includegraphics[width=0.07\textwidth]{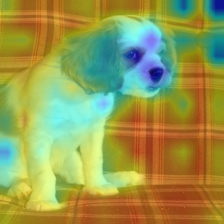} &
    \includegraphics[width=0.07\textwidth]{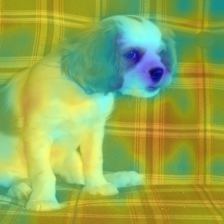} &
    \includegraphics[width=0.07\textwidth]{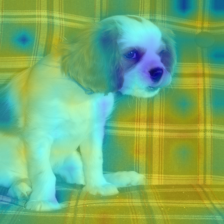} &
    \includegraphics[width=0.07\textwidth]{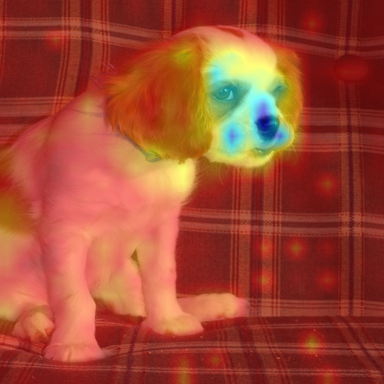} &
    \includegraphics[width=0.07\textwidth]{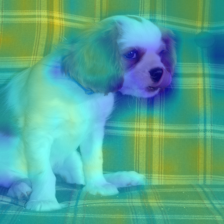} &
    \includegraphics[width=0.07\textwidth]{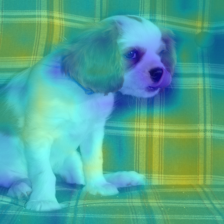} &
    \includegraphics[width=0.07\textwidth]{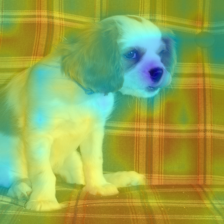} &
    &
    &
    \\
    
    &
    \includegraphics[width=0.07\textwidth]{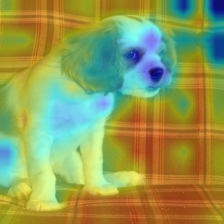} &
    \includegraphics[width=0.07\textwidth]{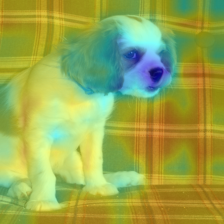} &
    \includegraphics[width=0.07\textwidth]{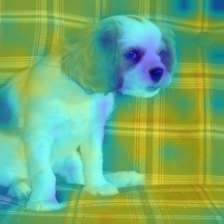} &
    \includegraphics[width=0.07\textwidth]{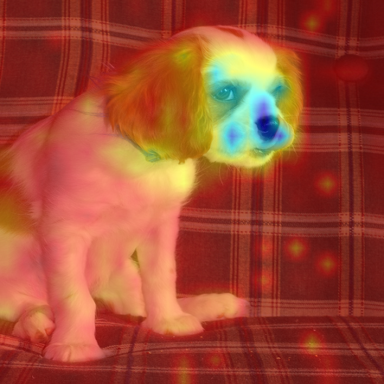} &
    \includegraphics[width=0.07\textwidth]{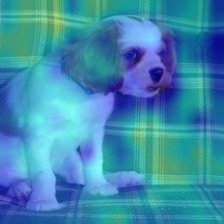} &
    \includegraphics[width=0.07\textwidth]{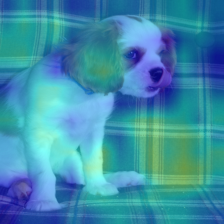} &
    \includegraphics[width=0.07\textwidth]{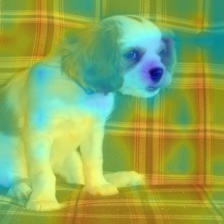} &
    &
    &
    \\ \hline
    \end{tabular}
}%
    \hspace{-3mm}\caption{\protect\rule{0ex}{5ex}\begin{footnotesize}\textbf{\ac{pgd}-$L_1$ $\epsilon=400$ attack:} The first row shows the clean sample and the \acp{ae}. The clean image is correctly classified by tested models and all \acp{ae} are successful attacks. (a) The perturbation (top) and the corresponding \acs{dct}-based spectral decomposition heatmap. Perturbation is scaled from [-1, 1] to [0, 255]. (b) \acs{gcam} of the clean (top) and \ac{ae} samples. (c) Feature map difference between clean and \ac{ae} feature maps that are computed after the first basic block of the model, attention block for \acp{vit} and convolutional layer for \acp{cnn}. (d) The attention map from last attention block for clean (top) and \ac{ae} samples. \end{footnotesize}}
    \vspace{-5mm}
    \label{fig:pgd_l1_400_maps}
\end{SCfigure*}

In Figure \ref{fig:sal_maps} and Figure \ref{fig:pgd_l1_400_maps}, for instance, the first row shows the \acp{ae}, while the top row of Figures \ref{fig:sal_maps_b} and \ref{fig:pgd_l1_400_maps_b}, and Figures \ref{fig:sal_maps_d}, \ref{fig:pgd_l1_400_maps_d} represent the \ac{gcam} and the attention maps of the clean(top) and \acp{ae} imags, respectively. It is clear that \acp{vit} have the capability to track the global features of the input image while \acp{cnn} track local and centered features. Figures \ref{fig:sal_maps_a} and \ref{fig:pgd_l1_400_maps_a} represent the perturbations of the generated \acp{ae} for each model. It is clear that the perturbations highly target the main object of the image, the goose or the dog for both \acp{cnn} and vanilla \acp{vit}. Moreover, the effect of the perturbation can be noticed on the \ac{gcam}. The impacted regions of the attention and \ac{gcam} maps span to regions that are not related to the main object of the image. On the other hand, model's feature maps were known to be affected by the perturbations and propagate when the model goes deeper. As shown in \cite{yuan2021tokens}, \acp{cnn} and hybrid-\acp{vit} tend to better learn low-level features. Figures \ref{fig:sal_maps_c}, and \ref{fig:pgd_l1_400_maps_c} show the difference of clean and \ac{ae} feature maps after the first convolutional layer for \acp{cnn} and after the first attention block for \acp{vit}. As feature maps show, the perturbation impacted learned features, especially the low-level features.


\subsection{Vanilla \acp{vit} are more robust against $L_0$-based attacks}


\begin{figure*}
\vspace{-4mm}
\resizebox{\textwidth}{!}{%
    \setlength\tabcolsep{1.5pt}
    \tiny
    \begin{tabular}{cc}
        \begin{tabular}{l}\subfloat[]{\includegraphics[width=0.6\textwidth]{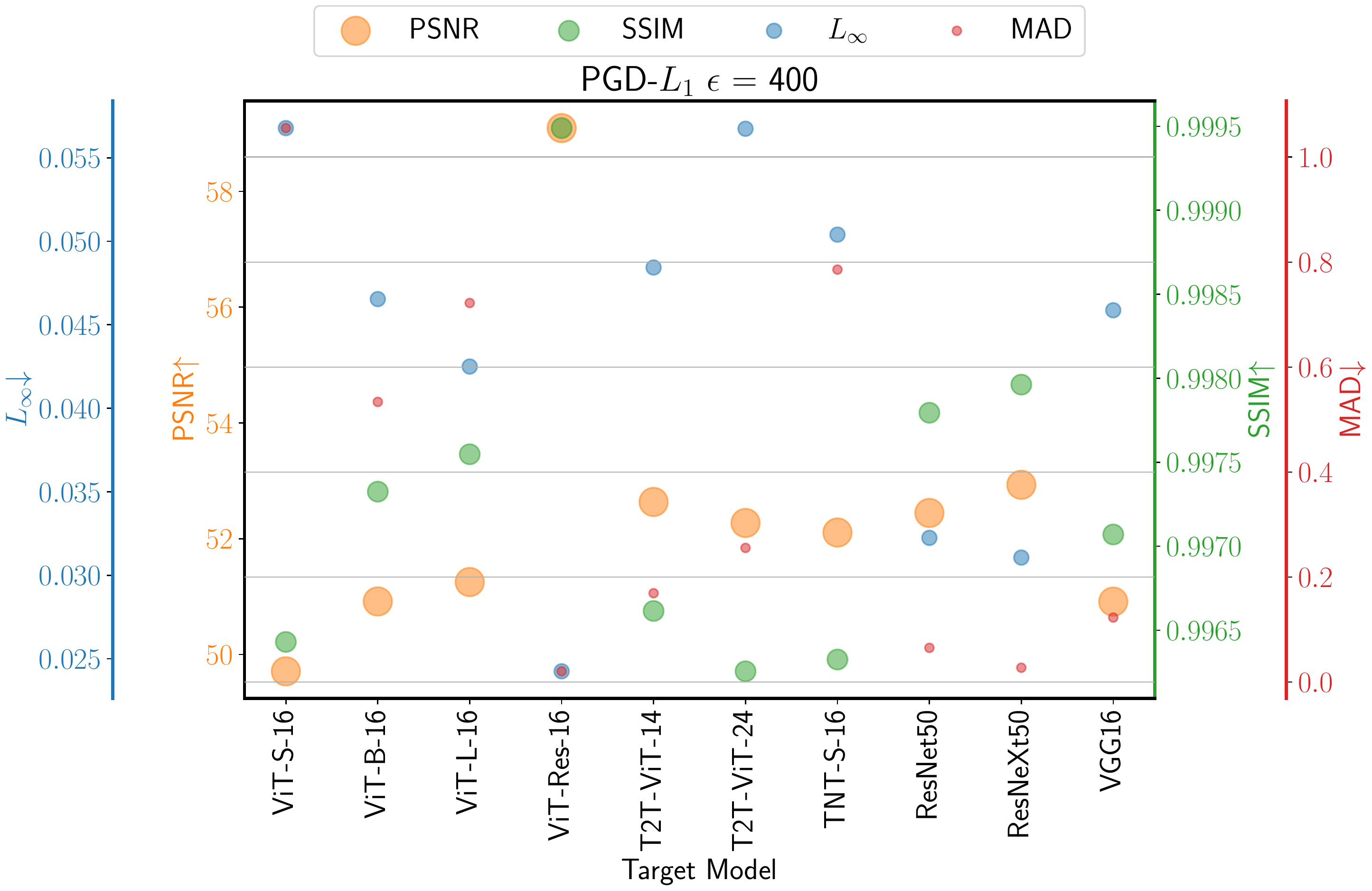}\label{fig:pgd_l1_400_quality}}\end{tabular} &
        \begin{tabular}{l}\subfloat[]{\includegraphics[width=0.4\textwidth]{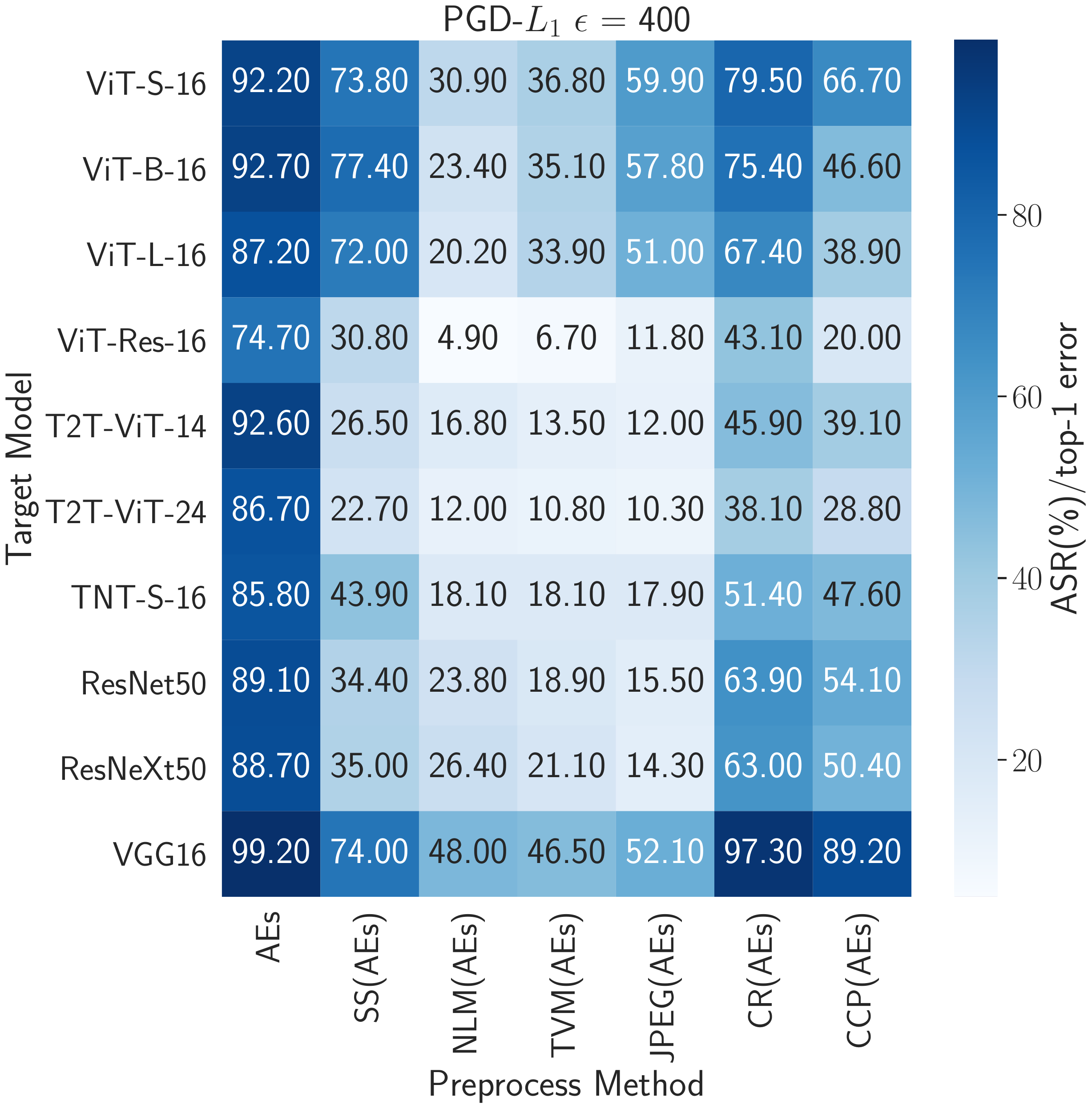}\label{fig:pgd_l1_400_defense}} \end{tabular} \\
    \end{tabular}
}%
    \vspace{-8mm}
    \hspace{-3mm}\caption{\protect\rule{0ex}{5ex}\begin{footnotesize}\textbf{\ac{pgd}-$L_1$ $\epsilon=400$  attack:} (a) \acp{ae} quality assessment measures. (b) The \ac{asr} of the \acp{ae} and  the top-1 error of the pre-processed \acp{ae}  on 1000 images from ImageNet-1k. SS: \acl{ss}. NLM: \acl{nlm}. TVM: \acl{tvm}. JPEG: \acl{jpg}. CR: \acl{cr}. CCP: \acl{ccp}. \vspace{-3mm} \end{footnotesize}}
    \vspace{-3mm}
    \label{fig:pgd_l1_400_defense_quality}
\end{figure*}

$L_0$-based attacks rely on changing few image pixels in order to fool the neural networks. Finding these few pixels is challenging and computationally consuming process since the attack algorithms need to search over the image space to identify these pixels. In this work we consider the white-box \ac{jsm} attack ~\cite{papernot2016limitations}, and generate 100 \acp{ae} from ImageNet-1k validation images and a sample is shown in the first row of Figure \ref{fig:sal_maps}. The results are investigated and the following observations are noted. 

\ac{jsm} identifies input features that significantly impact the classification output. In this experiment, \ac{jsm} achieves 100\% \ac{asr} for all tested models, see the first column of Figure \ref{fig:sal_defense}. Hence, to study the robustness of a model against $L_0$-based attacks, we analyzed 1) the perturbation \ac{dct}-based spectral decomposition \cite{ortizjimenez2020hold}, see Figure \ref{fig:sal_maps_a}, that shows that \acp{vit} have a wider spread energy spectrum which confirms a) \acp{vit} are less sensitive to low-level features and exploit the  global context of the image, and b) wider frequency spectrum has to be affected to generate \ac{ae} for \ac{vit} models and its variants, 2) the visual quality measures of \acp{ae} in Figure \ref{fig:sal_quality} show that \ac{jsm}, for vanilla \acp{vit}, targeted less than 1\% of the pixels ($L_0$) and has to increase the densities ($L_\infty$) of the input pixels to more than 0.8, in order to fool the models, which makes vanilla \acp{vit} more robust against $L_0$-based attacks than hybrid-\acp{vit}. 
\begin{SCfigure*}[1][t]
\centering
\resizebox{0.75\textwidth}{!}{%
    \setlength\tabcolsep{1.5pt}
    \tiny
    \begin{tabular}{ccccccccccc}
    
    \hline
    Clean &
    ViT-S-16 & 
    ViT-B-16 &
    ViT-L-16 &
    ViT-Res-16 &
    T2T-ViT-14 &
    T2T-ViT-24 &
    TNT-S-16 &
    ResNet50 &
    ResNeXt50 &
    VGG16 \\ \Large
    
    \includegraphics[width=0.07\textwidth]{imgs/clean_imagenet-1k_87.png} &
    \includegraphics[width=0.07\textwidth]{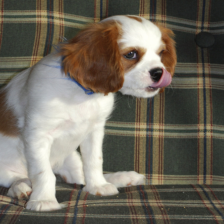} &
    \includegraphics[width=0.07\textwidth]{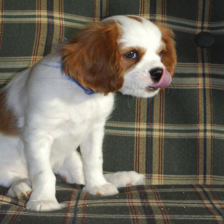} &
    \includegraphics[width=0.07\textwidth]{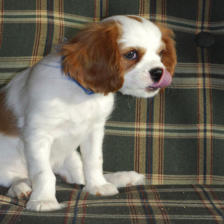} &
    \includegraphics[width=0.07\textwidth]{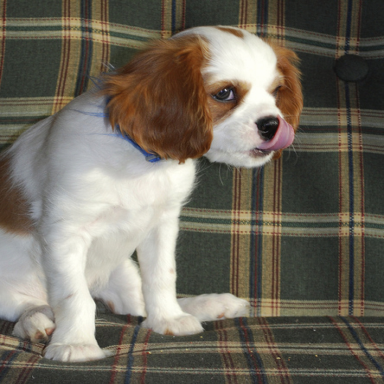} &
    \includegraphics[width=0.07\textwidth]{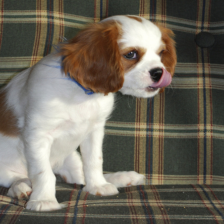} &
    \includegraphics[width=0.07\textwidth]{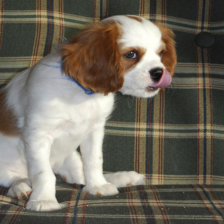} &
    \includegraphics[width=0.07\textwidth]{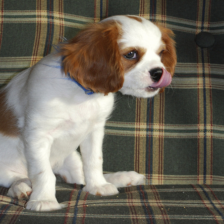} &
    \includegraphics[width=0.07\textwidth]{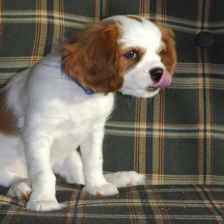} &
    \includegraphics[width=0.07\textwidth]{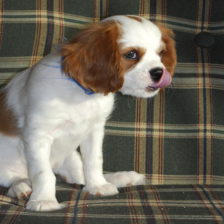} &
    \includegraphics[width=0.07\textwidth]{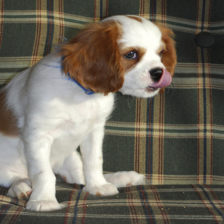} \\ \cline{0-0}
    
    \multirow{-3}{0.08\textwidth}{\subfloat[]{\label{fig:cw2_maps_a}}} &
    \includegraphics[width=0.07\textwidth]{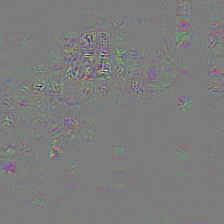} &
    \includegraphics[width=0.07\textwidth]{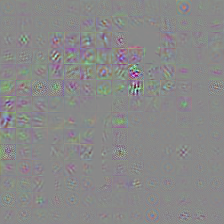} &
    \includegraphics[width=0.07\textwidth]{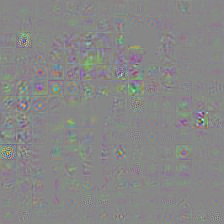} &
    \includegraphics[width=0.07\textwidth]{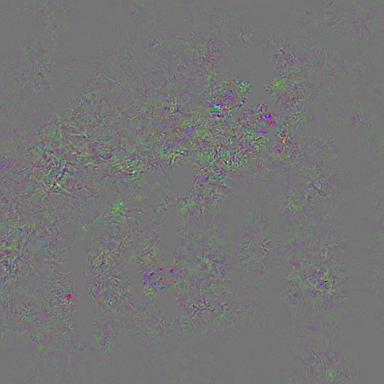} &
    \includegraphics[width=0.07\textwidth]{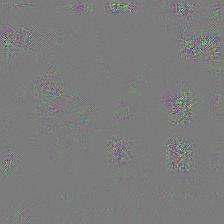} &
    \includegraphics[width=0.07\textwidth]{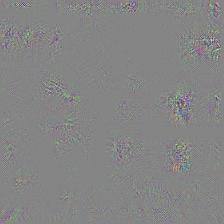} &
    \includegraphics[width=0.07\textwidth]{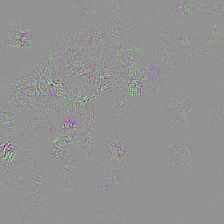} &
    \includegraphics[width=0.07\textwidth]{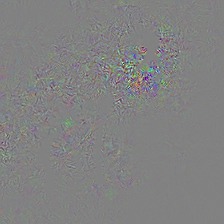} &
    \includegraphics[width=0.07\textwidth]{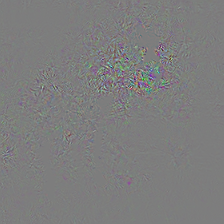} &
    \includegraphics[width=0.07\textwidth]{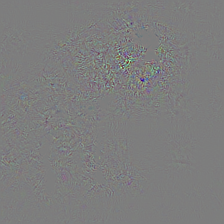} \\
    
    &
    \includegraphics[width=0.07\textwidth]{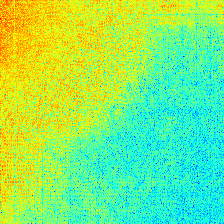} &
    \includegraphics[width=0.07\textwidth]{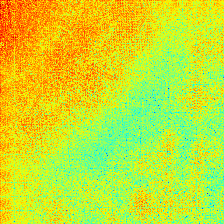} &
    \includegraphics[width=0.07\textwidth]{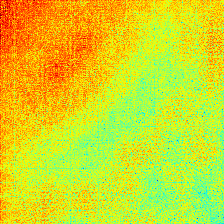} &
    \includegraphics[width=0.07\textwidth]{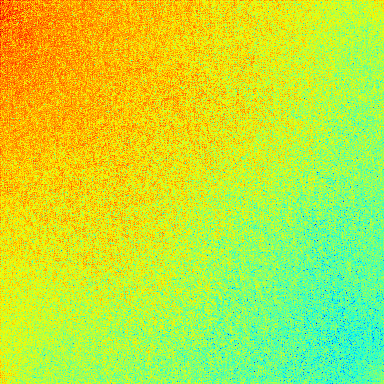} &
    \includegraphics[width=0.07\textwidth]{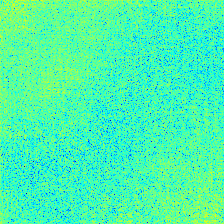} &
    \includegraphics[width=0.07\textwidth]{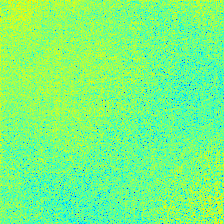} &
    \includegraphics[width=0.07\textwidth]{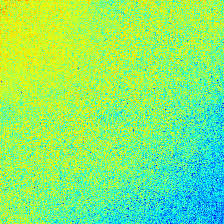} &
    \includegraphics[width=0.07\textwidth]{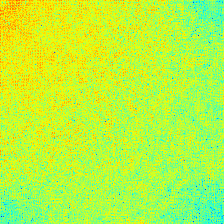} &
    \includegraphics[width=0.07\textwidth]{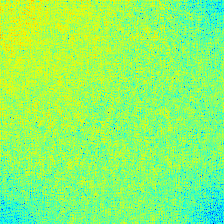} &
    \includegraphics[width=0.07\textwidth]{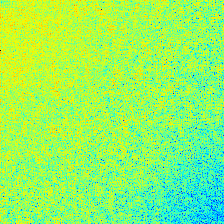} \\ \hline
    \end{tabular}
}%
    \hspace{-3mm}\caption{\protect\rule{0ex}{5ex}\begin{footnotesize}\textbf{\ac{cw}-$L_2$ attack:} The first row shows the clean sample and the \acp{ae}. The clean image is correctly classified by tested models and all \acp{ae} are successful attacks. (a) The perturbation (top) and the corresponding \acs{dct}-based spectral decomposition heatmap. Perturbation is scaled from [-1, 1] to [0, 255].\end{footnotesize}}
    \vspace{-5mm}
    \label{fig:cw2_maps}
\end{SCfigure*}

\begin{figure*}[h]
\vspace{-5mm}
\resizebox{\textwidth}{!}{%
    \setlength\tabcolsep{1.5pt}
    \tiny
    \begin{tabular}{cc}
        \begin{tabular}{l}\subfloat[]{\includegraphics[width=0.6\textwidth]{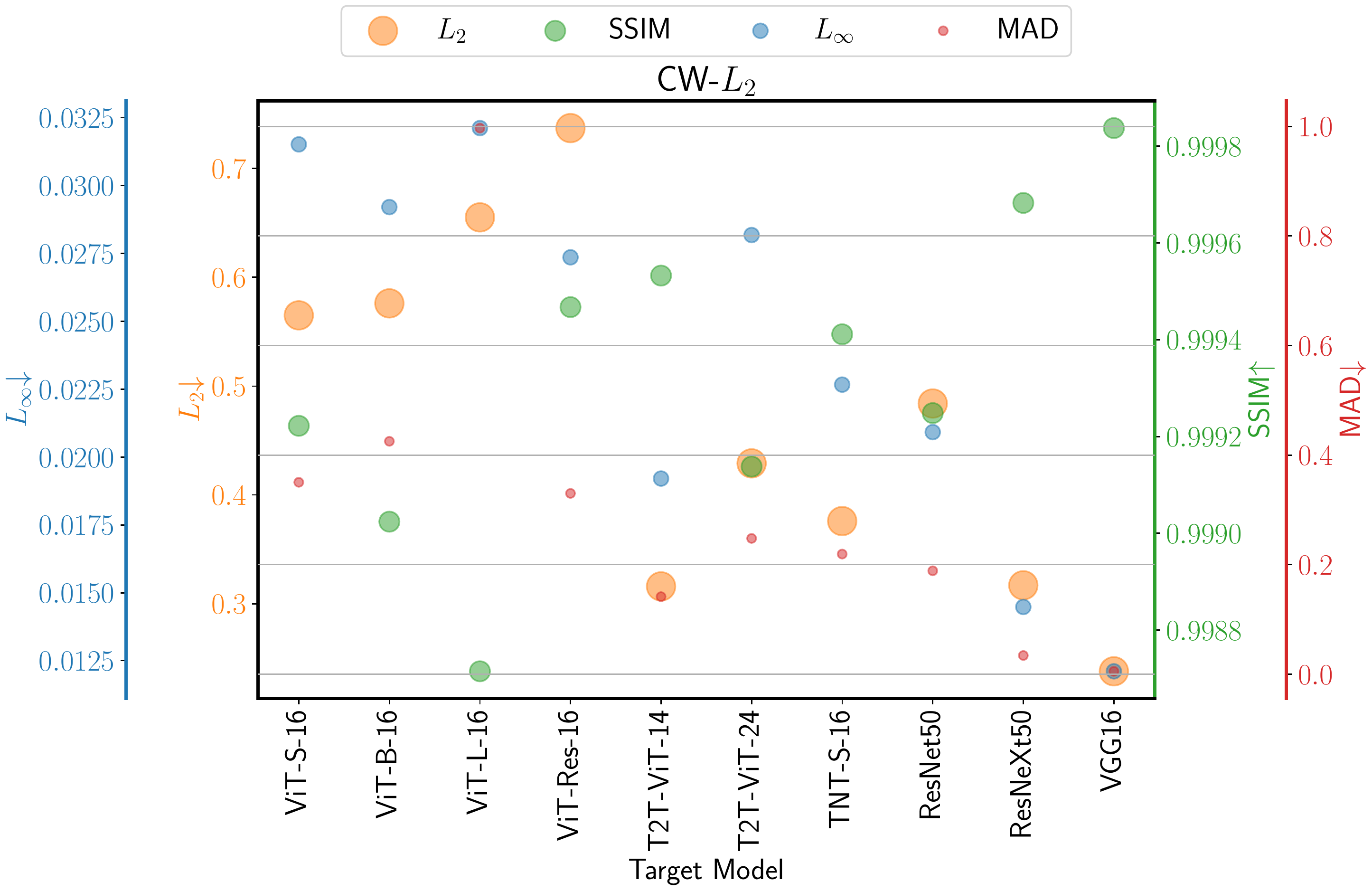}\label{fig:cw2_quality}}\end{tabular} &
        \begin{tabular}{l}\subfloat[]{\includegraphics[width=0.4\textwidth]{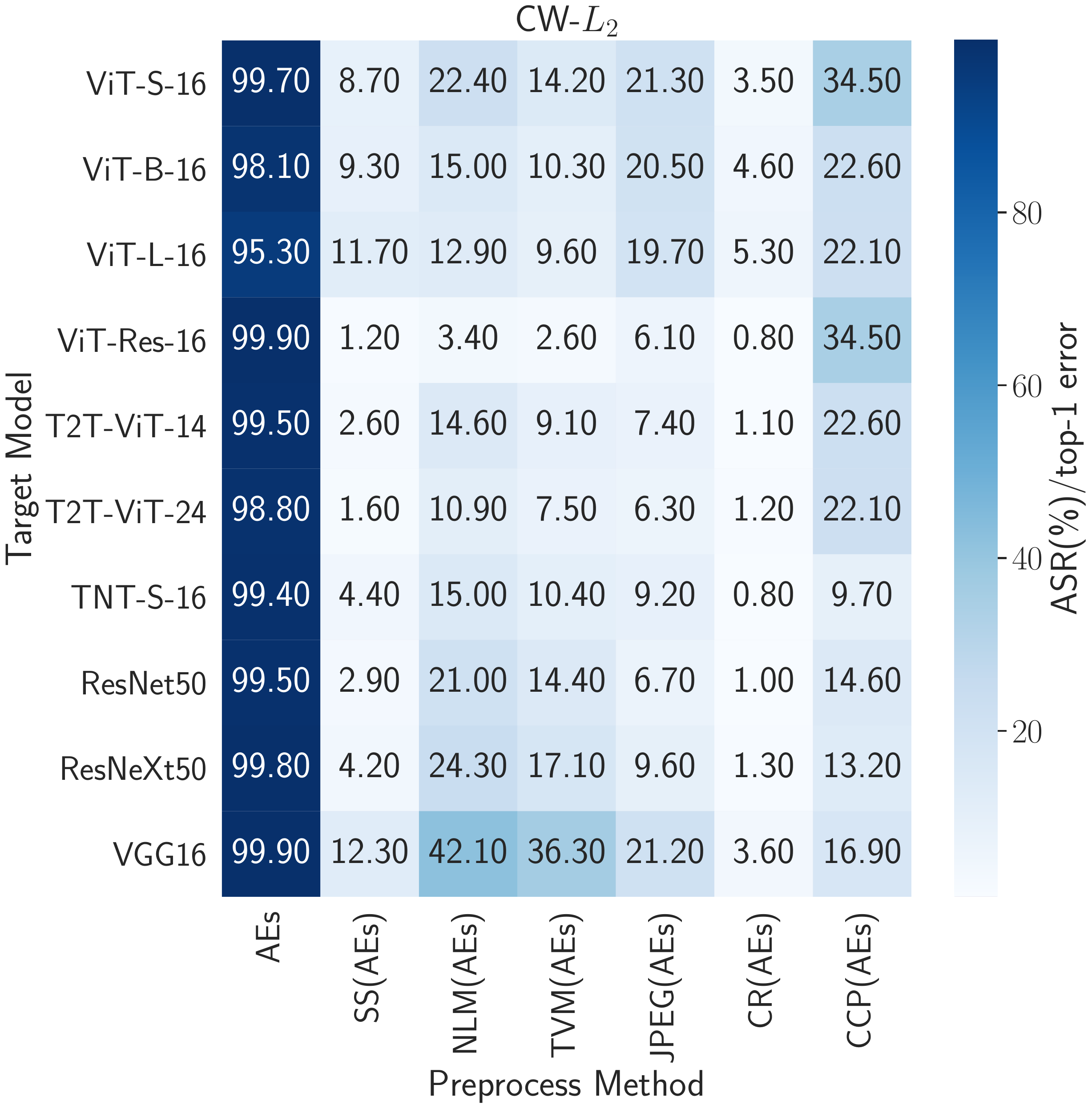}\label{fig:cw2_defense}}. \end{tabular} \\
    \end{tabular}
}%
    \vspace{-4mm}
    \hspace{-3mm}\caption{\begin{footnotesize}\textbf{\ac{cw}-$L_2$ attack:} (a) \acp{ae} quality assessment measures. (b) The \ac{asr} of the \acp{ae} and the top-1 error of the pre-processed \acp{ae} on 1000 images from ImageNet-1k. SS: \acl{ss}. NLM: \acl{nlm}. TVM: \acl{tvm}. JPEG: \acl{jpg}. CR: \acl{cr}. CCP: \acl{ccp}. \end{footnotesize}}    
    \vspace{-5mm}
    \label{fig:cw2_defense_quality}
\end{figure*}

On the other hand, in \ac{cnn}, a higher number of pixels are targeted with a high change in pixels intensities, which clearly affect the structure of the \ac{ae} as \ac{ssim} metric clearly shows. While, in hybrid-\ac{vit} models, less number of pixels with low pixel intensity change is enough to generate an \ac{ae} with high quality relative to human perception as \ac{mad} measure shows. Moreover, a narrower spread of spectrum energy has been noticed on the \ac{dct}-based decomposition, hence, bringing convolutional layers into \acp{vit} doesn't increase robustness under the $L_0$-based attacks.

\subsection{\ac{t2t}-24 and \ac{tnt}-S-16 are more robust against $L_1$-based attacks}

$L_1$-based attacks minimize the perturbation $\delta$, where $\delta = ||x-x^\prime||_1$,  $\delta\leq\epsilon$, and $f(x)\neq f(x^\prime)$, where $f(.)$ is a prediction function. In this work we consider the white-box \ac{pgd}-$L_1$ attack ~\cite{madry2017towards}, and generate 1000 \acp{ae} from ImageNet-1k validation images and a sample is shown in the first row of Figure \ref{fig:pgd_l1_400_maps}. In this experiment, we set $\epsilon$ to 400. 
As shown in the first column of Figure \ref{fig:pgd_l1_400_defense}, \ac{pgd}-$L_1$ achieves less \ac{asr} on \ac{vit}-L-16, \ac{t2t}-24 and \ac{tnt}-S-16 models. As shown in Figure \ref{fig:pgd_l1_400_maps_a}, perturbations that are generated using \ac{t2t}-24 and \ac{tnt}-S-16 models have wider frequency spectrum spread than \ac{vit}-Res-16 and \ac{vit}-L-16 models, which makes them more robust than other models. While perturbations that are generated using ResNets have a wider spread of frequency spectrum than \ac{vit}-S/B-16 and \ac{t2t}-14. From the visual quality assessment point of view, see Figure \ref{fig:pgd_l1_400_quality}, 1) as \ac{ssim} scores show, perturbations that are generated using \ac{t2t}-24 and \ac{tnt}-S-16 models have more influence to alter the adversarial image structure than the structure of \acp{ae} generated by other models, 2) as \ac{psnr} and \ac{mad} scores show, perturbations that are generated using \ac{vit} variants, in general and specifically \ac{t2t}-24 and \ac{tnt}-S-16, have higher $L_\infty$ than other tested models which yields to have lower image quality of \ac{ae}. VGG16 has the highest \ac{asr} which makes it less robust than other models. One reason for that is that VGG16 has less accuracy performance than other models which makes the learned features not robust. In this experiment, \ac{vit}-Res-16 is excluded from the analysis since the image size is different. 


\begin{SCfigure*}[1][h]
\centering
\resizebox{0.75\textwidth}{!}{%
    \setlength\tabcolsep{1.5pt}
    \tiny
    \begin{tabular}{ccccccccccc}
    \hline
    Clean &
    ViT-S-16 & 
    ViT-B-16 &
    ViT-L-16 &
    ViT-Res-16 &
    T2T-ViT-14 &
    T2T-ViT-24 &
    TNT-S-16 &
    ResNet50 &
    ResNeXt50 &
    VGG16 \\ \small
    
    \includegraphics[width=0.07\textwidth]{imgs/clean_imagenet-1k_87.png} &
    \includegraphics[width=0.07\textwidth]{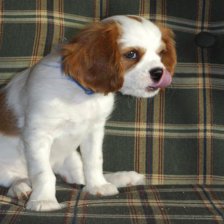} &
    \includegraphics[width=0.07\textwidth]{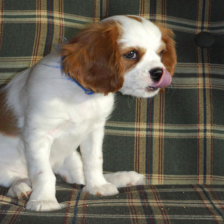} &
    \includegraphics[width=0.07\textwidth]{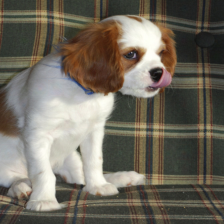} &
    \includegraphics[width=0.07\textwidth]{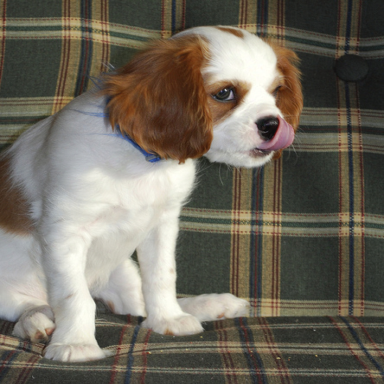} &
    \includegraphics[width=0.07\textwidth]{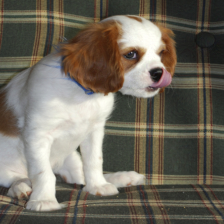} &
    \includegraphics[width=0.07\textwidth]{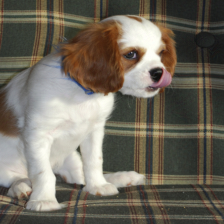} &
    \includegraphics[width=0.07\textwidth]{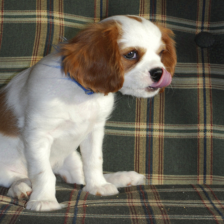} &
    \includegraphics[width=0.07\textwidth]{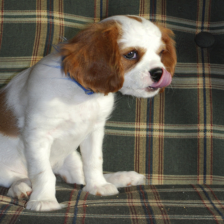} &
    \includegraphics[width=0.07\textwidth]{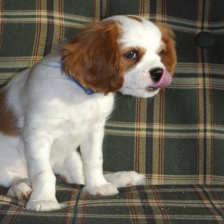} &
    \includegraphics[width=0.07\textwidth]{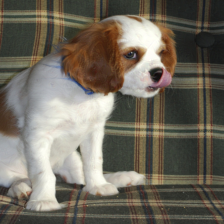} \\ \cline{0-0}
    
    \multirow{-3}{0.07\textwidth}{\subfloat[]{\label{fig:pgd_l2_2_maps_a}}} &
    \includegraphics[width=0.07\textwidth]{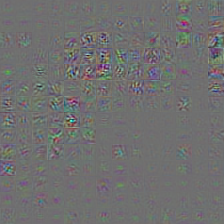} &
    \includegraphics[width=0.07\textwidth]{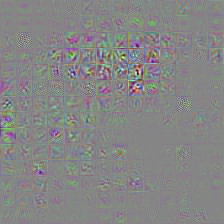} &
    \includegraphics[width=0.07\textwidth]{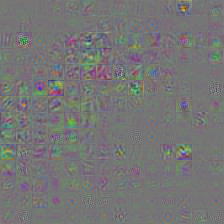} &
    \includegraphics[width=0.07\textwidth]{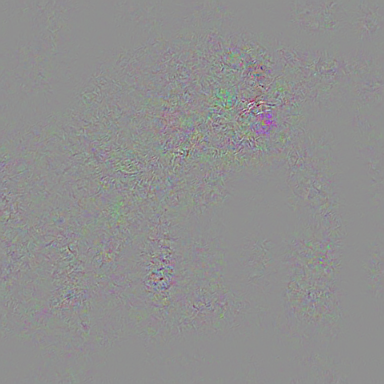} &
    \includegraphics[width=0.07\textwidth]{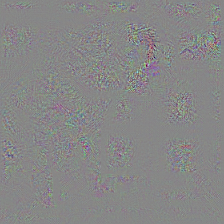} &
    \includegraphics[width=0.07\textwidth]{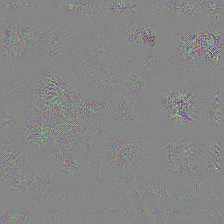} &
    \includegraphics[width=0.07\textwidth]{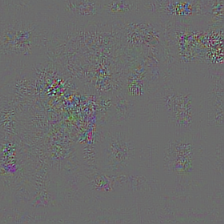} &
    \includegraphics[width=0.07\textwidth]{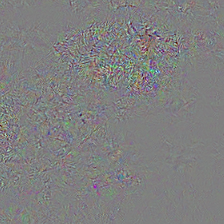} &
    \includegraphics[width=0.07\textwidth]{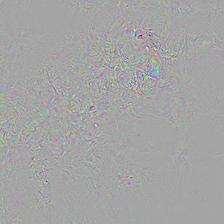} &
    \includegraphics[width=0.07\textwidth]{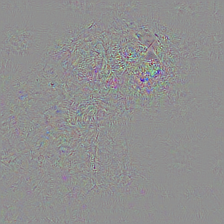} \\
    
    &
    \includegraphics[width=0.07\textwidth]{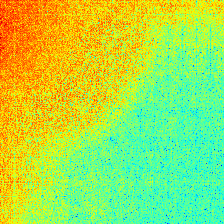} &
    \includegraphics[width=0.07\textwidth]{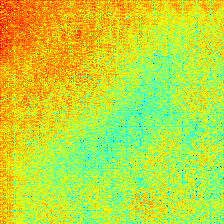} &
    \includegraphics[width=0.07\textwidth]{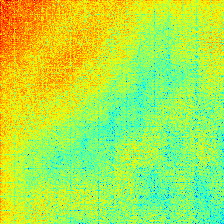} &
    \includegraphics[width=0.07\textwidth]{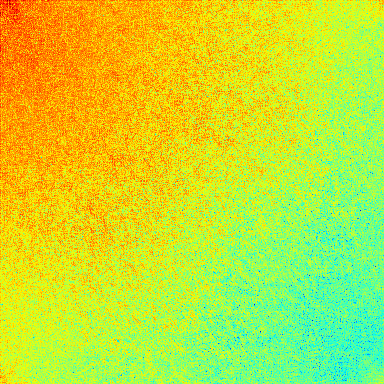} &
    \includegraphics[width=0.07\textwidth]{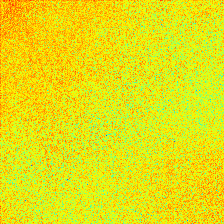} &
    \includegraphics[width=0.07\textwidth]{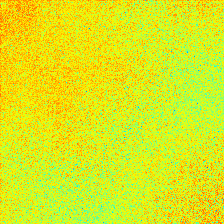} &
    \includegraphics[width=0.07\textwidth]{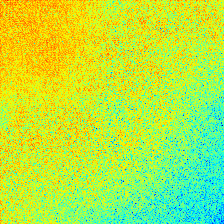} &
    \includegraphics[width=0.07\textwidth]{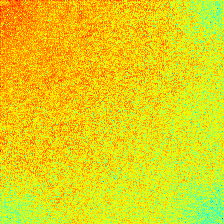} &
    \includegraphics[width=0.07\textwidth]{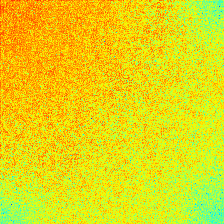} &
    \includegraphics[width=0.07\textwidth]{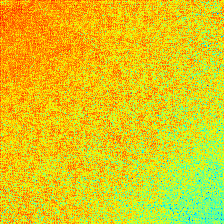} \\ \hline
    \end{tabular}
}%
    \hspace{-3mm}\caption{\protect\rule{0ex}{5ex}\begin{footnotesize}\textbf{\ac{pgd}-$L_2$ $\epsilon=2$ attack:} The first row shows the clean sample and the \acp{ae}. The clean image is correctly classified by tested models and all \acp{ae} are successful attacks. (a) The perturbation (top) and the corresponding \acs{dct}-based spectral decomposition heatmap. Perturbation is scaled from [-1, 1] to [0, 255].\end{footnotesize}}
    \label{fig:pgd_l2_2_maps}
    \vspace{-5mm}
\end{SCfigure*}


\begin{figure*}[h]
\vspace{-5mm}
\resizebox{\textwidth}{!}{%
    \setlength\tabcolsep{1.5pt}
    \tiny
    \begin{tabular}{cc}
        \begin{tabular}{l}\subfloat[]{\includegraphics[width=0.6\textwidth]{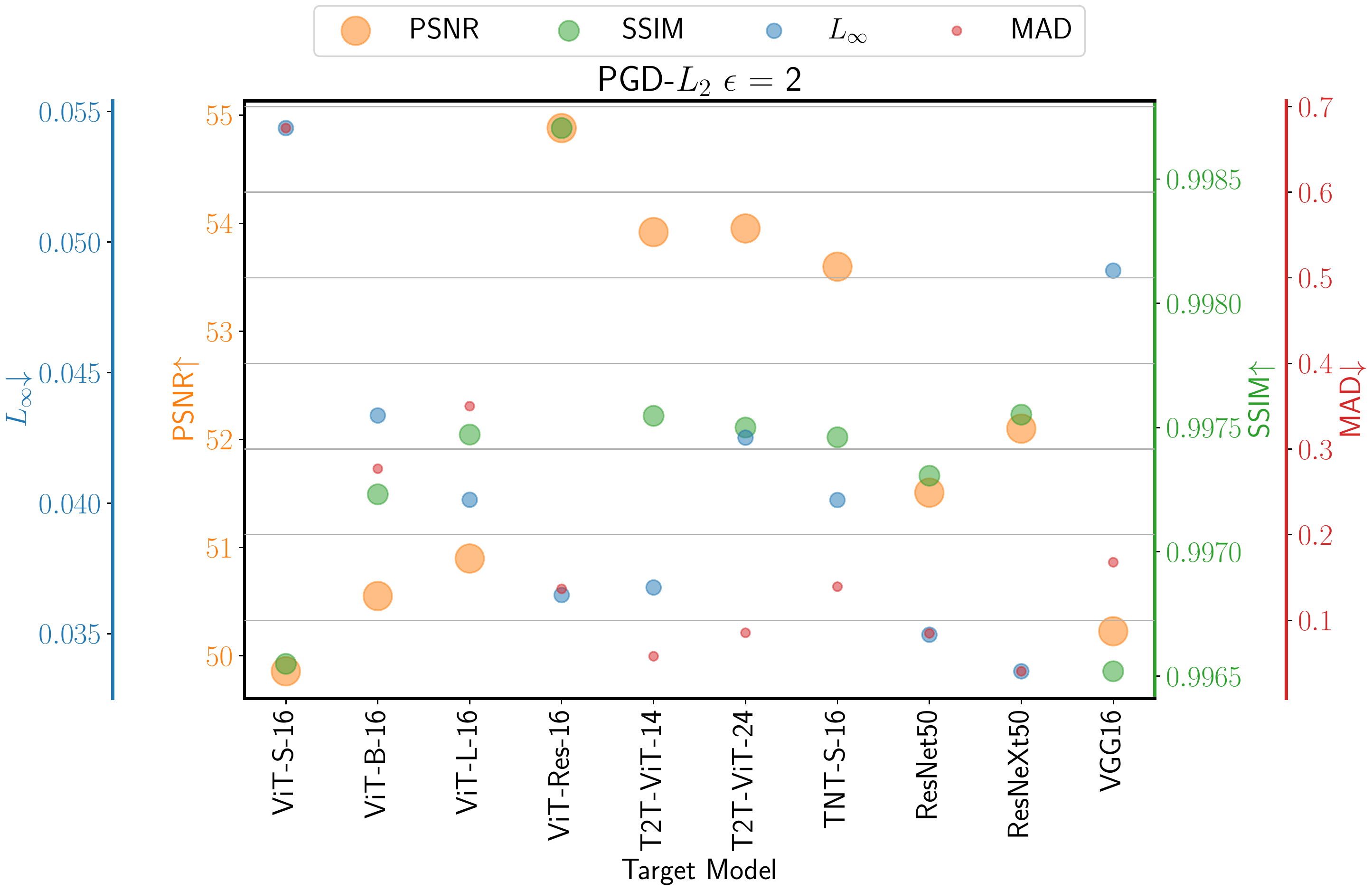}\label{fig:pgd_l2_2_quality}}\end{tabular} &
        \begin{tabular}{l}\subfloat[]{\includegraphics[width=0.4\textwidth]{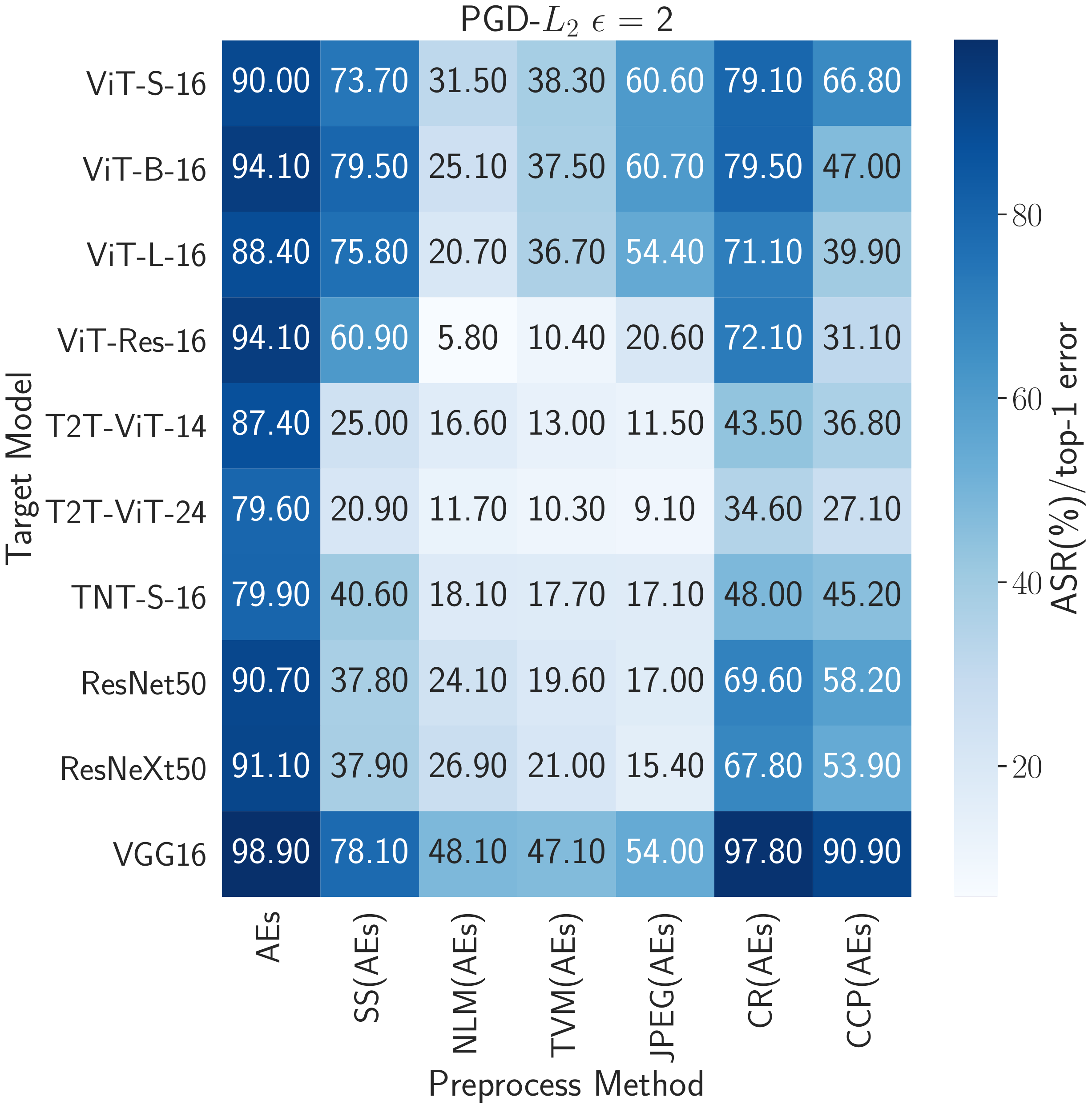}\label{fig:pgd_l2_2_defense}} \end{tabular} \\
    \end{tabular}
}%
    \vspace{-8mm}
    \hspace{-3mm}\caption{\protect\rule{0ex}{5ex}\begin{footnotesize}\textbf{\ac{pgd}-$L_2$ $\epsilon=2$  attack:} (a) \acp{ae} quality assessment measures. (b) The \ac{asr} of the \acp{ae} and  the top-1 error of the pre-processed \acp{ae}  on 1000 images from imagenet-1k. SS: \acl{ss}. NLM: \acl{nlm}. TVM: \acl{tvm}. JPEG: \acl{jpg}. CR: \acl{cr}. CCP: \acl{ccp}. \end{footnotesize}}    
    \vspace{-5mm}
    \label{fig:pgd_l2_2_defense_quality}
\end{figure*}

\subsection{Vanilla \acp{vit} are more robust under \ac{cw}-$L_2$, while hybrid-\acp{vit} are more robust under \ac{pgd}-$L_2$ attacks} \label{sec:l2_robust}
$L_2$-based attacks minimize the perturbation $\delta$, where $\delta = ||x-x^\prime||_2$,  $\delta\leq\epsilon$, and $f(x)\neq f(x^\prime)$. In this work we consider the white-box \ac{cw}-$L_2$ \cite{carlini2017towards} and \ac{pgd}-$L_2$ attacks ~\cite{madry2017towards}, and generate 1000 \acp{ae} from ImageNet-1k validation images and a sample is shown in the first row of Figure \ref{fig:cw2_maps}. 

\subsubsection{Robustness under \ac{cw}-$L_2$ attacks}
As shown in the the first column of Figure \ref{fig:cw2_defense}, \ac{cw}-$L_2$ attack achieves comparable \ac{asr} on all tested models except for \ac{vit}-L-16 which achieves less \ac{asr}. Compared to ResNet, the \ac{dct} decomposition of the perturbation shows the huge wide spread of energy spectrum for vanilla \acp{vit}, especially \ac{vit}-L-16. Moreover, the algorithm of \ac{cw}-$L_2$ has to increase $L_2$ and $L_\infty$ distortions, see Figure \ref{fig:cw2_quality}, for \acp{vit} in order to generate successful attacks. Hence, \ac{ssim} and \ac{mad} show less quality score for \acp{ae} that are generated using vanilla \acp{vit}. ResNet shows robustness over \ac{t2t}-14 since the perturbations that are generated using ResNet have a wider spread of frequency spectrum than \ac{t2t}-14 but not wider than \ac{t2t}-24 and \ac{tnt}-S-16. The visual quality scores, given in Figure \ref{fig:cw2_quality}, confirm that ResNet is more robust than  \ac{t2t}-14 but less robust than \ac{t2t}-24 and \ac{tnt}-S-16.

\subsubsection{Robustness under \ac{pgd}-$L_2$ attacks}
Figure \ref{fig:pgd_l2_2_maps} shows an example to study the target models' robustness against \ac{pgd}-$L_2$ attack with $\epsilon=2$. 
From the first column of Figure \ref{fig:pgd_l2_2_defense}, we concluded that the hybrid-\ac{vit} are more robust than other models against \ac{pgd}-$L_2$ attacks. The \ac{asr} of the hybrid-\acp{vit} is less than the \ac{asr} of the other target models. Moreover, when we look at Figure \ref{fig:pgd_l2_2_maps_a}, we can see that hybrid-\acp{vit} have a wider spread of discriminative features on all frequencies than other models. Moreover, the \ac{vit}-S/B/L-16 models show better robustness over ResNet. When looking at Figure \ref{fig:pgd_l2_2_quality}, we can see that \ac{vit} variants have higher $L_\infty$ score, especially the \ac{vit}-S-16, than ResNet which affects the \ac{ae} structure and visual quality as \ac{ssim} and \ac{mad} scores show. VGG16 has the highest \ac{asr} and high distortion which makes it less robust. One reason for that is that VGG16 has less accuracy performance than other models which makes the learned features not robust. In this experiment, \ac{vit}-Res-16 is excluded from the analysis since the image size is different. 


Hence, we can conclude that neither larger model's architecture, like \ac{vit}-L, nor bringing convolutional modules for tokenization in \acp{vit}, like \ac{tnt}, will necessarily enhance the robustness.

\subsection{Hybrid-\acp{vit} and small \ac{vit} are matters under some $L_\infty$-based attacks.} \label{sec:li_robust}
$L_\infty$-based attacks minimize the perturbation $\delta$, where $\delta = ||x-x^\prime||_\infty$, $\delta\leq\epsilon$, and $f(x) \neq f(x^\prime)$. In this work we consider the white-box \ac{fgsm}-$L_\infty$~\cite{goodfellow2014explaining}, \ac{cw}-$L_\infty$ \cite{carlini2017towards}, \ac{pgd}-$L_\infty$ ~\cite{madry2017towards}, and \ac{uap} \cite{moosavi2017universal} attacks. Moreover, two black box attacks are considered; \ac{rays} \cite{chen2020rays} and \ac{sa} \cite{andriushchenko2020square} attacks and the \ac{aa}-$L_\infty$ \cite{croce2020reliable} hybrid attack is considered as well. 


\begin{SCfigure*}[1][h]
\centering

\resizebox{0.75\textwidth}{!}{%
    \setlength\tabcolsep{1.5pt}
    \tiny
    \begin{tabular}{ccccccccccc}
    
    \hline
    Clean &
    ViT-S-16 & 
    ViT-B-16 &
    ViT-L-16 &
    ViT-Res-16 &
    T2T-ViT-14 &
    T2T-ViT-24 &
    TNT-S-16 &
    ResNet50 &
    ResNeXt50 &
    VGG16 \\ \Large
    
    \includegraphics[width=0.07\textwidth]{imgs/clean_imagenet-1k_87.png} &
    \includegraphics[width=0.07\textwidth]{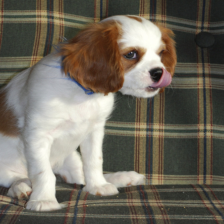} &
    \includegraphics[width=0.07\textwidth]{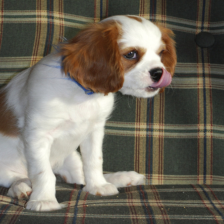} &
    \includegraphics[width=0.07\textwidth]{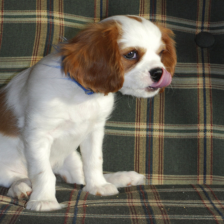} &
    \includegraphics[width=0.07\textwidth]{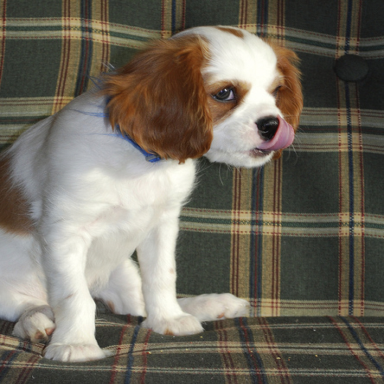} &
    \includegraphics[width=0.07\textwidth]{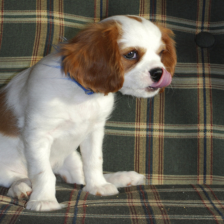} &
    \includegraphics[width=0.07\textwidth]{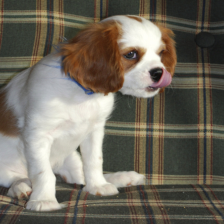} &
    \includegraphics[width=0.07\textwidth]{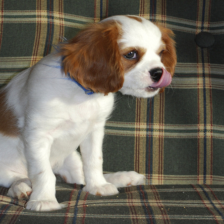} &
    \includegraphics[width=0.07\textwidth]{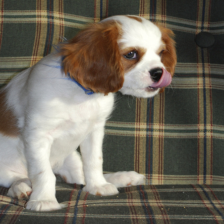} &
    \includegraphics[width=0.07\textwidth]{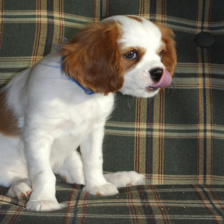} &
    \includegraphics[width=0.07\textwidth]{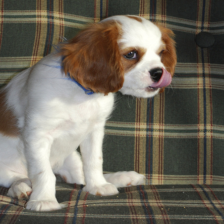} \\ \cline{0-0}
    
    \multirow{-3}{0.08\textwidth}{\subfloat[]{\label{fig:pgd_li_0.003_maps_a}}} &
    \includegraphics[width=0.07\textwidth]{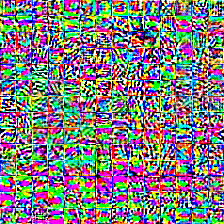} &
    \includegraphics[width=0.07\textwidth]{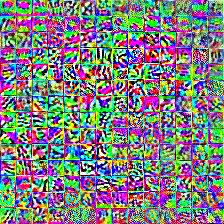} &
    \includegraphics[width=0.07\textwidth]{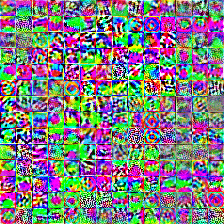} &
    \includegraphics[width=0.07\textwidth]{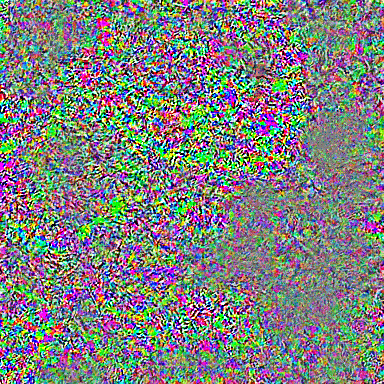} &
    \includegraphics[width=0.07\textwidth]{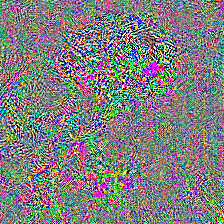} &
    \includegraphics[width=0.07\textwidth]{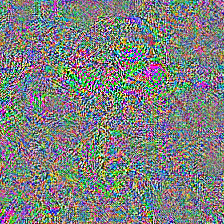} &
    \includegraphics[width=0.07\textwidth]{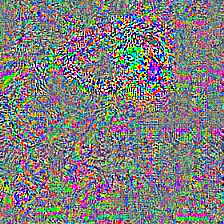} &
    \includegraphics[width=0.07\textwidth]{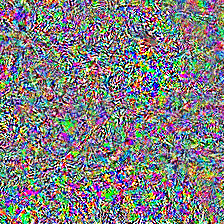} &
    \includegraphics[width=0.07\textwidth]{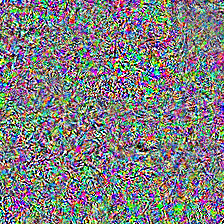} &
    \includegraphics[width=0.07\textwidth]{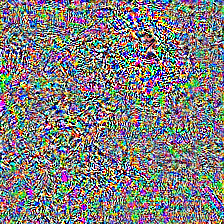} \\
    
    &
    \includegraphics[width=0.07\textwidth]{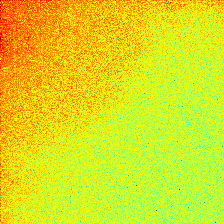} &
    \includegraphics[width=0.07\textwidth]{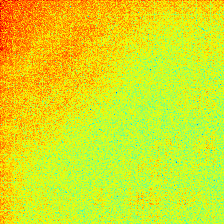} &
    \includegraphics[width=0.07\textwidth]{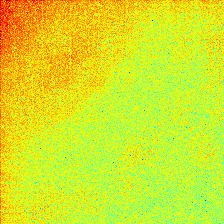} &
    \includegraphics[width=0.07\textwidth]{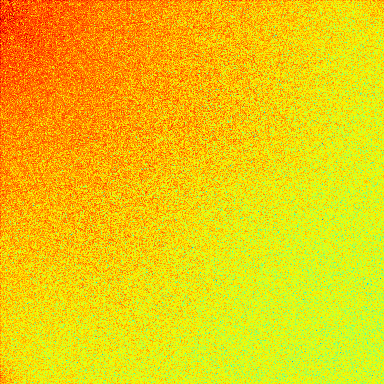} &
    \includegraphics[width=0.07\textwidth]{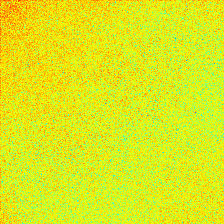} &
    \includegraphics[width=0.07\textwidth]{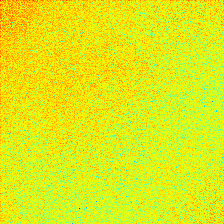} &
    \includegraphics[width=0.07\textwidth]{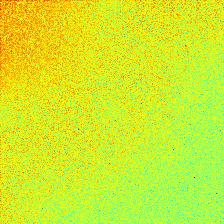} &
    \includegraphics[width=0.07\textwidth]{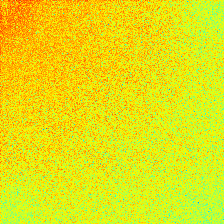} &
    \includegraphics[width=0.07\textwidth]{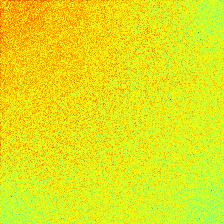} &
    \includegraphics[width=0.07\textwidth]{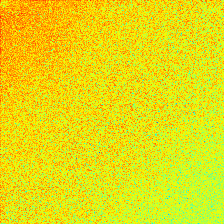} \\ \hline
    \end{tabular}
}%
    \hspace{-3mm}\caption{\begin{footnotesize}\textbf{\ac{pgd}-$L_\infty$ $\epsilon=1/255$ attack:} First row shows the clean sample and the \acp{ae}. The clean image is correctly classified by tested models and all \acp{ae} are successful attacks. (a) The perturbation (top) and the corresponding \acs{dct}-based spectral decomposition heatmap. Perturbation is scaled from [-1, 1] to [0, 255].\end{footnotesize}}
    \vspace{-5mm}
    \label{fig:pgd_li_0.003_maps}
\end{SCfigure*}

\begin{figure*}
\vspace{-3mm}
\resizebox{\textwidth}{!}{%
    \setlength\tabcolsep{1.5pt}
    \tiny
    \begin{tabular}{cc}
        \begin{tabular}{l}\subfloat[]{\includegraphics[width=0.6\textwidth]{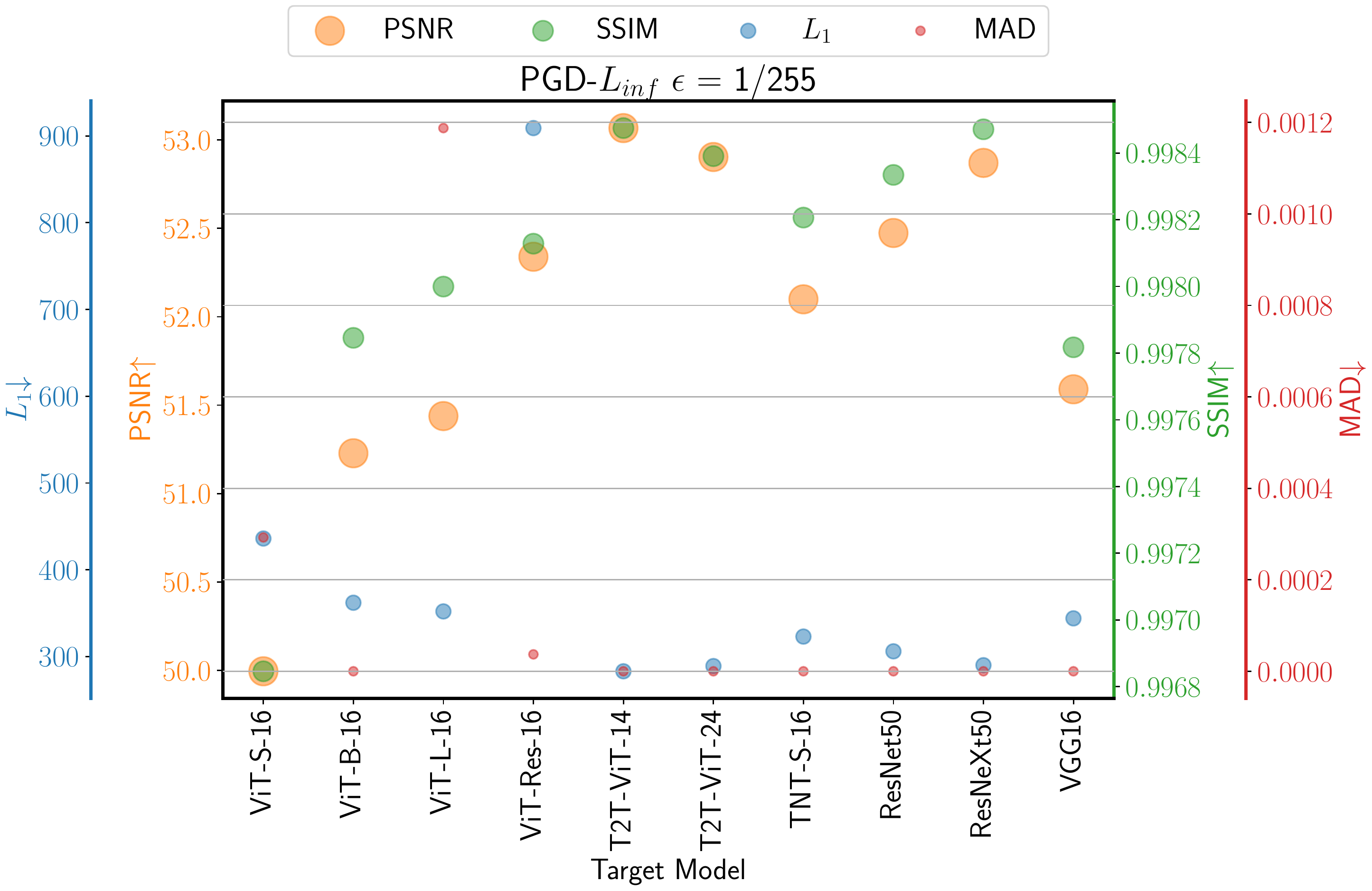}\label{fig:pgd_li_0.003_quality}}\end{tabular} &
        \begin{tabular}{l}\subfloat[]{\includegraphics[width=0.4\textwidth]{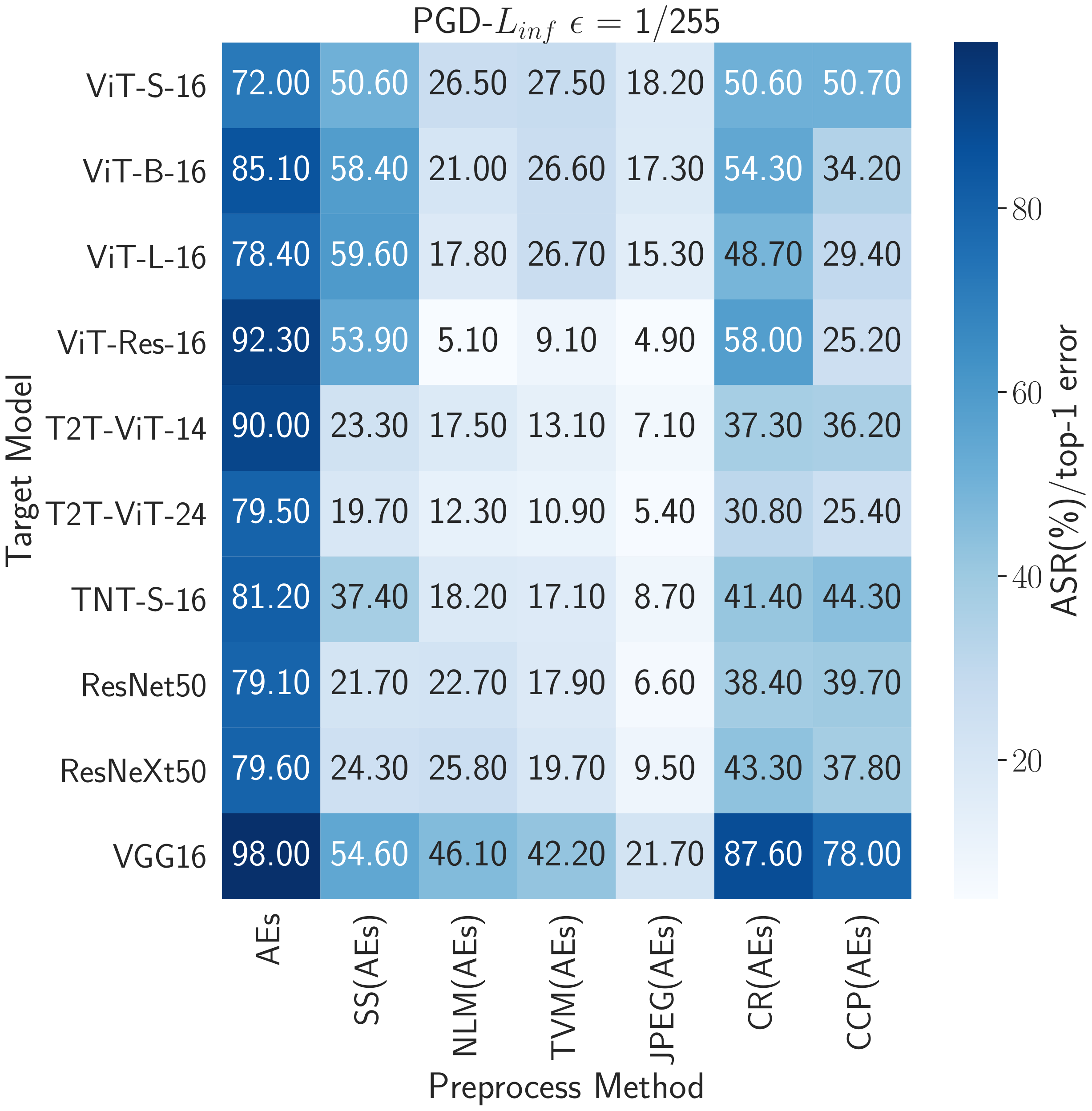}\label{fig:pgd_li_0.003_defense}.} \end{tabular} \\
    \end{tabular}
}%
\vspace{-8mm}
    \hspace{-3mm}\caption{\protect\rule{0ex}{5ex}\begin{footnotesize}\textbf{\ac{pgd}-$L_\infty$ $\epsilon=1/255$ attack:} (a) \acp{ae} quality assessment measures. (b) The \ac{asr} of the \acp{ae} and the top-1 error of the pre-processed \acp{ae} on 1000 images from imagenet-1k. SS: \acl{ss}. NLM: \acl{nlm}. TVM: \acl{tvm}. JPEG: \acl{jpg}. CR: \acl{cr}. CCP: \acl{ccp}.\end{footnotesize}}
    \vspace{-5mm}
    \label{fig:pgd_li_0.003_defense_quality}
\end{figure*}

\subsubsection{Robustness under \ac{pgd}-$L_\infty$, \ac{aa}, and \ac{uap} attacks}
We generate 1000 \acp{ae} from ImageNet-1k validation images and a sample is shown in the first row of Figure \ref{fig:pgd_li_0.003_maps}. \ac{pgd}-$L_\infty$ attack achieves less \ac{asr} on \ac{vit}-S-16 than other tested models. While ResNets have comparable robustness to \ac{vit}-L-16, \ac{t2t}-24, and \ac{tnt}-S-16, and have better robustness over \ac{vit}-B-16, \ac{vit}-Res-16, and \ac{t2t}-14. The perturbations that are generated using \ac{vit}-S-16 have the following properties: 1) more spread of energy spectrum as \ac{dct} decomposition shows in Figure~\ref{fig:pgd_li_0.003_maps_a} , 2) lower visual quality as \ac{psnr}, \ac{ssim} scores show in Figure \ref{fig:pgd_li_0.003_quality}, and 3) higher $L_1$ score. $L_1$ score for \ac{vit}-Res-16 is not considered in the analysis since the image size is different. Similar to $L_1$-based attack, VGG16 has the highest \ac{asr} which makes it less robust. VGG16 has the highest \ac{asr} and high distortion which makes it less robust than other target tested models. 


Under the \ac{aa} and \ac{uap} attacks, we noticed that \ac{vit}-S-16 has better robustness than other target models, as Figure \ref{fig:asr_vs_mad_autoattack} shows for \ac{aa}, but due to space limitation, we only visualized \ac{pgd}-$L_\infty$ attacks.


\subsubsection{Robustness under \ac{fgsm}-$L_\infty$ attacks}
\begin{SCfigure*}[1][h]
\centering
\resizebox{0.75\textwidth}{!}{%
    \setlength\tabcolsep{1.5pt}
    \tiny
    \begin{tabular}{ccccccccccc}
    \hline
    Clean &
    ViT-S-16 & 
    ViT-B-16 &
    ViT-L-16 &
    ViT-Res-16 &
    T2T-ViT-14 &
    T2T-ViT-24 &
    TNT-S-16 &
    ResNet50 &
    ResNeXt50 &
    VGG16 \\ \small
    
    \includegraphics[width=0.07\textwidth]{imgs/clean_imagenet-1k_81.png} &
    \includegraphics[width=0.07\textwidth]{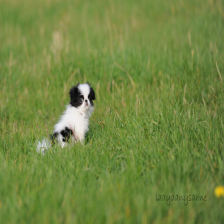} &
    \includegraphics[width=0.07\textwidth]{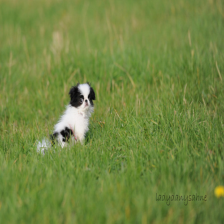} &
    \includegraphics[width=0.07\textwidth]{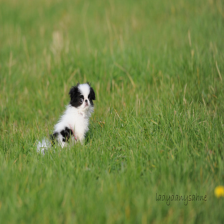} &
    \includegraphics[width=0.07\textwidth]{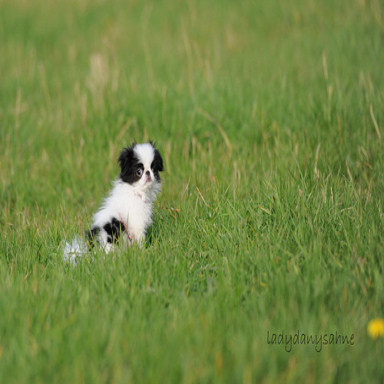} &
    \includegraphics[width=0.07\textwidth]{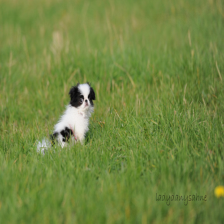} &
    \includegraphics[width=0.07\textwidth]{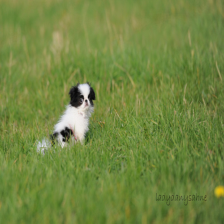} &
    \includegraphics[width=0.07\textwidth]{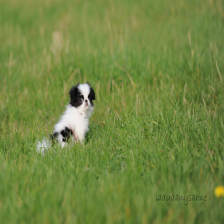} &
    \includegraphics[width=0.07\textwidth]{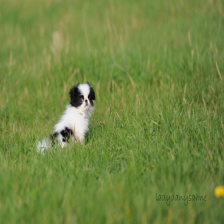} &
    \includegraphics[width=0.07\textwidth]{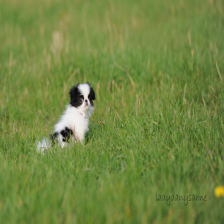} &
    \includegraphics[width=0.07\textwidth]{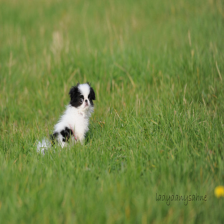} \\ \cline{0-0}
    
    \multirow{-3}{0.07\textwidth}{\subfloat[]{\label{fig:fgsm_li_0.003_maps_a}}} &
    \includegraphics[width=0.07\textwidth]{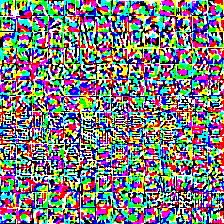} &
    \includegraphics[width=0.07\textwidth]{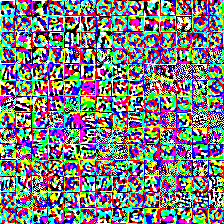} &
    \includegraphics[width=0.07\textwidth]{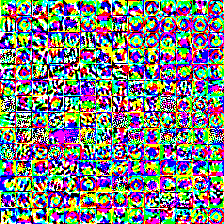} &
    \includegraphics[width=0.07\textwidth]{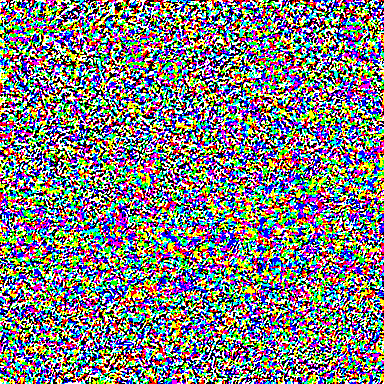} &
    \includegraphics[width=0.07\textwidth]{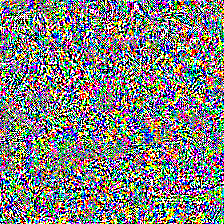} &
    \includegraphics[width=0.07\textwidth]{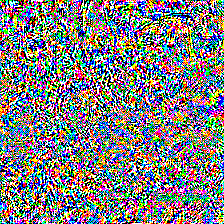} &
    \includegraphics[width=0.07\textwidth]{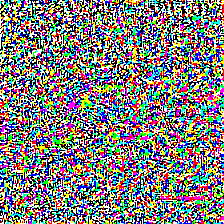} &
    \includegraphics[width=0.07\textwidth]{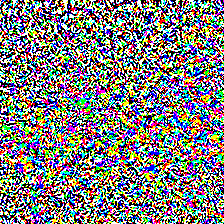} &
    \includegraphics[width=0.07\textwidth]{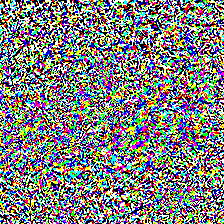} &
    \includegraphics[width=0.07\textwidth]{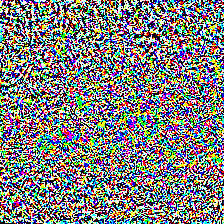} \\
    
    &
    \includegraphics[width=0.07\textwidth]{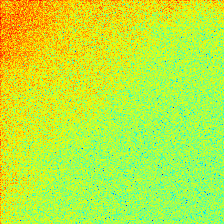} &
    \includegraphics[width=0.07\textwidth]{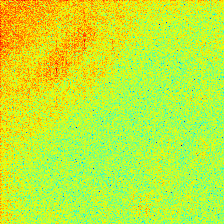} &
    \includegraphics[width=0.07\textwidth]{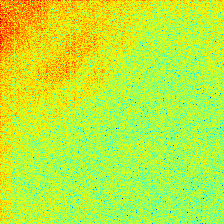} &
    \includegraphics[width=0.07\textwidth]{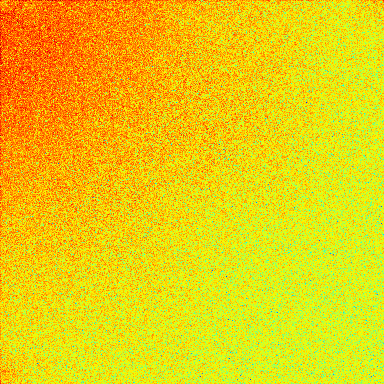} &
    \includegraphics[width=0.07\textwidth]{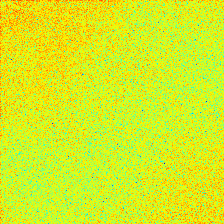} &
    \includegraphics[width=0.07\textwidth]{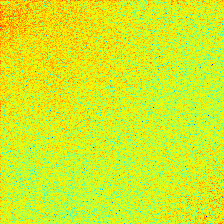} &
    \includegraphics[width=0.07\textwidth]{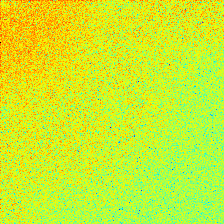} &
    \includegraphics[width=0.07\textwidth]{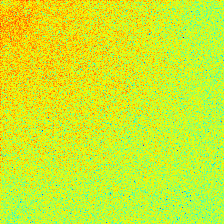} &
    \includegraphics[width=0.07\textwidth]{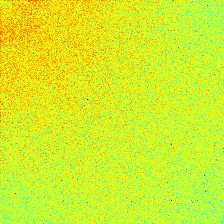} &
    \includegraphics[width=0.07\textwidth]{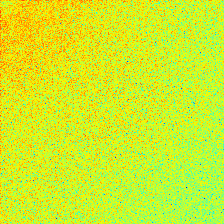} \\ \hline
    \end{tabular}
}%
    \hspace{-3mm}\caption{\protect\rule{0ex}{5ex}\begin{footnotesize}\textbf{\ac{fgsm}-$L_\infty$ $\epsilon=1/255$ attack:} The first row shows the clean sample and the \acp{ae}. The clean image is correctly classified by tested models and all \acp{ae} are successful attacks. (a) The perturbation (top) and the corresponding \acs{dct}-based spectral decomposition heatmap. Perturbation is scaled from [-1, 1] to [0, 255].\end{footnotesize}}
    \label{fig:fgsm_li_0.003_maps}
    \vspace{-4mm}
\end{SCfigure*}

\begin{figure*}
\vspace{-6mm}
\resizebox{\textwidth}{!}{%
    \setlength\tabcolsep{1.5pt}
    \tiny
    \begin{tabular}{cc}
        \begin{tabular}{l}\subfloat[]{\includegraphics[width=0.6\textwidth]{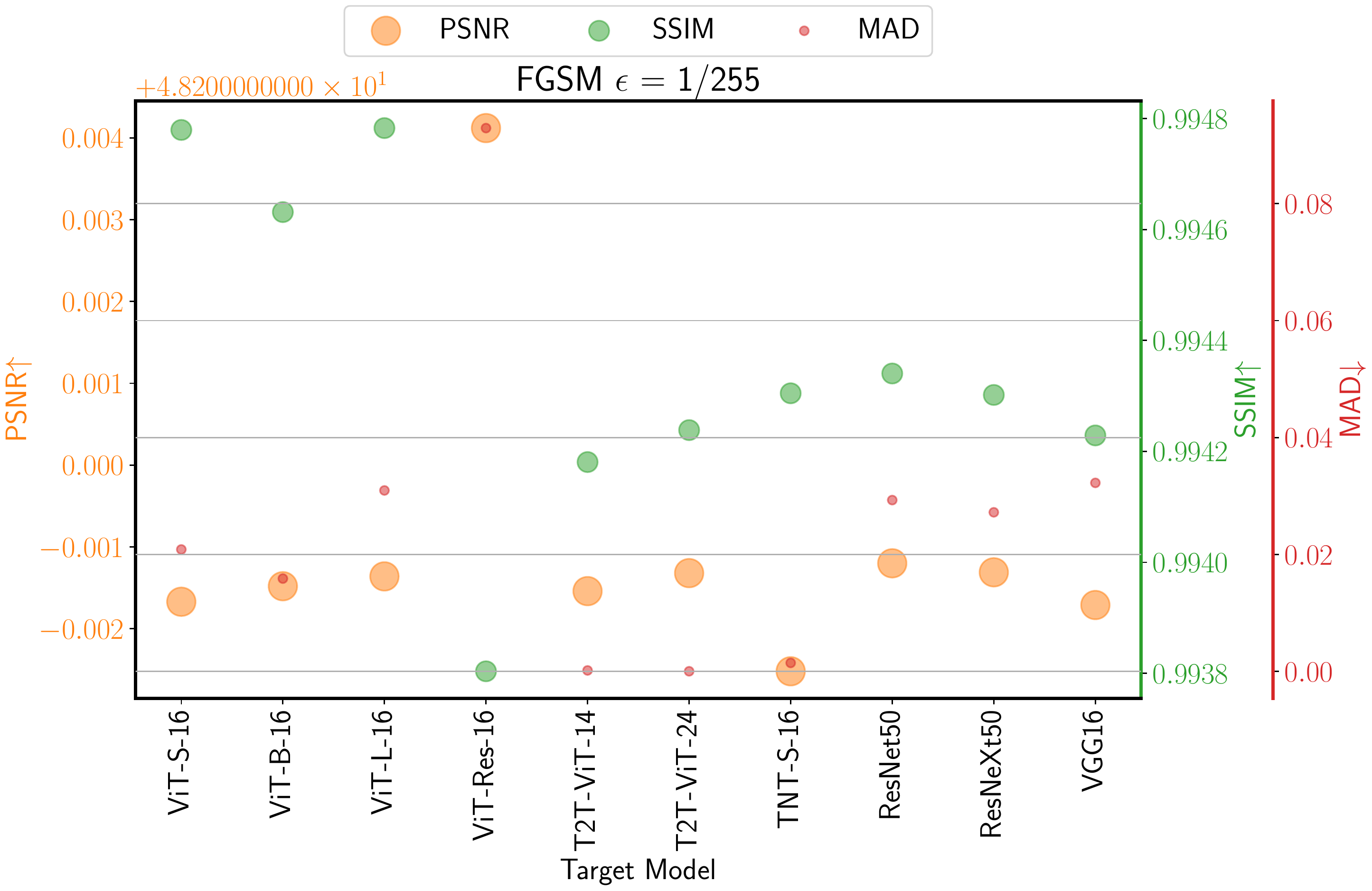}\label{fig:fgsm_li_0.003_quality}}\end{tabular} &
        \begin{tabular}{l}\subfloat[]{\includegraphics[width=0.4\textwidth]{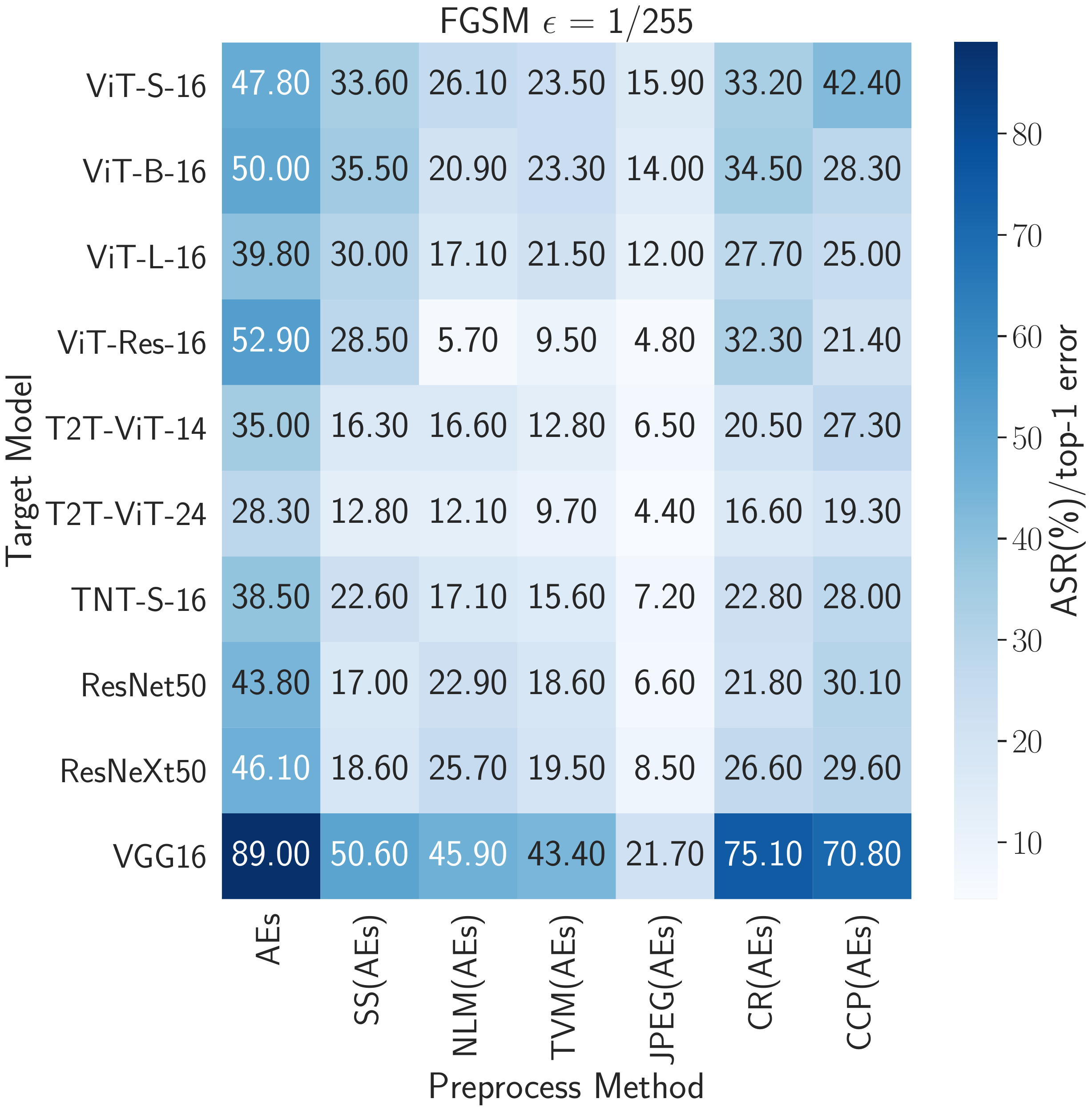}\label{fig:fgsm_li_0.003_defense}} \end{tabular} \\
    \end{tabular}
}%
\vspace{-8mm}
    \hspace{-3mm}\caption{\protect\rule{0ex}{5ex}\begin{footnotesize}\textbf{\ac{fgsm}-$L_\infty$ $\epsilon=1/255$  attack:} (a) \acp{ae} quality assessment measures. (b) The \ac{asr} of the \acp{ae} and  the top-1 error of the pre-processed \acp{ae}  on 1000 images from imagenet-1k. SS: \acl{ss}. NLM: \acl{nlm}. TVM: \acl{tvm}. JPEG: \acl{jpg}. CR: \acl{cr}. CCP: \acl{ccp}. \end{footnotesize}}
    \vspace{-4mm}
    \label{fig:fgsm_li_0.003_defense_quality}
\end{figure*}

Figure \ref{fig:fgsm_li_0.003_maps} shows an example to study the target models' robustness against \ac{fgsm}-$L_\infty$ attack. Figure \ref{fig:fgsm_li_0.003_quality}, \ref{fig:cwi_quality} shows the visual quality scores of the \acp{ae} that are generated using \ac{fgsm}-$L_\infty$ attack. While Figure \ref{fig:fgsm_li_0.003_defense} shows the \ac{asr} of the \acp{ae} and  the top-1 error of the pre-processed \acp{ae} that are generated using \ac{fgsm}-$L_\infty$ attack.

Figure \ref{fig:fgsm_li_0.003_defense} shows that hybrid-\acp{vit} have lower \ac{asr} than other models. Vanilla \ac{vit}-L-16 shows more robustness over ResNet and ReNeXt, while ResNet shows robustness over vanilla \ac{vit}-S/B-16. The wider spread of the energy spectrum of the \ac{dct} decomposition confirms the robustness of hybrid-\acp{vit} over other models, as shown in Figure \ref{fig:fgsm_li_0.003_maps_a}. Moreover, \ac{ssim} scores, illustrated in Figure \ref{fig:fgsm_li_0.003_quality}, show that hybrid-\ac{vit} has lower score than other models, except for VGG16. VGG16 has the lowest robustness since it has the highest \ac{asr} and high distortion as \ac{ssim} score indicates. In \ac{fgsm} attack, it is hard to use \ac{psnr} and \ac{mad} to judge the robustness of the target models since the $||\delta||_2$ of all generated \acp{ae} are equal.

\subsubsection{Robustness under \ac{cw}-$L_\infty$ attacks}
\begin{SCfigure*}[1][h]
\centering
\resizebox{0.75\textwidth}{!}{%
    \setlength\tabcolsep{1.5pt}
    \tiny
    \begin{tabular}{ccccccccccc}
    \hline
    Clean &
    ViT-S-16 & 
    ViT-B-16 &
    ViT-L-16 &
    ViT-Res-16 &
    T2T-ViT-14 &
    T2T-ViT-24 &
    TNT-S-16 &
    ResNet50 &
    ResNeXt50 &
    VGG16 \\ \small
    
    \includegraphics[width=0.07\textwidth]{imgs/clean_imagenet-1k_81.png} &
    \includegraphics[width=0.07\textwidth]{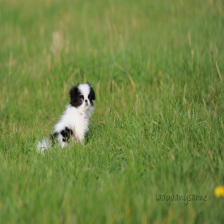} &
    \includegraphics[width=0.07\textwidth]{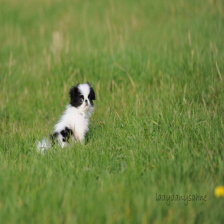} &
    \includegraphics[width=0.07\textwidth]{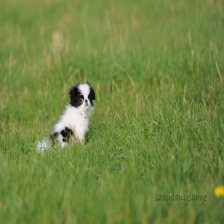} &
    \includegraphics[width=0.07\textwidth]{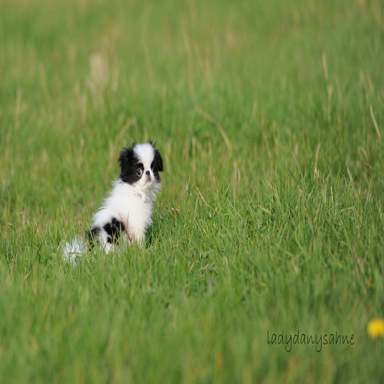} &
    \includegraphics[width=0.07\textwidth]{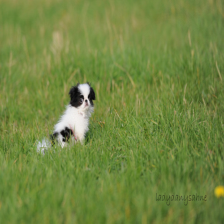} &
    \includegraphics[width=0.07\textwidth]{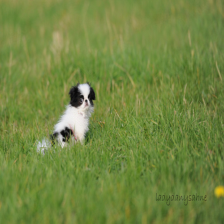} &
    \includegraphics[width=0.07\textwidth]{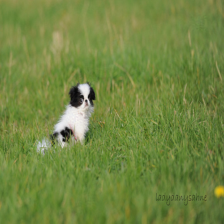} &
    \includegraphics[width=0.07\textwidth]{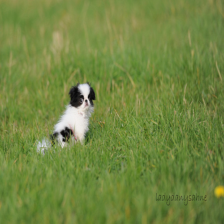} &
    \includegraphics[width=0.07\textwidth]{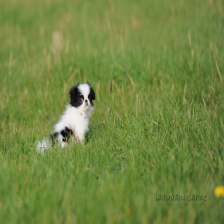} &
    \includegraphics[width=0.07\textwidth]{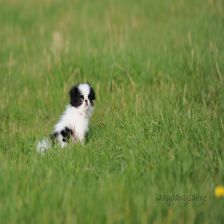} \\ \cline{0-0}
    
    \multirow{-3}{0.07\textwidth}{\subfloat[]{\label{fig:cwi_maps_a}}} &
    \includegraphics[width=0.07\textwidth]{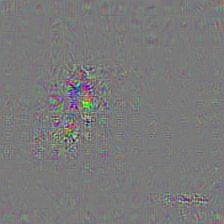} &
    \includegraphics[width=0.07\textwidth]{imgs/perturb_cwi_vit_base_patch16_224_imagenet-1k_81.png} &
    \includegraphics[width=0.07\textwidth]{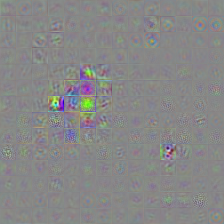} &
    \includegraphics[width=0.07\textwidth]{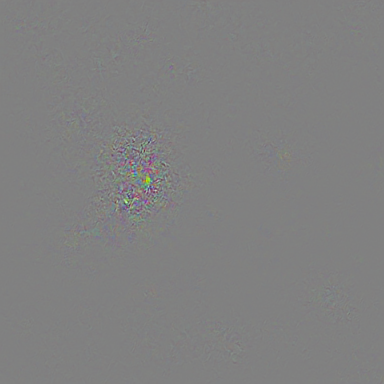} &
    \includegraphics[width=0.07\textwidth]{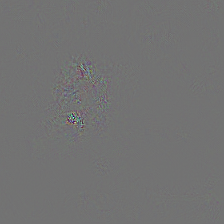} &
    \includegraphics[width=0.07\textwidth]{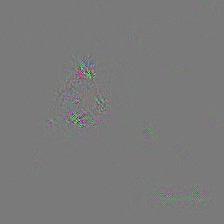} &
    \includegraphics[width=0.07\textwidth]{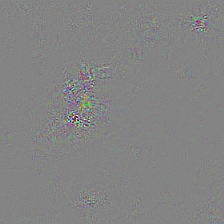} &
    \includegraphics[width=0.07\textwidth]{imgs/perturb_cwi_resnet50_imagenet-1k_81.png} &
    \includegraphics[width=0.07\textwidth]{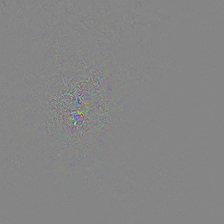} &
    \includegraphics[width=0.07\textwidth]{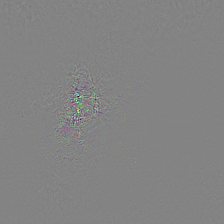} \\
    
    &
    \includegraphics[width=0.07\textwidth]{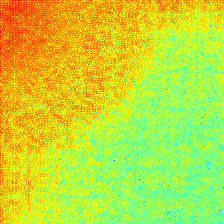} &
    \includegraphics[width=0.07\textwidth]{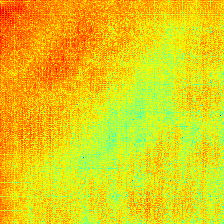} &
    \includegraphics[width=0.07\textwidth]{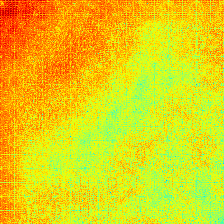} &
    \includegraphics[width=0.07\textwidth]{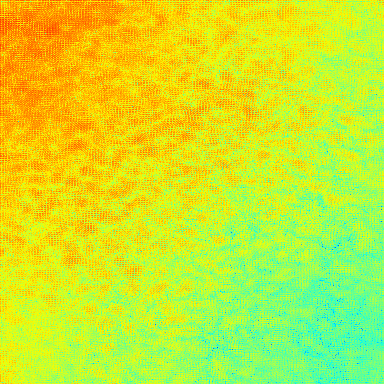} &
    \includegraphics[width=0.07\textwidth]{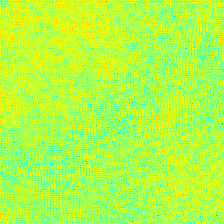} &
    \includegraphics[width=0.07\textwidth]{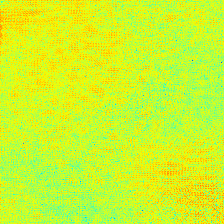} &
    \includegraphics[width=0.07\textwidth]{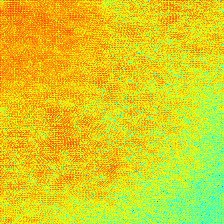} &
    \includegraphics[width=0.07\textwidth]{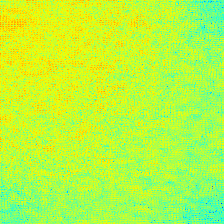} &
    \includegraphics[width=0.07\textwidth]{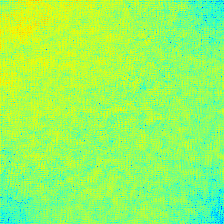} &
    \includegraphics[width=0.07\textwidth]{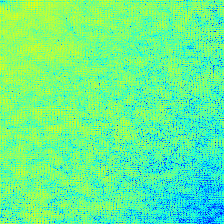} \\ \hline
    \end{tabular}
}%
    \hspace{-3mm}\caption{\protect\rule{0ex}{5ex}\begin{footnotesize}\textbf{\ac{cw}-$L_\infty$ attack:} The first row shows the clean sample and the \acp{ae}. The clean image is correctly classified by tested models and all \acp{ae} are successful attacks. (a) The perturbation (top) and the corresponding \acs{dct}-based spectral decomposition heatmap. Perturbation is scaled from [-1, 1] to [0, 255]. \end{footnotesize}}
    \label{fig:cwi_maps}
    \vspace{-5mm}
\end{SCfigure*}

Figure \ref{fig:cwi_maps} shows an example to study the target models' robustness against \ac{cw}-$L_\infty$ attack. Figure \ref{fig:cwi_quality} shows the visual quality scores of the \acp{ae} that are generated using \ac{cw}-$L_\infty$ attack. While Figure \ref{fig:cwi_defense} shows the \ac{asr} of the \acp{ae} and  the top-1 error of the pre-processed \acp{ae} that are generated using \ac{cw}-$L_\infty$ attack.

For \ac{cw}-$L_\infty$, Figure \ref{fig:cwi_defense} shows that the \ac{asr} of the \ac{vit}-B/L-16 is lower than the \ac{asr} of other models. On the other hand, the \ac{dct} decomposition, see Figure \ref{fig:cwi_maps_a}, shows that the spread of the discriminative features is wider on \ac{vit}-B/L-16 than other models. Moreover, it is shown that \ac{t2t}-24 and \ac{tnt}-S-16 have wider spread of \ac{dct} decomposition than ResNet. Finally, from Figure \ref{fig:cwi_quality}, we can conclude that \ac{cw}-$L_\infty$ generates \acp{ae} with higher perturbations for \ac{vit}-B/L-16 and for \ac{t2t}-24 models than other model which clearly affected the image structure and the visual perception as \ac{ssim} and \ac{mad} scores indicate. 

\subsubsection{Robustness under \ac{rays} attacks}
\begin{SCfigure*}[1][h]
\centering
\resizebox{0.75\textwidth}{!}{%
    \setlength\tabcolsep{1.5pt}
    \tiny
    \begin{tabular}{ccccccccccc}
    \hline
    Clean &
    ViT-S-16 & 
    ViT-B-16 &
    ViT-L-16 &
    ViT-Res-16 &
    T2T-ViT-14 &
    T2T-ViT-24 &
    TNT-S-16 &
    ResNet50 &
    ResNeXt50 &
    VGG16 \\ \small
    
    \includegraphics[width=0.07\textwidth]{imgs/clean_imagenet-1k_6.png} &
    \includegraphics[width=0.07\textwidth]{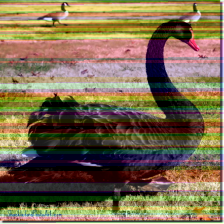} &
    \includegraphics[width=0.07\textwidth]{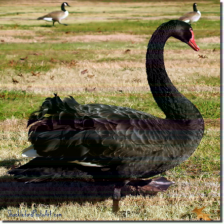} &
    \includegraphics[width=0.07\textwidth]{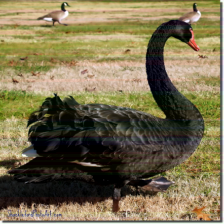} &
    \includegraphics[width=0.07\textwidth]{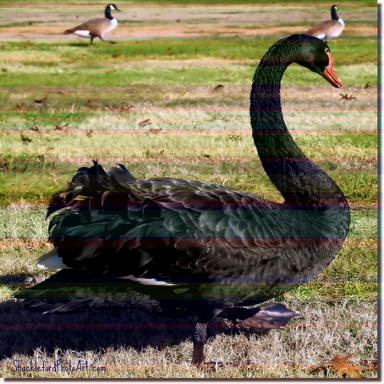} &
    \includegraphics[width=0.07\textwidth]{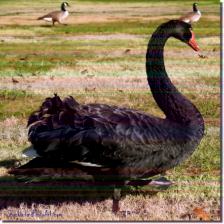} &
    \includegraphics[width=0.07\textwidth]{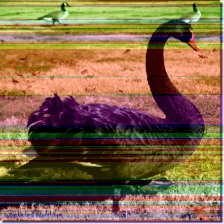} &
    \includegraphics[width=0.07\textwidth]{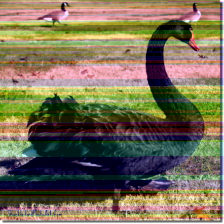} &
    \includegraphics[width=0.07\textwidth]{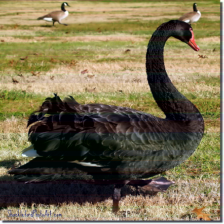} &
    \includegraphics[width=0.07\textwidth]{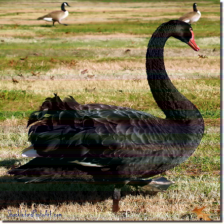} &
    \includegraphics[width=0.07\textwidth]{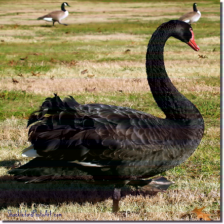} \\ \cline{0-0}
    
    \multirow{-3}{0.07\textwidth}{\subfloat[]{\label{fig:rays_maps_a}}} &
    \includegraphics[width=0.07\textwidth]{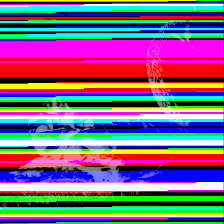} &
    \includegraphics[width=0.07\textwidth]{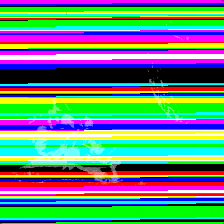} &
    \includegraphics[width=0.07\textwidth]{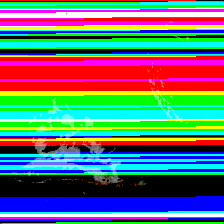} &
    \includegraphics[width=0.07\textwidth]{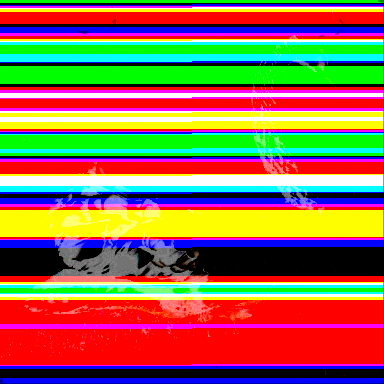} &
    \includegraphics[width=0.07\textwidth]{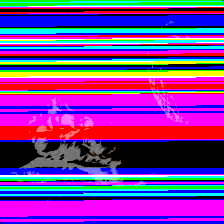} &
    \includegraphics[width=0.07\textwidth]{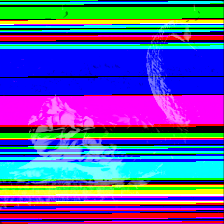} &
    \includegraphics[width=0.07\textwidth]{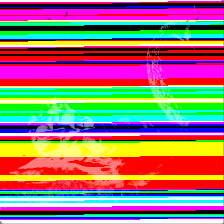} &
    \includegraphics[width=0.07\textwidth]{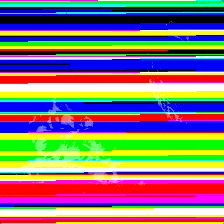} &
    \includegraphics[width=0.07\textwidth]{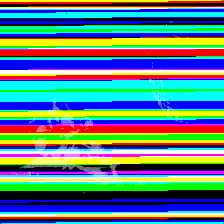} &
    \includegraphics[width=0.07\textwidth]{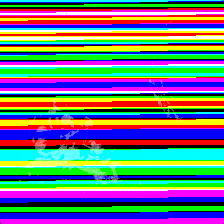} \\
    
    &
    \includegraphics[width=0.07\textwidth]{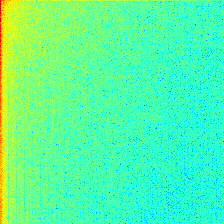} &
    \includegraphics[width=0.07\textwidth]{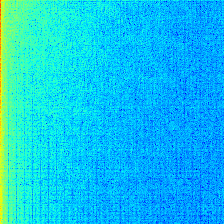} &
    \includegraphics[width=0.07\textwidth]{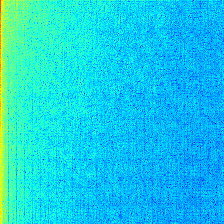} &
    \includegraphics[width=0.07\textwidth]{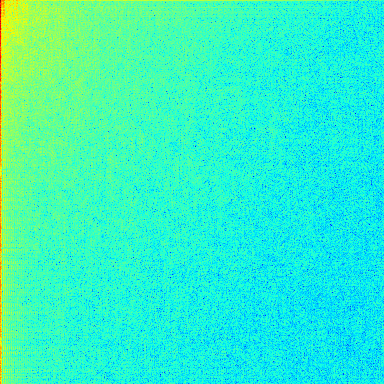} &
    \includegraphics[width=0.07\textwidth]{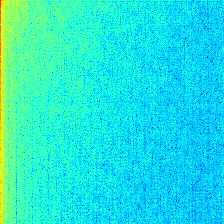} &
    \includegraphics[width=0.07\textwidth]{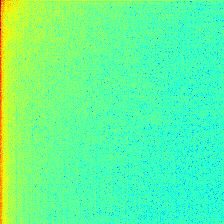} &
    \includegraphics[width=0.07\textwidth]{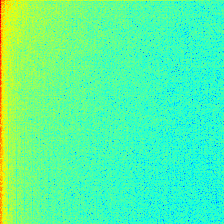} &
    \includegraphics[width=0.07\textwidth]{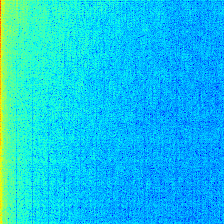} &
    \includegraphics[width=0.07\textwidth]{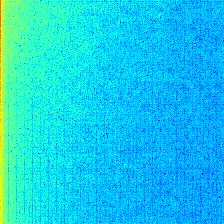} &
    \includegraphics[width=0.07\textwidth]{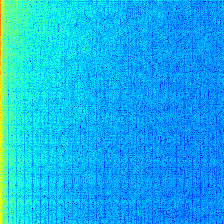} \\ \hline
    \end{tabular}
}%
    \hspace{-3mm}\caption{\protect\rule{0ex}{5ex}\begin{footnotesize}\textbf{\ac{rays} attack:} The first row shows the clean sample and the \acp{ae}. The clean image is correctly classified by tested models and all \acp{ae} are successful attacks. (a) The perturbation (top) and the corresponding \acs{dct}-based spectral decomposition heatmap. Perturbation is scaled from [-1, 1] to [0, 255]. \end{footnotesize}}
    \label{fig:rays_maps}
    \vspace{-5mm}
\end{SCfigure*}

\begin{figure*}
\resizebox{\textwidth}{!}{%
    \setlength\tabcolsep{1.5pt}
    \tiny
    \begin{tabular}{cc}
        \begin{tabular}{l}\subfloat[]{\includegraphics[width=0.6\textwidth]{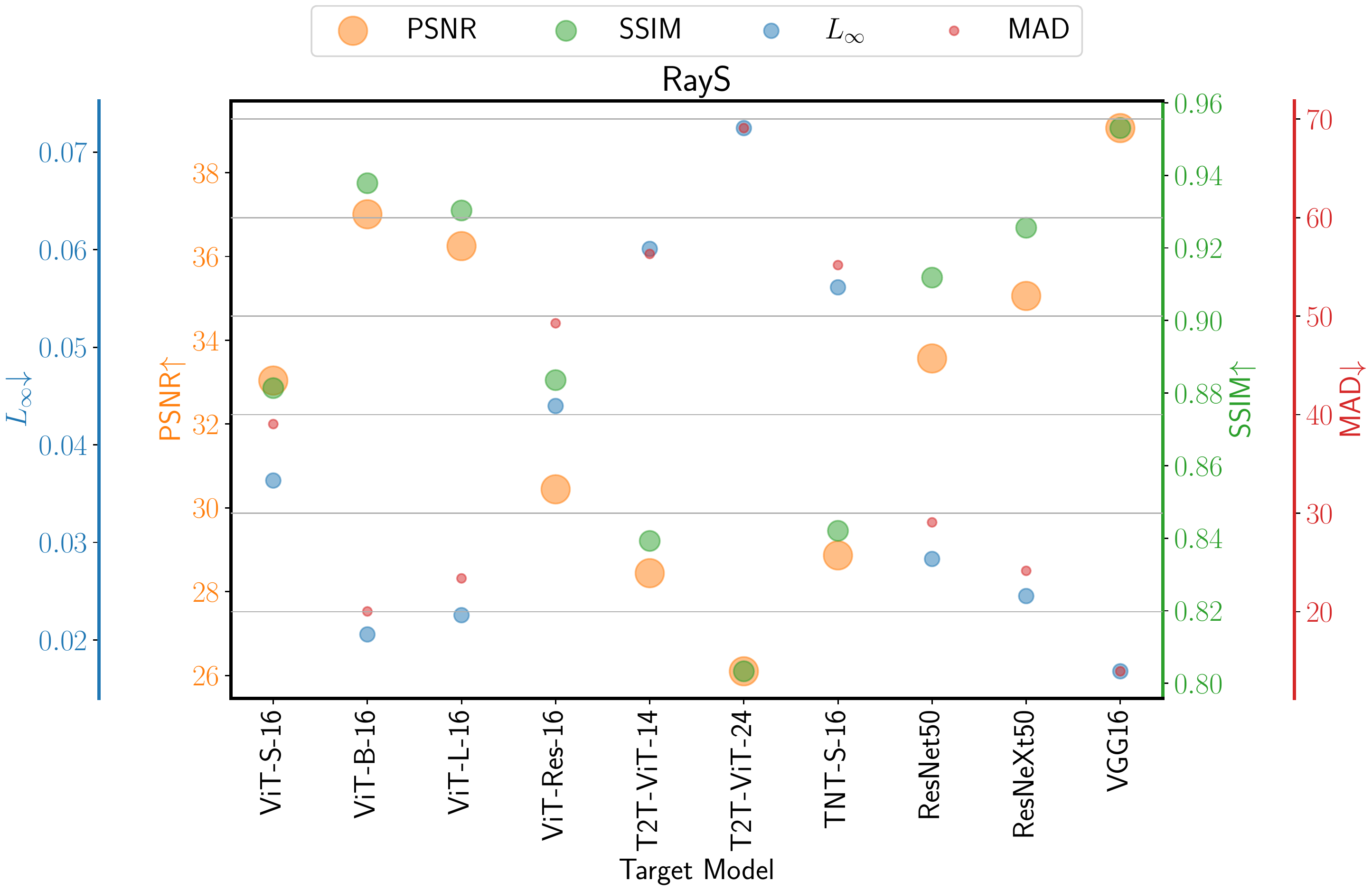}\label{fig:rays_quality}}\end{tabular} &
        \begin{tabular}{l}\subfloat[]{\includegraphics[width=0.4\textwidth]{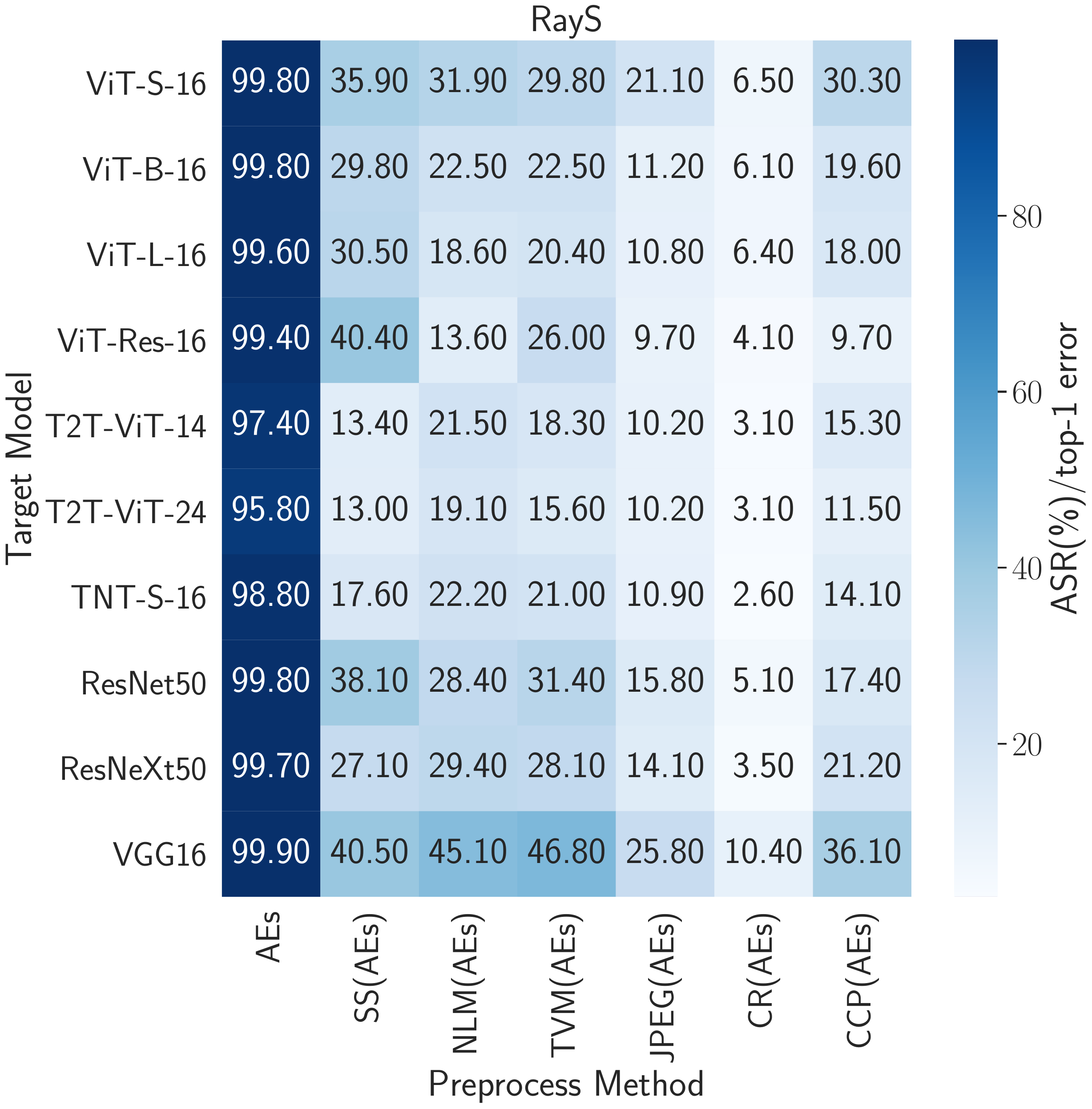}\label{fig:rays_defense}} \end{tabular} \\
    \end{tabular}
}%
    \vspace{-8mm}
    \hspace{-3mm}\caption{\protect\rule{0ex}{5ex}\begin{footnotesize}\textbf{\ac{rays} attack:} (a) \acp{ae} quality assessment measures. (b) The \ac{asr} of the \acp{ae} and  the top-1 error of the pre-processed \acp{ae}  on 1000 images from imagenet-1k. SS: \acl{ss}. NLM: \acl{nlm}. TVM: \acl{tvm}. JPEG: \acl{jpg}. CR: \acl{cr}. CCP: \acl{ccp}. \end{footnotesize}}
    \vspace{-3mm}
    \label{fig:rays_defense_quality}
\end{figure*}

Figure \ref{fig:rays_maps} shows an example to study the target models' robustness against \ac{rays} attack. Figure \ref{fig:rays_quality} shows the visual quality scores of the \acp{ae} that are generated using \ac{rays} attack. While Figure \ref{fig:rays_defense} shows the \ac{asr} of the \acp{ae} and  the top-1 error of the pre-processed \acp{ae} that are generated using \ac{rays} attack.

For \ac{rays} attack, by looking at Figure \ref{fig:rays_defense}, Figure \ref{fig:rays_maps_a}, and Figure \ref{fig:rays_quality}, we can conclude that hybrid-\acp{vit} are more robust than other target models. The figures show 1) \ac{t2t}-24 has lower \ac{asr} than other models. 2) hybrid-\acp{vit} have wider energy spectrum spread than other models. 3) \ac{rays} generates \acp{ae} with higher perturbations for hybrid-\acp{vit} that are perceptible to human.

\subsection{Transfer attacks: increasing the number of attention blocks reduces the transferability}
\begin{figure*}
\resizebox{\textwidth}{!}{%
    \setlength\tabcolsep{1.5pt}
    \tiny

    \begin{tabular}{cc}
        \begin{tabular}{l}\subfloat[]{\includegraphics[width=0.5\textwidth]{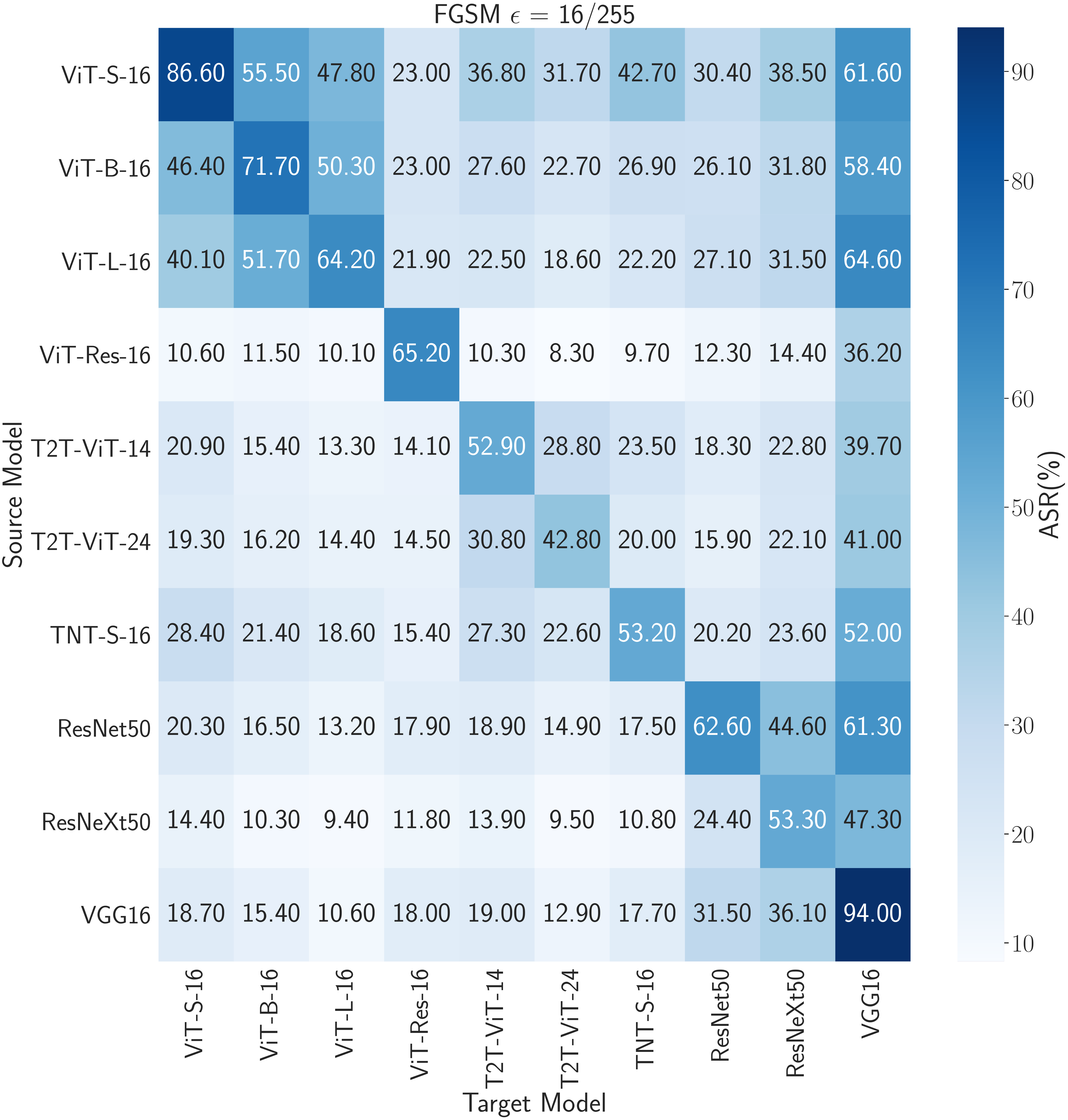}\label{fig:fgsm_transfer}}\end{tabular} &
        \begin{tabular}{l}\subfloat[]{\includegraphics[width=0.5\textwidth]{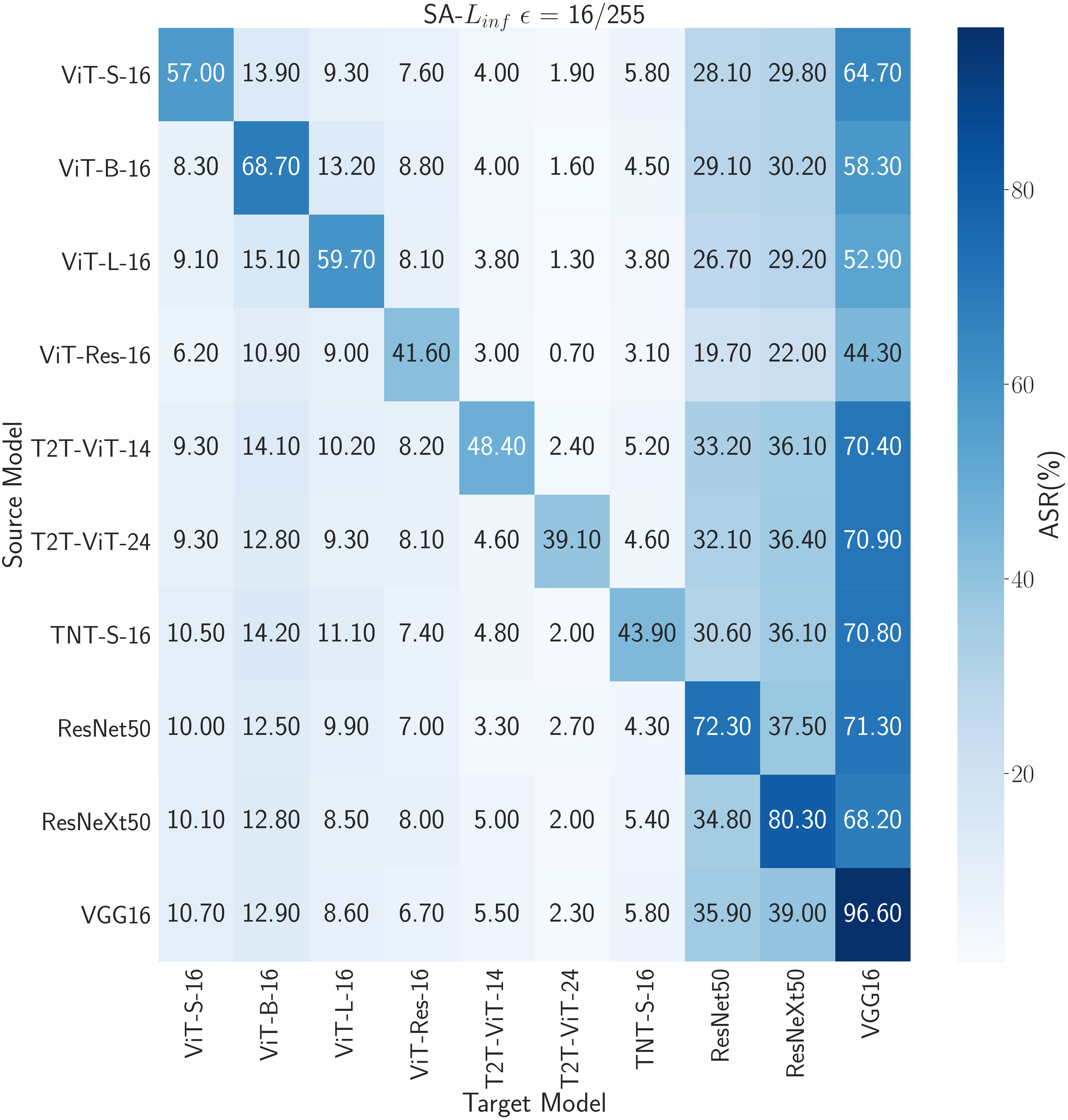}\label{fig:sa_transfer}} \end{tabular} \\
    \end{tabular}
}%
    \vspace{-8mm}
    \hspace{-3mm}\caption{\protect\rule{0ex}{5ex}\begin{footnotesize}\ac{asr} of transfer attack using \ac{fgsm} and \ac{sa} on 1000 images from ImageNet-1k . The row represents the source model that is used to generate \ac{ae}. The column represents the target model. Dark blue column means that the target model is vulnerable to transfer attacks.\end{footnotesize}}
    \vspace{-2mm}
    \label{fig:fgsm_sa_transfere}
\end{figure*}

\begin{SCfigure*}[1][h]
\centering
\resizebox{0.75\textwidth}{!}{%
    \setlength\tabcolsep{1.5pt}
    \tiny
    \begin{tabular}{ccccccccccc}
    
    \hline
    Clean &
    ViT-S-16 & 
    ViT-B-16 &
    ViT-L-16 &
    ViT-Res-16 &
    T2T-ViT-14 &
    T2T-ViT-24 &
    TNT-S-16 &
    ResNet50 &
    ResNeXt50 &
    VGG16 \\ \Large
    
    \includegraphics[width=0.07\textwidth]{imgs/clean_imagenet-1k_81.png} &
    \includegraphics[width=0.07\textwidth]{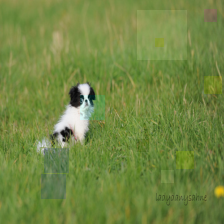} &
    \includegraphics[width=0.07\textwidth]{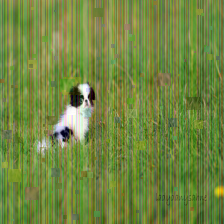} &
    \includegraphics[width=0.07\textwidth]{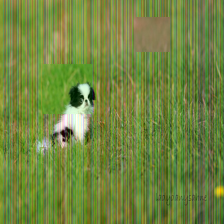} &
    \includegraphics[width=0.07\textwidth]{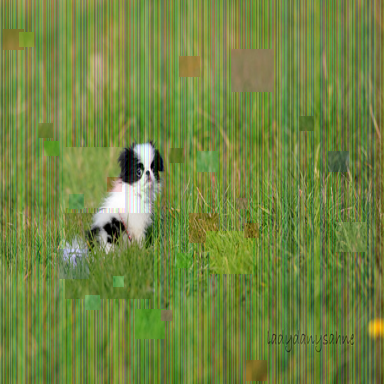} &
    \includegraphics[width=0.07\textwidth]{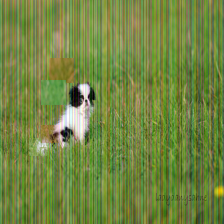} &
    \includegraphics[width=0.07\textwidth]{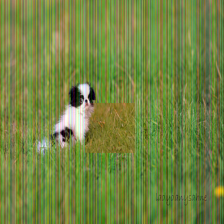} &
    \includegraphics[width=0.07\textwidth]{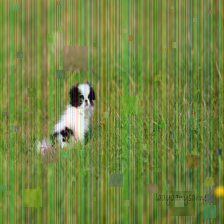} &
    \includegraphics[width=0.07\textwidth]{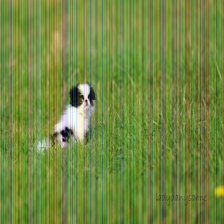} &
    \includegraphics[width=0.07\textwidth]{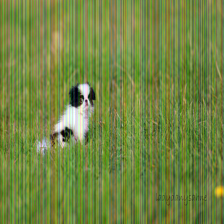} &
    \includegraphics[width=0.07\textwidth]{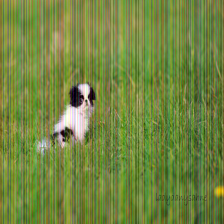} \\ \cline{0-0}
    
    \multirow{-3}{0.08\textwidth}{\subfloat[]{\label{fig:sa_maps_a}}} &
    \includegraphics[width=0.07\textwidth]{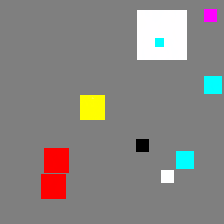} &
    \includegraphics[width=0.07\textwidth]{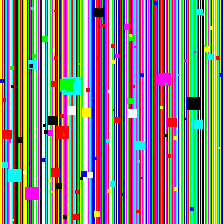} &
    \includegraphics[width=0.07\textwidth]{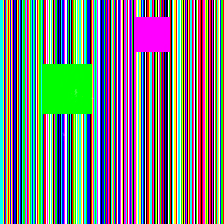} &
    \includegraphics[width=0.07\textwidth]{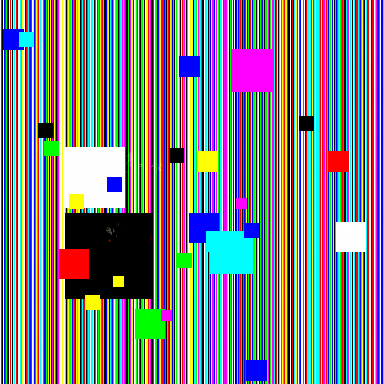} &
    \includegraphics[width=0.07\textwidth]{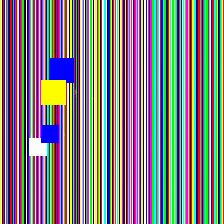} &
    \includegraphics[width=0.07\textwidth]{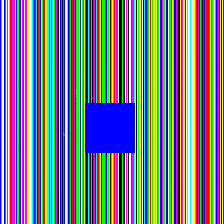} &
    \includegraphics[width=0.07\textwidth]{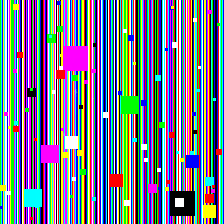} &
    \includegraphics[width=0.07\textwidth]{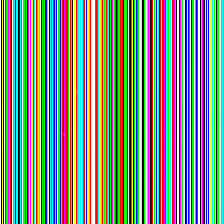} &
    \includegraphics[width=0.07\textwidth]{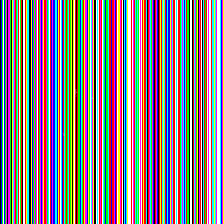} &
    \includegraphics[width=0.07\textwidth]{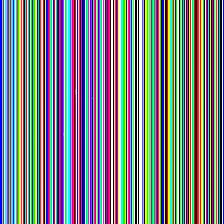} \\
    
    &
    \includegraphics[width=0.07\textwidth]{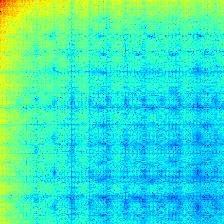} &
    \includegraphics[width=0.07\textwidth]{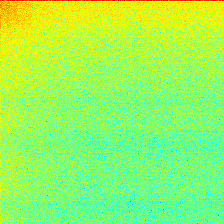} &
    \includegraphics[width=0.07\textwidth]{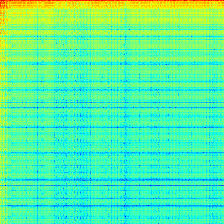} &
    \includegraphics[width=0.07\textwidth]{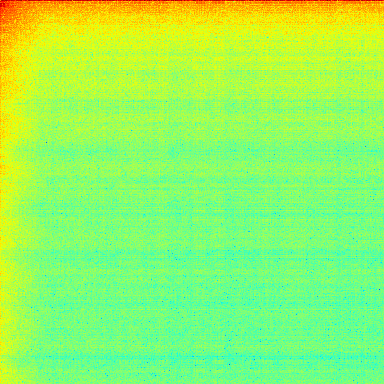} &
    \includegraphics[width=0.07\textwidth]{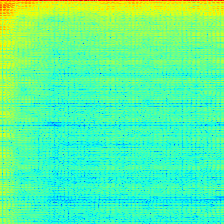} &
    \includegraphics[width=0.07\textwidth]{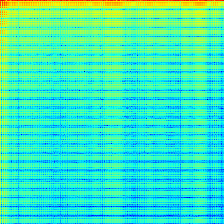} &
    \includegraphics[width=0.07\textwidth]{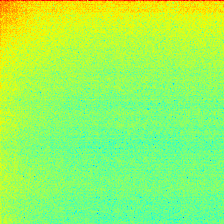} &
    \includegraphics[width=0.07\textwidth]{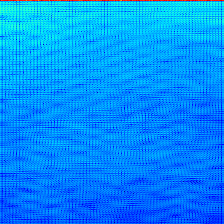} &
    \includegraphics[width=0.07\textwidth]{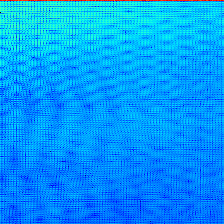} &
    \includegraphics[width=0.07\textwidth]{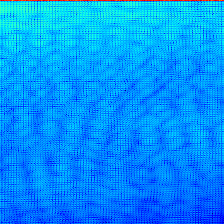} \\ \hline
    \end{tabular}
}%
    \hspace{-3mm}\caption{\begin{footnotesize}\textbf{\ac{sa}-$L_\infty$ $\epsilon=8/255$ attack:} The first row shows the clean sample and the \acp{ae}. The clean image is correctly classified by tested models and all \acp{ae} are successful attacks. (a) The perturbation (top) and the corresponding \acs{dct}-based spectral decomposition heatmap. Perturbation is scaled from [-1, 1] to [0, 255]. \end{footnotesize}}
    \label{fig:sa_maps}
\end{SCfigure*}

Recent studies in~\cite{bhojanapalli2021understanding,shao2021adversarial,mahmood2021robustness} showed that there is low transferability between different models' families. In this work we confirm that and the results are shown in Figure \ref{fig:fgsm_transfer}. Here, we show two new observations. The first one is that the transferability, within the same model family, is becoming lower when the model is becoming larger. Hence, adding more attention blocks to \ac{vit} variant models reduces the effect of the transferability property, as shown in Figure \ref{fig:fgsm_transfer}. The second\footnote{Using source model to generate black box \acp{ae} for a different target model is not popular in real-world but, here the goal is to note the observation.} observation is that the black box based \acp{ae}, \ac{rays} and \ac{sa}, that are generated using \ac{vit} variants are more transferable to \acp{cnn} while the black box based \acp{ae} that are generated using \ac{cnn} are much less transferable to \ac{vit} variants. As shown in Figure \ref{fig:sa_transfer}, we notice that when \acp{cnn} serve as target models, last three columns, the \ac{asr} is higher than those of when \acp{cnn} serve as the source model, last three rows. According to the \ac{dct} decomposition of \ac{sa}-based \acp{ae}, see Figure \ref{fig:sa_maps}, one possible explanation to this case is that the generated perturbations highly affect the local features than global features and since \acp{vit} are less sensitive to local features making perturbation effect less transferable.

\begin{figure*}
\vspace{-5mm}
\resizebox{\textwidth}{!}{%
    \setlength\tabcolsep{1.5pt}
    \tiny
     \begin{tabular}{cc}
        \begin{tabular}{l}\subfloat[]{\includegraphics[width=0.6\textwidth]{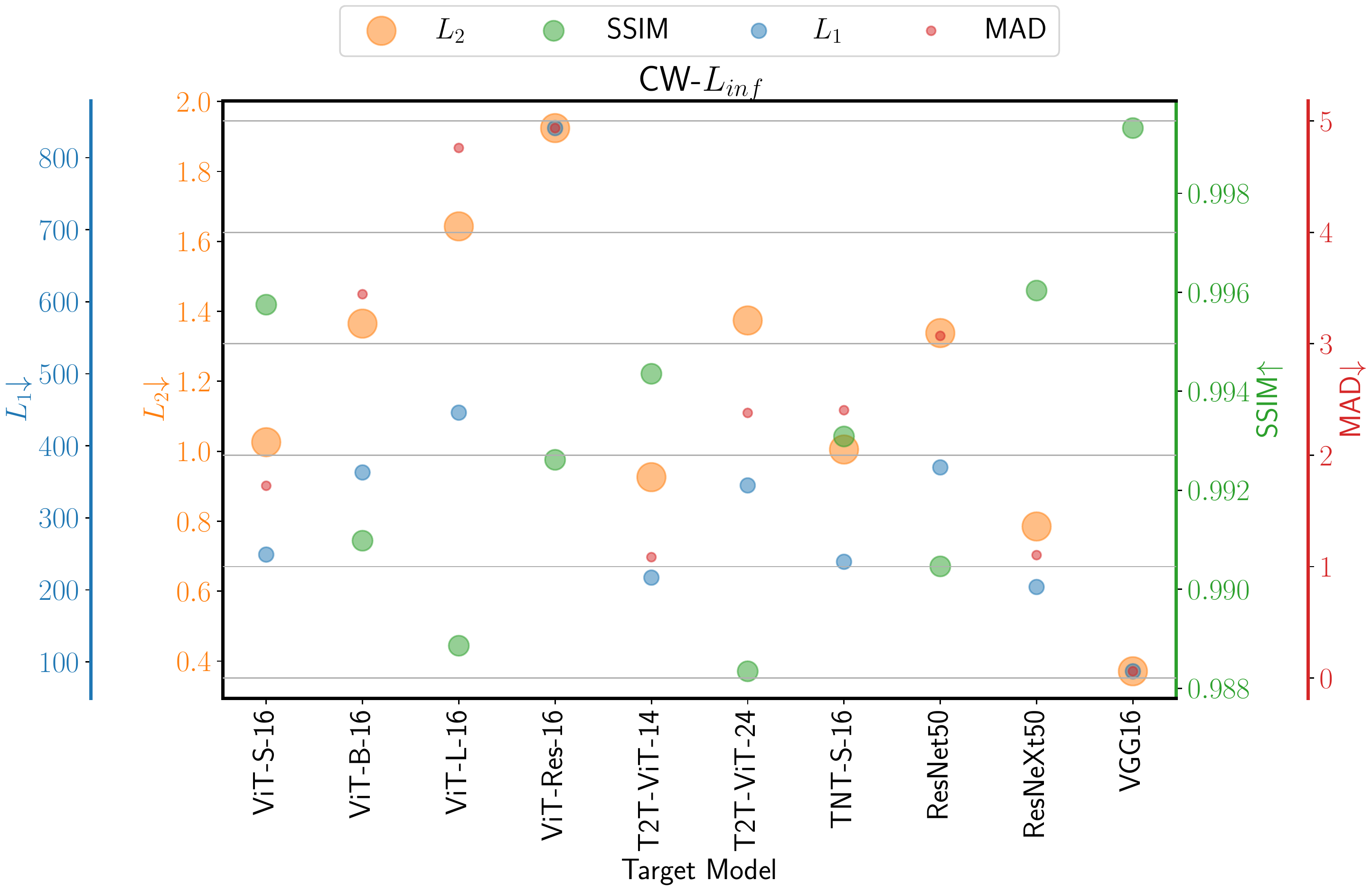}\label{fig:cwi_quality}}\end{tabular} &
        \begin{tabular}{l}\subfloat[]{\includegraphics[width=0.4\textwidth]{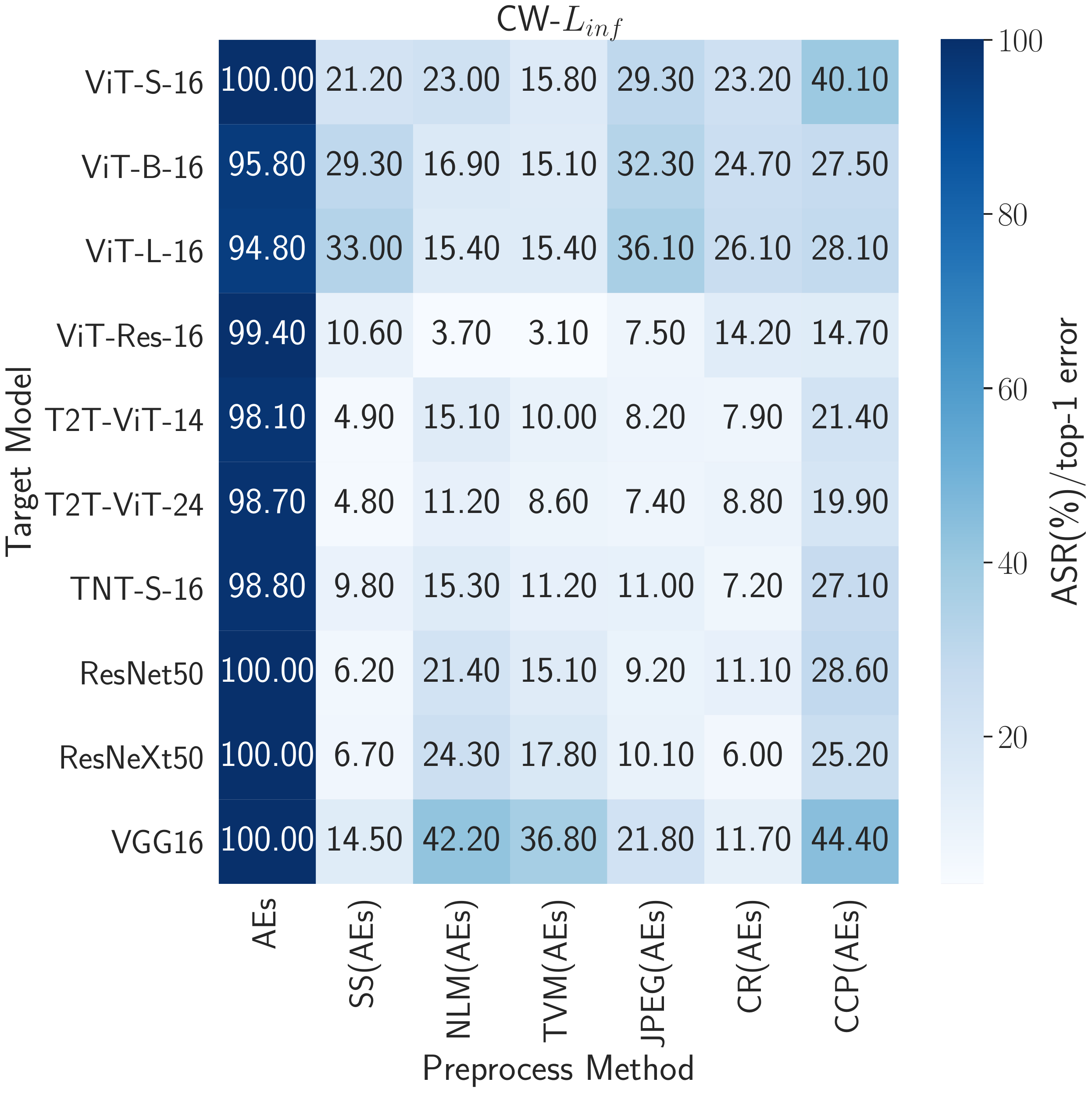}\label{fig:cwi_defense}} \end{tabular} \\
    \end{tabular}
}%
\vspace{-8mm}
    \hspace{-3mm}\caption{\protect\rule{0ex}{5ex}\begin{footnotesize}\textbf{\ac{cw}-$L_\infty$ attack:} (a) \acp{ae} quality assessment measures. (b) The \ac{asr} of the \acp{ae} and  the top-1 error of the pre-processed \acp{ae}  on 1000 images from imagenet-1k. SS: \acl{ss}. NLM: \acl{nlm}. TVM: \acl{tvm}. JPEG: \acl{jpg}. CR: \acl{cr}. CCP: \acl{ccp}. \end{footnotesize}}
    \vspace{-3mm}
    \label{fig:cwi_defense_quality}
\end{figure*}

\subsection{\Acf{ccp} is an attack and defense!}
\begin{figure}[!ht]
    \centering
    \vspace{-3mm}
    \resizebox{0.9\linewidth}{!}{%
    \begin{tabular}{ccc}
         \includegraphics[width=0.35\textwidth]{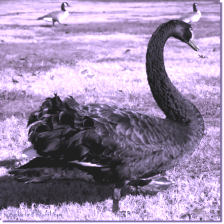}& \includegraphics[width=0.35\textwidth]{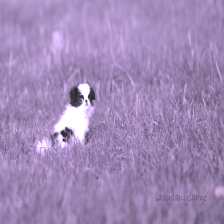}&\includegraphics[width=0.35\textwidth]{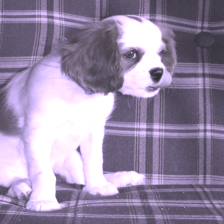}\\
    \end{tabular}
    }%
    \caption{Examples for \ac{ccp} attack. Images are from ImageNet-1k validation set.}
    \label{fig:ccp_examples}
\end{figure}

The \ac{ccp} attack is based on the color property of the image. It uses the original color channels, Red $R$, Green $G$ and Blue $B$ of the image input $x$ to generate the \ac{ae} ($x^\prime$). The \ac{ae} is composed of new transformed color channels, $R^\prime$, $G^\prime$, and $B^\prime$  of the transformed image ($x^\prime$). The transformed channels are calculated as follows:

\begin{ceqn}
    \begin{equation} \label{eq:ccp}
    \begin{gathered} 
        R^\prime=s\left(\frac{\alpha^r R + \alpha^g G + \alpha^b B}{3}\right)+b, \\
        G^\prime=s\left(\frac{\beta^r R + \beta^g G + \beta^b B}{3}\right)+b, \\
        B^\prime=s\left(\frac{\gamma^r  R + \gamma^g  G + \gamma^b B}{3}\right)+b,
    \end{gathered}
    \end{equation}
\end{ceqn}
where $s$ is a scale factor hyperparameter, $b$ is a bias hyperparameter, \{$\alpha$, $\beta$, $\gamma$\}  $\in [0, 1]$, and \{$R^\prime$, $G^\prime$, $B^\prime$\} $\in \mathbb{R}$. The scale $s$ and bias $b$ are used to adjust the visual appearance of the generated \ac{ae}. Figure \ref{fig:ccp_examples} shows examples for \ac{ccp} attack.

Images might be vulnerable to natural perturbations like color brightness change. 
Table \ref{tab:ccp_asr} shows the \ac{asr} of the target models against the \ac{ccp} attack. Although this attack doesn't achieve high \ac{asr} when tested on neural network models, it can reveal robustness of these models. The results in Table \ref{tab:ccp_asr} show that hybrid-\acp{vit} have the highest robustness, while ResNets and \ac{vit}-B/L-16 have comparable robustness. One explanation to that is that \ac{ccp} limits \acp{cnn}' and vanilla \acp{vit}' capabilities to extract the local and global features, respectively. Hence, enhancing the tokenization process of the vanilla \acp{vit} as in hybrid-\acp{vit} has an added value to model's robustness. Surprisingly, we found that when applying \ac{ccp} attack on \acp{ae}, \ac{ccp} is able to project the \acp{ae} back to its original manifold. Last columns of Figures \ref{fig:sal_defense}, \ref{fig:pgd_l1_400_defense}, \ref{fig:cw2_defense}, and, \ref{fig:pgd_li_0.003_defense} show the \ac{asr} after considering \ac{ccp} as a preprocess method for the \acp{ae}. Although removing the effect of perturbations depends on noise amount, hybrid-\acp{vit} show better performance on predicting the original class due the presence of the tokenization process enhancements in hybrid-\acp{vit}. Future investigations are highly recommended to study this phenomenon since brightness change is a common process in many real-world applications. 
\begin{table*}[h]
\vspace{1mm}
    \centering
    \caption{The \ac{asr} of the target models against \acf{ccp}.}
    \label{tab:ccp_asr}
    \resizebox{\textwidth}{!}{%
        \begin{tabular}{c|c|c|c|c|c|c|c|c|c|c|}
        \cline{2-11}
        &
        ViT-S-16 & 
        ViT-B-16 &
        ViT-L-16 &
        ViT-Res-16 &
        T2T-ViT-14 &
        T2T-ViT-24 &
        TNT-S-16 &
        ResNet50 &
        ResNeXt50 &
        VGG16 \\ \cline{1-11}
        \multicolumn{1}{|c|}{ASR(\%)} &
        22.6 &
        15.2 &
        12.5 &
        \textbf{5.4} &
        \textbf{10.1} &
        \textbf{7.1} &
        \textbf{8.1} &
        11.9 &
        14.9 &
        32.6 \\ \cline{1-11}

        \end{tabular}
    }%
\end{table*}

\begin{table*}[h]
\renewcommand{\arraystretch}{1.1} 
\centering
\caption{The top-1 error(\%) of the preprocessing defense methods that are applied to \ac{pgd}-$L_\infty$ $\epsilon=4/255$ under the \ac{eot}. 1000 images from imagenet-1k are used. SS: \acl{ss}. NLM: \acl{nlm}. TVM: \acl{tvm}. JPEG: \acl{jpg}. CR: \acl{cr}. EOT: \acl{eot}.}
\label{tab:asr_eot_defense}
\resizebox{\textwidth}{!}{%
\begin{tabular}{|l|c||c|c||c|c||c|c||c|c||c|c|}
\hline
{Model}                    & {\shortstack{No defense}} & \multicolumn{2}{c||}{SS}  & \multicolumn{2}{c||}{NLM}  & \multicolumn{2}{c||}{TVM}  & \multicolumn{2}{c||}{JPEG} & \multicolumn{2}{c|}{CR}  \\ \cline{2-12}
\ac{eot} &      \XSolidBrush     & \XSolidBrush & \Checkmark & \XSolidBrush     & \Checkmark & \XSolidBrush   & \Checkmark & \XSolidBrush    & \Checkmark & \XSolidBrush     & \Checkmark   \\ \hline \hline
\ac{vit}-S-16 & 99.9          & 96.3 & 98     & 43.8 & 77.7    & 62   & 91.2    & 91.8 & 94.3     & 97.4 & 90.3   \\ \hline
\ac{vit}-B-16   & 99.5          & 97.2 & 97.9   & 35.9 & 86.4    & 58.5 & 94.4    & 90.2 & 94.8     & 96.4 & 95.6   \\ \hline
\ac{vit}-L-16  & 98.8          & 96.5 & 97     & 33.2 & 83.9    & 58.6 & 91.6    & 88.2 & 94       & 93.7 & 92     \\ \hline
ViT-Res-16 & 100           & 96.7 & 99.6   & 12.3 & 89.6    & 34.1 & 97.8    & 57.5 & 96.9     & 98.1 & 98.2   \\ \hline
\ac{t2t}-14          & 99.9          & 58.1 & 96.8   & 22.8 & 77.1    & 24.3 & 87.4    & 25   & 77.4     & 74.7 & 70     \\ \hline
\ac{t2t}-24         & 99.6          & 52.3 & 94.6   & 18.1 & 66      & 21.2 & 82.1    & 23.9 & 70.1     & 67.4 & 63.8   \\ \hline
\ac{tnt}-S-16     & 99.7          & 80.7 & 97.8   & 27.1 & 78.8    & 34.2 & 91.1    & 38.4 & 82.2     & 83   & 79.7   \\ \hline
ResNet50                 & 98.7          & 61.4 & 95.7   & 26.9 & 81.3    & 26.3 & 89      & 27.9 & 85.4     & 86.8 & 79.4   \\ \hline
ResNet50-32x4d         & 98.4          & 56.3 & 95     & 29.4 & 79.8    & 25.4 & 87.2    & 22.5 & 79.3     & 81.8 & 71.9   \\ \hline
VGG16                    & 99.6          & 94.1 & 96.5   & 55   & 89.2    & 57.7 & 90.2    & 73.8 & 95.5     & 99.4 & 94.5  \\\hline
\end{tabular}
}%
\end{table*}

\subsection{Vanilla \acp{vit} are not responding to preprocessing defenses that mainly reduce the high frequency components.}

\begin{figure}
    \centering
    \vspace{-3mm}
    \includegraphics[width=\linewidth]{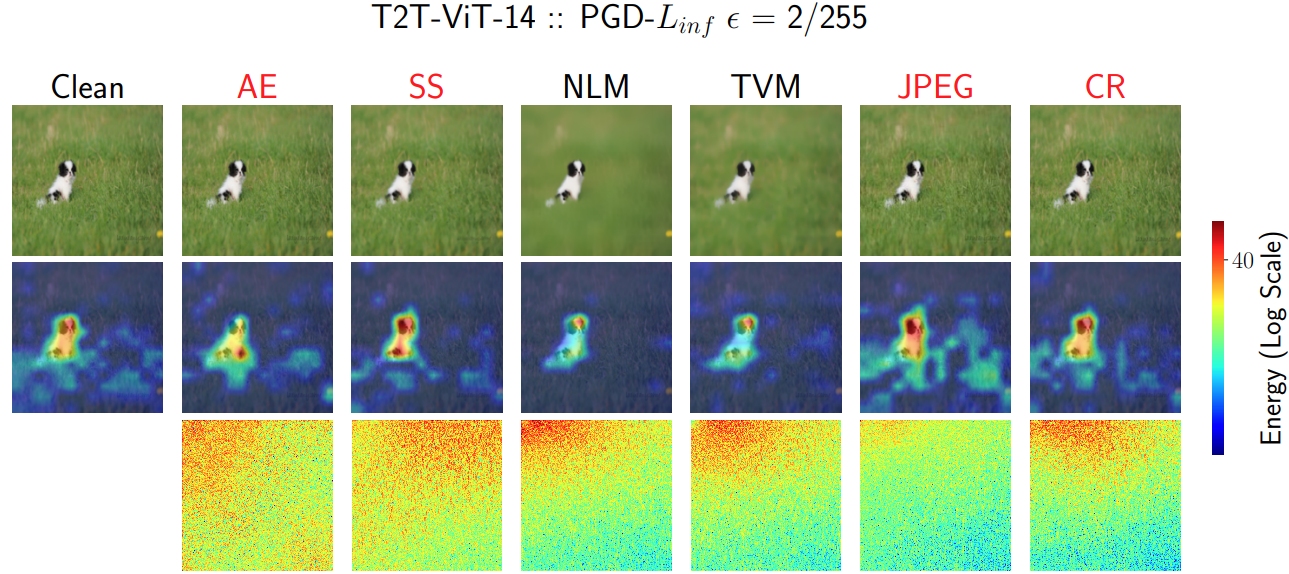}
    
    \vspace{-5mm}
    \caption{\protect\rule{0ex}{5ex}\begin{footnotesize}Examples of the defense preprocessing. The top row shows the clean, \ac{ae}, and the preprocessed \ac{ae}. The middle row shows the \ac{gcam} of the clean, \ac{ae}, and the preprocessed \ac{ae}. The bottom row shows the energy spectrum of the \ac{dct} decomposition of \ac{ae}, and the preprocessed \ac{ae}. Red label means the image is misclassified. \end{footnotesize}}
    \vspace{-5mm}
    \label{fig:adv_defense}
\end{figure}

Preprocessing is one of the defense methods that is applied to the model's input to remove the effect of perturbations that are added to the input image. In this experiment, we apply five preprocessing methods that are used in the literature and briefly mentioned in Section \ref{sec:defense_methods}. Samples of top-1 error after perprocessing of some attacks are shown in Figures \ref{fig:sal_defense}, \ref{fig:pgd_l1_400_defense}, \ref{fig:cw2_defense}, \ref{fig:pgd_l2_2_defense}, \ref{fig:pgd_li_0.003_defense}, \ref{fig:fgsm_li_0.003_defense}, \ref{fig:cwi_defense}, and \ref{fig:rays_defense}, while Figure \ref{fig:defense_all} shows the top-1 error \ac{aa}, in average. 
For $L_0$-based and \ac{cw}-$L_2$ attacks, \ac{ss} and \ac{cr}, as expected, have the capability to remove the perturbations effect and to project \acp{ae} back to input manifold, see Figure \ref{fig:sal_defense}, by re-positioning the perturbation structure. The limited success of the preprocessing against $L_0$-based \acp{ae} that are generated using VGG16 is due the large number of the impacted pixels. For other attacks, \ac{nlm} and \ac{tvm} show better performance on the preprocessing of \acp{ae}. That's because these two denoising methods try to restore the original image while preserving the global image structure and contours. While other preprocessing methods including \ac{ss}, \ac{jpg}, and \ac{cr}, are highly impacting the high frequency components of the \acp{ae}, hence ResNets show better top-1 error over vanilla \acp{vit}. Hybrid-\acp{vit} have lower top-1 error than vanilla \acp{vit} and \acp{cnn} due to its power of identifying global and local features. Figure \ref{fig:adv_defense} shows examples of the preprocessing process for \ac{pgd}-$L_\infty$ attacks. 

\subsection{Robustness under \ac{eot} attack}
\ac{eot} is a framework that constructs \acp{ae} that remain adversarial over a chosen transformation distribution $T$, i.e. the preprocessing defenses. When processing defenses are used the stochastic gradients issue of the classifier $f(.)$ arises and hence, to have successful attack, it is necessary to estimate the gradient over the expected transformation to the input $t(x)$, where, $t(.)$ is the transformation function. \ac{eot} optimizes the expectation over the transformation $\mathbb{e}_{t\sim T} f(t(x))$ which can be solved by minimizing the expected perceived distance as seen by the classifier $\mathbb{E}_{t\sim T} [d(t(x^\prime), t(x))]$, where $d(.)$ is the distance function.

We consider the distribution of transformations that includes \ac{ss}, \ac{nlm}, \ac{tvm}, \ac{jpg}, and \ac{cr}. We 1) use 1000 images from the ImageNet-1k validation set, 2) generate \acp{ae} using \acp{pgd}-$L_\infty$ $\epsilon=4/255$ and apply the preprocessing defense methods for the generated \acp{ae}, 3) use \ac{eot} to synthesize \acp{ae} that are robust over the given distribution $T$ and apply the preprocessing defense methods for the synthesized \acp{ae}.  Table \ref{tab:asr_eot_defense} shows the top-1 error for the target models against the preprocessed \acp{ae} with and without  considering the \ac{eot}.

It is clear that the \ac{eot} kept the input samples as adversarial over the tested transformations except for the \acf{cr} transformation. For \ac{ss} and \ac{jpg} , the top-1 error is highly increased in \ac{t2t} and ResNets models and the \ac{t2t} models show less top-1 error. While for \ac{nlm}, the small \ac{vit}, \ac{t2t}, and \ac{tnt} models have less top-1 error compared to other models. It is interesting to notice that, with the use \ac{eot}, the ResNet models have less top-1 error than \ac{vit}-B/L models. One reason for that is that the transformation under the \ac{eot} highly impacts the global structure of the input sample. For \ac{tvm}, the top-1 error of \ac{t2t} models is less than other target models. The top-1 error for using \ac{cr} under the \ac{eot} is comparable to the top-1 error without \ac{cr}. One reason for that is that \ac{cr}, under any framework, targets restructuring the adversarial sample by re-positioning the input pixels.

In general, we conclude that \ac{t2t} and \ac{tnt} models show better robustness than vanilla \acp{vit} and ResNets under the \ac{eot} robustness test.

\section{Conclusion}
In this paper, we have studied the robustness of vision transformers and their variants and compared them with \acp{cnn}. Our analysis showed 1) either vanilla \acp{vit} or hybrid-\acp{vit} are more robust than \acp{cnn} against $L_p$-based attacks and \ac{ccp} attacks. We analysed the energy spectrum of \ac{dct} decomposition of the perturbations and the different visual quality measures. 2) \ac{ccp} can be used as a preprocessing defense method. 3) Increasing the number of attention blocks will increase the robustness against the transfer attacks but not against white box attacks. 4) Enhancing \ac{vit} tokenization might not increase the robustness against the \acp{ae} but will increase the robustness under the preprocessed \acp{ae}.

\ifCLASSOPTIONcompsoc
  \section*{Acknowledgments}
\else
  \section*{Acknowledgment}
\fi
The project is funded by both R\'egion Bretagne (Brittany region), France, and direction g\'en\'erale de l'armement (DGA).

\ifCLASSOPTIONcaptionsoff
  \newpage
\fi





\bibliographystyle{IEEEtran}
\bibliography{refs}

\begin{thebibliography}{10}
\providecommand{\url}[1]{#1}
\csname url@samestyle\endcsname
\providecommand{\newblock}{\relax}
\providecommand{\bibinfo}[2]{#2}
\providecommand{\BIBentrySTDinterwordspacing}{\spaceskip=0pt\relax}
\providecommand{\BIBentryALTinterwordstretchfactor}{4}
\providecommand{\BIBentryALTinterwordspacing}{\spaceskip=\fontdimen2\font plus
\BIBentryALTinterwordstretchfactor\fontdimen3\font minus
  \fontdimen4\font\relax}
\providecommand{\BIBforeignlanguage}[2]{{%
\expandafter\ifx\csname l@#1\endcsname\relax
\typeout{** WARNING: IEEEtran.bst: No hyphenation pattern has been}%
\typeout{** loaded for the language `#1'. Using the pattern for}%
\typeout{** the default language instead.}%
\else
\language=\csname l@#1\endcsname
\fi
#2}}
\providecommand{\BIBdecl}{\relax}
\BIBdecl

\bibitem{he2016deep}
K.~He, X.~Zhang, S.~Ren, and J.~Sun, ``Deep residual learning for image
  recognition,'' in \emph{Proceedings of the IEEE conference on computer vision
  and pattern recognition}, 2016, pp. 770--778.

\bibitem{yuan2021tokens}
L.~Yuan, Y.~Chen, T.~Wang, W.~Yu, Y.~Shi, F.~E.~H. Tay, J.~Feng, and S.~Yan,
  ``{Tokens-to-Token} {ViT}: Training vision transformers from scratch on
  imagenet,'' \emph{CoRR}, vol. abs/2101.11986, 2021.

\bibitem{vaswani2017attention}
A.~Vaswani, N.~Shazeer, N.~Parmar, J.~Uszkoreit, L.~Jones, A.~N. Gomez,
  L.~Kaiser, and I.~Polosukhin, ``Attention is all you need,'' in
  \emph{Advances in Neural Information Processing Systems 30, December 4-9,
  2017, Long Beach, CA, {USA}}, 2017, pp. 5998--6008.

\bibitem{dosovitskiy2020image}
A.~Dosovitskiy, L.~Beyer, A.~Kolesnikov, D.~Weissenborn, X.~Zhai,
  T.~Unterthiner, M.~Dehghani, M.~Minderer, G.~Heigold, S.~Gelly, J.~Uszkoreit,
  and N.~Houlsby, ``An image is worth 16x16 words: Transformers for image
  recognition at scale,'' in \emph{9th International Conference on Learning
  Representations, {ICLR} 2021, Virtual Event, Austria, May 3-7, 2021}.\hskip
  1em plus 0.5em minus 0.4em\relax OpenReview.net, 2021.

\bibitem{sun2017revisiting}
C.~Sun, A.~Shrivastava, S.~Singh, and A.~Gupta, ``Revisiting unreasonable
  effectiveness of data in deep learning era,'' in \emph{{IEEE} International
  Conference on Computer Vision, {ICCV} 2017, Venice, Italy, October 22-29,
  2017}.\hskip 1em plus 0.5em minus 0.4em\relax {IEEE} Computer Society, 2017,
  pp. 843--852.

\bibitem{imagenet_cvpr09}
J.~Deng, W.~Dong, R.~Socher, L.-J. Li, K.~Li, and L.~Fei-Fei, in \emph{CVPR09},
  2009.

\bibitem{han2021transformer}
K.~Han, A.~Xiao, E.~Wu, J.~Guo, C.~Xu, and Y.~Wang, ``Transformer in
  transformer,'' \emph{CoRR}, vol. abs/2103.00112, 2021.

\bibitem{wu2021cvt}
H.~Wu, B.~Xiao, N.~Codella, M.~Liu, X.~Dai, L.~Yuan, and L.~Zhang, ``{CvT}:
  Introducing convolutions to vision transformers,'' \emph{CoRR}, vol.
  abs/2103.15808, 2021.

\bibitem{simonyan2014very}
K.~Simonyan and A.~Zisserman, ``Very deep convolutional networks for
  large-scale image recognition,'' in \emph{3rd International Conference on
  Learning Representations, {ICLR} 2015, San Diego, CA, USA, May 7-9, 2015,
  Conference Track Proceedings}, Y.~Bengio and Y.~LeCun, Eds., 2015.

\bibitem{szegedy2016rethinking}
C.~Szegedy, V.~Vanhoucke, S.~Ioffe, J.~Shlens, and Z.~Wojna, ``Rethinking the
  inception architecture for computer vision,'' in \emph{Proceedings of the
  IEEE conference on computer vision and pattern recognition}, 2016, pp.
  2818--2826.

\bibitem{howard2017mobilenets}
A.~G. Howard, M.~Zhu, B.~Chen, D.~Kalenichenko, W.~Wang, T.~Weyand,
  M.~Andreetto, and H.~Adam, ``{MobileNets}: Efficient convolutional neural
  networks for mobile vision applications,'' \emph{arXiv preprint
  arXiv:1704.04861}, 2017.

\bibitem{szegedy2013intriguing}
C.~Szegedy, W.~Zaremba, I.~Sutskever, J.~Bruna, D.~Erhan, I.~J. Goodfellow, and
  R.~Fergus, ``Intriguing properties of neural networks,'' in \emph{2nd
  International Conference on Learning Representations, {ICLR} 2014, Banff, AB,
  Canada, April 14-16, 2014, Conference Track Proceedings}, 2014.

\bibitem{carlini2017adversarial}
N.~Carlini and D.~Wagner, ``Adversarial examples are not easily detected:
  Bypassing ten detection methods,'' in \emph{Proceedings of the 10th ACM
  Workshop on Artificial Intelligence and Security}, 2017, pp. 3--14.

\bibitem{ilyas2019adversarial}
A.~Ilyas, S.~Santurkar, D.~Tsipras, L.~Engstrom, B.~Tran, and A.~Madry,
  ``Adversarial examples are not bugs, they are features,'' in \emph{Advances
  in Neural Information Processing Systems}, 2019, pp. 125--136.

\bibitem{akhtar2018threat}
N.~Akhtar and A.~Mian, ``Threat of adversarial attacks on deep learning in
  computer vision: A survey,'' \emph{IEEE Access}, vol.~6, pp.
  14\,410--14\,430, 2018.

\bibitem{hao2020adversarial}
H.~X. Y.~M. Hao-Chen, L.~D. Deb, H.~L. J.-L.~T. Anil, and K.~Jain,
  ``Adversarial attacks and defenses in images, graphs and text: A review,''
  \emph{International Journal of Automation and Computing}, vol.~17, no.~2, pp.
  151--178, 2020.

\bibitem{bhojanapalli2021understanding}
S.~Bhojanapalli, A.~Chakrabarti, D.~Glasner, D.~Li, T.~Unterthiner, and
  A.~Veit, ``Understanding robustness of transformers for image
  classification,'' \emph{CoRR}, vol. abs/2103.14586, 2021.

\bibitem{shao2021adversarial}
R.~Shao, Z.~Shi, J.~Yi, P.~Chen, and C.~Hsieh, ``On the adversarial robustness
  of visual transformers,'' \emph{CoRR}, vol. abs/2103.15670, 2021.

\bibitem{mahmood2021robustness}
K.~Mahmood, R.~Mahmood, and M.~van Dijk, ``On the robustness of vision
  transformers to adversarial examples,'' 2021.

\bibitem{paul2021vision}
S.~Paul and P.~Chen, ``Vision transformers are robust learners,'' \emph{CoRR},
  vol. abs/2105.07581, 2021.

\bibitem{yuan2019adversarial}
X.~Yuan, P.~He, Q.~Zhu, and X.~Li, ``Adversarial examples: Attacks and defenses
  for deep learning,'' \emph{IEEE transactions on neural networks and learning
  systems}, vol.~30, no.~9, pp. 2805--2824, 2019.

\bibitem{chakraborty2018adversarial}
A.~Chakraborty, M.~Alam, V.~Dey, A.~Chattopadhyay, and D.~Mukhopadhyay,
  ``Adversarial attacks and defences: {A} survey,'' \emph{CoRR}, vol.
  abs/1810.00069, 2018.

\bibitem{aldahdooh2021adversarial}
A.~Aldahdooh, W.~Hamidouche, S.~A. Fezza, and O.~D{\'{e}}forges, ``Adversarial
  example detection for {DNN} models: {A} review,'' \emph{CoRR}, vol.
  abs/2105.00203, 2021.

\bibitem{alamri2021transformer}
F.~Alamri, S.~Kalkan, and N.~Pugeault, ``Transformer-encoder detector module:
  Using context to improve robustness to adversarial attacks on object
  detection,'' in \emph{2020 25th International Conference on Pattern
  Recognition (ICPR)}.\hskip 1em plus 0.5em minus 0.4em\relax IEEE, 2021, pp.
  9577--9584.

\bibitem{ren2016faster}
S.~Ren, K.~He, R.~B. Girshick, and J.~Sun, ``Faster {R-CNN:} towards real-time
  object detection with region proposal networks,'' \emph{{IEEE} Trans. Pattern
  Anal. Mach. Intell.}, vol.~39, no.~6, pp. 1137--1149, 2017.

\bibitem{goodfellow2014explaining}
I.~J. Goodfellow, J.~Shlens, and C.~Szegedy, ``Explaining and harnessing
  adversarial examples,'' in \emph{3rd International Conference on Learning
  Representations, {ICLR} 2015, San Diego, CA, USA, May 7-9, 2015, Conference
  Track Proceedings}, 2015.

\bibitem{madry2017towards}
A.~Madry, A.~Makelov, L.~Schmidt, D.~Tsipras, and A.~Vladu, ``Towards deep
  learning models resistant to adversarial attacks,'' in \emph{6th
  International Conference on Learning Representations, {ICLR} 2018, Vancouver,
  Canada, 2018}.\hskip 1em plus 0.5em minus 0.4em\relax OpenReview.net, 2018.

\bibitem{imagenet21}
J.~Deng, W.~Dong, R.~Socher, L.-J. Li, K.~Li, and L.~Fei-Fei, ``{ImageNet: A
  large-scale hierarchical image database},'' in \emph{2009 IEEE Conference on
  Computer Vision and Pattern Recognition}, 2009, pp. 248--255.

\bibitem{kolesnikov2020big}
A.~Kolesnikov, L.~Beyer, X.~Zhai, J.~Puigcerver, J.~Yung, S.~Gelly, and
  N.~Houlsby, ``Big transfer {(BiT)}: General visual representation learning,''
  in \emph{Computer Vision - {ECCV} 2020 - 16th European Conference, Glasgow,
  UK, August 23-28, 2020, Proceedings, Part {V}}, ser. Lecture Notes in
  Computer Science, vol. 12350.\hskip 1em plus 0.5em minus 0.4em\relax
  Springer, 2020, pp. 491--507.

\bibitem{croce2020reliable}
F.~Croce and M.~Hein, ``Reliable evaluation of adversarial robustness with an
  ensemble of diverse parameter-free attacks,'' in \emph{Proceedings of the
  37th International Conference on Machine Learning, {ICML} 2020, 13-18 July
  2020, Virtual Event}, ser. Proceedings of Machine Learning Research, vol.
  119.\hskip 1em plus 0.5em minus 0.4em\relax {PMLR}, 2020, pp. 2206--2216.

\bibitem{carlini2017towards}
N.~Carlini and D.~Wagner, ``Towards evaluating the robustness of neural
  networks,'' in \emph{2017 ieee symposium on security and privacy (sp)}.\hskip
  1em plus 0.5em minus 0.4em\relax IEEE, 2017, pp. 39--57.

\bibitem{dong2018boosting}
Y.~Dong, F.~Liao, T.~Pang, H.~Su, J.~Zhu, X.~Hu, and J.~Li, ``Boosting
  adversarial attacks with momentum,'' in \emph{2018 {IEEE} Conference on
  Computer Vision and Pattern Recognition, {CVPR} 2018, Salt Lake City, UT,
  USA, June 18-22, 2018}.\hskip 1em plus 0.5em minus 0.4em\relax Computer
  Vision Foundation / {IEEE} Computer Society, 2018, pp. 9185--9193.

\bibitem{athalye2018obfuscated}
A.~Athalye, N.~Carlini, and D.~A. Wagner, ``Obfuscated gradients give a false
  sense of security: Circumventing defenses to adversarial examples,'' in
  \emph{Proceedings of the 35th International Conference on Machine Learning,
  {ICML} 2018, Stockholmsm{\"{a}}ssan, Stockholm, Sweden, July 10-15, 2018},
  ser. Proceedings of Machine Learning Research, vol.~80.\hskip 1em plus 0.5em
  minus 0.4em\relax {PMLR}, 2018, pp. 274--283.

\bibitem{hendrycks2020gaussian}
D.~Hendrycks and K.~Gimpel, ``Gaussian error linear units {(GELUs)},''
  \emph{CoRR}, vol. abs/1606.08415, 2016.

\bibitem{rw2019timm}
R.~Wightman, ``Pytorch image models,''
  \url{https://github.com/rwightman/pytorch-image-models}, 2019.

\bibitem{xie2017aggregated}
S.~Xie, R.~B. Girshick, P.~Doll{\'{a}}r, Z.~Tu, and K.~He, ``Aggregated
  residual transformations for deep neural networks,'' in \emph{2017 {IEEE}
  Conference on Computer Vision and Pattern Recognition, {CVPR} 2017, Honolulu,
  HI, USA, July 21-26, 2017}.\hskip 1em plus 0.5em minus 0.4em\relax {IEEE}
  Computer Society, 2017, pp. 5987--5995.

\bibitem{li2020survey}
Z.~Li, W.~Yang, S.~Peng, and F.~Liu, ``A survey of convolutional neural
  networks: Analysis, applications, and prospects,'' \emph{CoRR}, vol.
  abs/2004.02806, 2020.

\bibitem{papernot2016limitations}
N.~Papernot, P.~McDaniel, S.~Jha, M.~Fredrikson, Z.~B. Celik, and A.~Swami,
  ``The limitations of deep learning in adversarial settings,'' in \emph{2016
  IEEE European symposium on security and privacy (EuroS\&P)}.\hskip 1em plus
  0.5em minus 0.4em\relax IEEE, 2016, pp. 372--387.

\bibitem{moosavi2017universal}
S.-M. Moosavi-Dezfooli, A.~Fawzi, O.~Fawzi, and P.~Frossard, ``Universal
  adversarial perturbations,'' in \emph{Proceedings of the IEEE conference on
  computer vision and pattern recognition}, 2017, pp. 1765--1773.

\bibitem{andriushchenko2020square}
M.~Andriushchenko, F.~Croce, N.~Flammarion, and M.~Hein, ``Square attack: a
  query-efficient black-box adversarial attack via random search,'' in
  \emph{European Conference on Computer Vision}.\hskip 1em plus 0.5em minus
  0.4em\relax Springer, 2020, pp. 484--501.

\bibitem{chen2020rays}
J.~Chen and Q.~Gu, ``Rays: A ray searching method for hard-label adversarial
  attack,'' in \emph{Proceedings of the 26th ACM SIGKDD International
  Conference on Knowledge Discovery \& Data Mining}, 2020, pp. 1739--1747.

\bibitem{colorchannel}
J.~Kantipudi, S.~R. Dubey, and S.~Chakraborty, ``Color channel perturbation
  attacks for fooling convolutional neural networks and a defense against such
  attacks,'' \emph{IEEE Transactions on Artificial Intelligence}, vol.~1,
  no.~2, pp. 181--191, 2020.

\bibitem{croce2020minimally}
F.~Croce and M.~Hein, ``Minimally distorted adversarial examples with a fast
  adaptive boundary attack,'' in \emph{International Conference on Machine
  Learning}.\hskip 1em plus 0.5em minus 0.4em\relax PMLR, 2020, pp. 2196--2205.

\bibitem{xu2017feature}
W.~Xu, D.~Evans, and Y.~Qi, ``Feature squeezing: Detecting adversarial examples
  in deep neural networks,'' in \emph{25th Annual Network and Distributed
  System Security Symposium, {NDSS} 2018, San Diego, California, USA, February
  18-21, 2018}.\hskip 1em plus 0.5em minus 0.4em\relax The Internet Society,
  2018.

\bibitem{guo2017countering}
C.~Guo, M.~Rana, M.~Ciss{\'{e}}, and L.~van~der Maaten, ``Countering
  adversarial images using input transformations,'' in \emph{6th International
  Conference on Learning Representations, {ICLR} 2018, Vancouver, BC, Canada,
  April 30 - May 3, 2018, Conference Track Proceedings}.\hskip 1em plus 0.5em
  minus 0.4em\relax OpenReview.net, 2018.

\bibitem{dziugaite2016study}
G.~K. Dziugaite, Z.~Ghahramani, and D.~M. Roy, ``A study of the effect of {JPG}
  compression on adversarial images,'' \emph{CoRR}, vol. abs/1608.00853, 2016.

\bibitem{das2017keeping}
N.~Das, M.~Shanbhogue, S.~Chen, F.~Hohman, L.~Chen, M.~E. Kounavis, and D.~H.
  Chau, ``Keeping the bad guys out: Protecting and vaccinating deep learning
  with {JPEG} compression,'' \emph{CoRR}, vol. abs/1705.02900, 2017.

\bibitem{graese2016assessing}
A.~Graese, A.~Rozsa, and T.~E. Boult, ``Assessing threat of adversarial
  examples on deep neural networks,'' in \emph{2016 15th IEEE International
  Conference on Machine Learning and Applications (ICMLA)}.\hskip 1em plus
  0.5em minus 0.4em\relax IEEE, 2016, pp. 69--74.

\bibitem{donoho1994ideal}
D.~L. Donoho and J.~M. Johnstone, ``Ideal spatial adaptation by wavelet
  shrinkage,'' \emph{biometrics}, vol.~81, no.~3, pp. 425--455, 1994.

\bibitem{chambolle2004algorithm}
A.~Chambolle, ``An algorithm for total variation minimization and
  applications,'' \emph{Journal of Mathematical imaging and vision}, vol.~20,
  no.~1, pp. 89--97, 2004.

\bibitem{athalye2018synthesizing}
A.~Athalye, L.~Engstrom, A.~Ilyas, and K.~Kwok, ``Synthesizing robust
  adversarial examples,'' in \emph{International conference on machine
  learning}.\hskip 1em plus 0.5em minus 0.4em\relax PMLR, 2018, pp. 284--293.

\bibitem{Selvaraju_2019}
R.~R. Selvaraju, M.~Cogswell, A.~Das, R.~Vedantam, D.~Parikh, and D.~Batra,
  ``{Grad-CAM}: Visual explanations from deep networks via gradient-based
  localization,'' \emph{International Journal of Computer Vision}, vol. 128,
  no.~2, p. 336–359, Oct 2019.

\bibitem{ortizjimenez2020hold}
G.~Ortiz{-}Jim{\'{e}}nez, A.~Modas, S.~Moosavi{-}Dezfooli, and P.~Frossard,
  ``Hold me tight! influence of discriminative features on deep network
  boundaries,'' in \emph{Advances in Neural Information Processing Systems 33:
  NeurIPS 2020, December 6-12, 2020, virtual}, 2020.

\bibitem{fezza2019perceptual}
S.~A. Fezza, Y.~Bakhti, W.~Hamidouche, and O.~D{\'e}forges, ``Perceptual
  evaluation of adversarial attacks for cnn-based image classification,'' in
  \emph{2019 Eleventh International Conference on Quality of Multimedia
  Experience (QoMEX)}.\hskip 1em plus 0.5em minus 0.4em\relax IEEE, 2019, pp.
  1--6.

\bibitem{wang2004image}
Z.~Wang, A.~C. Bovik, H.~R. Sheikh, and E.~P. Simoncelli, ``Image quality
  assessment: from error visibility to structural similarity,'' \emph{IEEE
  transactions on image processing}, vol.~13, no.~4, pp. 600--612, 2004.

\bibitem{larson2010most}
E.~C. Larson and D.~M. Chandler, ``Most apparent distortion: full-reference
  image quality assessment and the role of strategy,'' \emph{Journal of
  electronic imaging}, vol.~19, no.~1, p. 011006, 2010.

\bibitem{jacobgilpytorchcam}
J.~Gildenblat and contributors, ``Pytorch library for {CAM} methods,''
  \url{https://github.com/jacobgil/pytorch-grad-cam}, 2021.

\bibitem{sharma2019effectiveness}
Y.~Sharma, G.~W. Ding, and M.~A. Brubaker, ``On the effectiveness of low
  frequency perturbations,'' in \emph{Proceedings of the Twenty-Eighth
  International Joint Conference on Artificial Intelligence, {IJCAI} 2019,
  Macao, China, August 10-16, 2019}.\hskip 1em plus 0.5em minus 0.4em\relax
  ijcai.org, 2019, pp. 3389--3396.

\end{thebibliography}

\vfill







\end{document}